\documentclass[twoside,11pt]{article}
\usepackage{graphicx} %
\usepackage[hidelinks,breaklinks=true]{hyperref}
\usepackage{arxiv, theapa, rawfonts}
\usepackage{url}

\usepackage{tikz}
\usetikzlibrary{shapes,arrows}
\usepackage{color-edits}

\usepackage{enumitem}

\newcounter{rowcount}
\newcounter{rowcountrecs}
\usepackage{xcolor}
\usepackage{longtable,booktabs,array}

\makeatletter
\newbox\LT@firstfoot
\def\endfirstfoot{\LT@end@hd@ft\LT@firstfoot}
\newdimen\LT@footdiff
\def\LT@start{%
  \let\LT@start\endgraf
  \endgraf\penalty\z@
  \vskip\LTpre\endgraf
  \LT@footdiff-\ht\LT@foot
  \advance\LT@footdiff\ht\LT@firstfoot
  \dimen@\pagetotal
  \advance\dimen@ \ht\ifvoid\LT@firsthead\LT@head\else\LT@firsthead\fi
  \advance\dimen@ \dp\ifvoid\LT@firsthead\LT@head\else\LT@firsthead\fi
  \advance\dimen@ \ht\ifvoid\LT@firstfoot\LT@foot\else\LT@firstfoot\fi
  \dimen@ii\vfuzz
  \vfuzz\maxdimen
  \setbox\tw@\copy\z@
  \setbox\tw@\vsplit\tw@ to \ht\@arstrutbox
  \setbox\tw@\vbox{\unvbox\tw@}%
  \vfuzz\dimen@ii
  \advance\dimen@ \ht
      \ifdim\ht\@arstrutbox>\ht\tw@\@arstrutbox\else\tw@\fi
  \advance\dimen@\dp
      \ifdim\dp\@arstrutbox>\dp\tw@\@arstrutbox\else\tw@\fi
  \advance\dimen@ -\pagegoal
  \ifdim \dimen@>\z@\vfil\break\fi
  \global\@colroom\@colht
  \ifvoid\LT@firstfoot
    \ifvoid\LT@foot
    \else
      \advance\vsize-\ht\LT@foot
      \global\advance\@colroom-\ht\LT@foot
      \dimen@\pagegoal\advance\dimen@-\ht\LT@foot\pagegoal\dimen@
      \maxdepth\z@
    \fi
  \else
    \advance\vsize-\ht\LT@firstfoot
    \global\advance\@colroom-\ht\LT@firstfoot
    \dimen@\pagegoal\advance\dimen@-\ht\LT@firstfoot\pagegoal\dimen@
    \maxdepth\z@
  \fi
  \ifvoid\LT@firsthead\copy\LT@head\else\box\LT@firsthead\fi\nobreak
  \output{\LT@output}%
}
\def\LT@output{%
  \ifnum\outputpenalty <-\@Mi
    \ifnum\outputpenalty > -\LT@end@pen
      \LT@err{floats and marginpars not allowed in a longtable}\@ehc
    \else
      \setbox\z@\vbox{\unvbox\@cclv}%
      \ifdim \ht\LT@lastfoot>\ht\LT@foot
        \dimen@\pagegoal
        \advance\dimen@-\ht\LT@lastfoot
        \ifdim\dimen@<\ht\z@
          \setbox\@cclv\vbox{\unvbox\z@\copy\LT@foot\vss}%
          \@makecol
          \@outputpage
          \setbox\z@\vbox{\box\LT@head}%
        \fi
      \fi  
      \global\@colroom\@colht
      \global\vsize\@colht   
      \vbox
        {\unvbox\z@\box\ifvoid\LT@lastfoot\LT@foot\else\LT@lastfoot\fi}%
    \fi
  \else
    \ifvoid\LT@firstfoot
      \setbox\@cclv\vbox{\unvbox\@cclv\copy\LT@foot\vss}%
      \@makecol
      \@outputpage
      \global\vsize\@colroom
    \else
      \setbox\@cclv\vbox{\unvbox\@cclv\box\LT@firstfoot\vss}%
      \@makecol
      \@outputpage
      \global\advance\@colroom\LT@footdiff
      \global\vsize\@colroom
    \fi
    \copy\LT@head\nobreak
  \fi
}
\makeatother

\usepackage{multirow}
\usepackage{calc} %

\def\MultiplyA#1,{#1\MultiplyB}
\def\MultiplyB#1,{\ifx,#1,\else\cdot#1\expandafter\MultiplyB\fi}

\usepackage{wrapfig}

\ShortHeadings{Evaluating Task-oriented Dialogue Systems}{Braggaar, Liebrecht, Van Miltenburg, Krahmer}
\firstpageno{1}

\title{Evaluating Task-oriented Dialogue Systems: \\ A Systematic Review of Measures, Constructs and their Operationalisations}

\author{\name Anouck Braggaar \email A.R.Y.Braggaar@tilburguniversity.edu
\AND
\name Christine Liebrecht \email C.C.Liebrecht@tilburguniversity.edu
\AND
\name Emiel van Miltenburg \email C.W.J.vanMiltenburg@tilburguniversity.edu
\AND
\name Emiel Krahmer \email E.J.Krahmer@tilburguniversity.edu\\
\addr Department of Communication and Cognition,\\
Tilburg University, P.O. Box 90153, NL-5000 LE,\\
Tilburg, The Netherlands
}

\begin{document}

\maketitle

\begin{abstract}
This review gives an extensive overview of evaluation methods for task-oriented dialogue systems, paying special attention to practical applications of dialogue systems, for example for customer service. The review (1) provides an overview of the used constructs and metrics in previous work, (2) discusses challenges in the context of dialogue system evaluation and (3) develops a research agenda for the future of dialogue system evaluation. We conducted a systematic review of four databases (ACL, ACM, IEEE and Web of Science), which after screening resulted in 122 studies. Those studies were carefully analysed for the constructs and methods they proposed for evaluation. We found a wide variety in both constructs and methods. Especially the operationalisation is not always clearly reported. Newer developments concerning large language models are discussed in two contexts: to power dialogue systems and to use in the evaluation process. We hope that future work will take a more critical approach to the operationalisation and specification of the used constructs. To work towards this aim, this review ends with recommendations for evaluation and suggestions for outstanding questions.
\end{abstract}

\section{Introduction}
\label{sec:introduction}
Over the last few years, dialogue systems have become much more robust, and practical applications are within reach and sometimes even in existence already. Evaluation of such systems remains an important task. However, many challenges emerge when trying to evaluate a dialogue system. Especially the selection of constructs and methods seems to be a challenging task. Evaluation needs to be done with great care and at the same time there seems to be a lack of standardisation, regarding both metrics and constructs (as mentioned for example by \shortciteR{casas2020trends}). As will become clear there is a high diversity in terms used and how to operationalise these terms. Next to this, it is hard to define what a `good' dialogue or dialogue system is and how this aspect could eventually be captured in a measure \shortcite{deriu2021survey}. Proper evaluation of dialogue systems is important as a good working system is essential for both the user and the organisation behind the dialogue system.

\begin{figure}[t]
   \centering
   \includegraphics[width=0.55\textwidth]{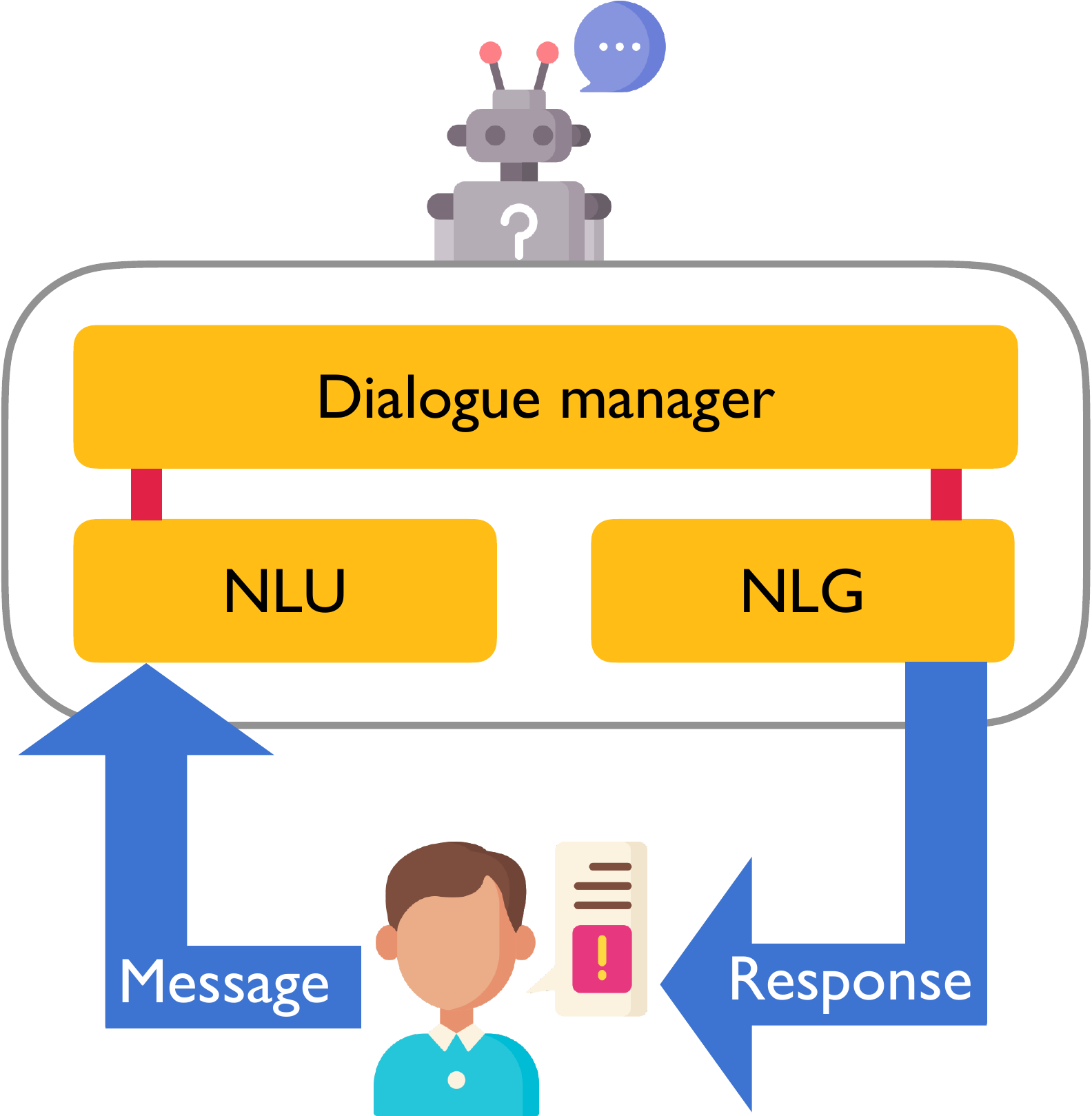}
   \caption{Simplified depiction of an interaction with a task-oriented dialogue system. The user comes to the system with a particular problem that they would like to solve. Through a series of messages to and responses from the dialogue system, both interlocutors work towards finding a resolution. Internally, messages are traditionally processed through a Natural Language Understanding (NLU) module, after which a dialogue manager updates the internal state of the system, and selects an appropriate response, which is then realised by the Natural Language Generation (NLG) module. (Icons via Freepik.com.)}
   \label{fig:dsinteraction}
\end{figure}

Dialogue systems (schematically depicted in Figure~\ref{fig:dsinteraction}) are employed within multiple domains, such as health care (e.g. \shortciteR{chung2021chatbot,crutzen2011artificially}), e-commerce (e.g. \shortciteR{bhawiyuga2017design,qiu2009evaluating}), customer service (e.g. \shortciteR{eren2021determinants}) and insurance (e.g. \shortciteR{nuruzzaman2020intellibot}). When dialogue systems are used for such practical applications, evaluation becomes both more complex (because the contribution towards achieving the task needs to be included in the evaluation) and more important (since a system that fails to achieve its task is bound to be rejected in practice). Therefore, in this review we will pay special attention to such applied evaluation of dialogue systems, with a focus on customer service applications (in any domain). Within customer service, dialogue systems do not just interact with users to answer their questions or help with basic tasks (i.e., task-based systems), but they also serve as \emph{brand ambassadors}.\footnote{For a discussion of this concept, see \citeA{harris2001corporate}.} 
Both users and organisations perceive customer service chatbots as an extension - and sometimes even as a (partial) replacement - of the human customer service agent with whom (potential) customers could chat or phone call with \cite{zhang2023organizational}. Thus, whatever these systems do also reflects on the corporate image of the organisation that they serve. Bad experiences with a dialogue system may give (potential) customers a bad impression of the organisation as a whole, or they may not want to use the system again. We should thus not just focus on the task alone (do the users achieve their goal?), but instead take a broader perspective (how do different users \emph{experience} the system?). This perspective involves all kinds of other metrics relevant in customer service, such as assurance, coherence, and civility \cite{ghosh2020webcare}. These metrics are hard to capture through any single evaluation measure, and therefore it is important to consider the full evaluation spectrum: from intrinsic to extrinsic measures, from manual to automatic evaluation, and so on \cite{resnik2010evaluation}. This paper provides an overview of different evaluation metrics, organised by the different constructs (i.e. quality dimensions) that one might be interested in when evaluating their dialogue system.

\subsection{Dialogue systems}
\label{ds}
There is a high degree of variation in the terminology used to refer to task-oriented dialogue systems. Common labels are: \emph{dialogue system, chatbot, conversational agent, conversational AI, virtual assistant} and \emph{digital agent}. While \citeA{jurafsky2021} argue that chatbots are distinct from dialogue systems as chatbots are designed for more unstructured conversations, in the literature the aforementioned terms tend to be used interchangeably. As this review puts an emphasis on task-oriented systems in the customer service domain, both the terms `dialogue systems' and `chatbot' will be used interchangeably. But, papers that make use of other related terms (such as `customer service chatbot' or `customer care agent') will of course also be covered.

ELIZA \cite{weizenbaum_eliza} is seen as the first chatbot to have been developed. It was based on a small set of rules and keywords, enabling it to respond to users with either a pre-programmed response or a variation of the user's own utterances. Although Weizenbaum has repeatedly stated that ELIZA was created as a \emph{parody} of Rogerian psychotherapists,\footnote{See \citeA{yao2023person} for a brief introduction to Rogerian psychotherapy.} others nonetheless took ELIZA quite seriously, as a first step towards automating psychological treatment (\citeA{Weizenbaum1976computer} cites \shortciteA{colby1966computer} as an example). More importantly for our purposes, and again to Weizenbaum's surprisal, people started to have deep conversations with ELIZA, and were quick to anthropomorphise the system \cite{Weizenbaum1976computer}. This shows the impact that dialogue systems can have on users, even with relatively simple means. To this day, many chatbots still use similar rules and scripted responses to have meaningful conversations with users. A final observation \citeA{Weizenbaum1976computer} made (perhaps connected to the anthropomorphisation of ELIZA) was ``the spread of a belief that [ELIZA] demonstrated a general solution to the problem of computer understanding of natural language'' \cite[p.\thinspace 7]{Weizenbaum1976computer}, even though this is demonstrably false. We currently see a similar kind of wishful thinking around the performance of Large Language Models \cite{mitchell2021why}. To us, these observations highlight the importance of both (i) critically thinking about what it \emph{means} for a system to have particular cognitive/linguistic abilities, and (ii) the use and developments of valid and reliable evaluation methods that test the abilities that a system has, and the impact that the system has on the user.

Nowadays, task-oriented dialogue systems can be much more complex. The processing and generating components shown in Figure~\ref{fig:dsinteraction} may vary from straightforward rule-based to more complex machine learning approaches \shortcite{harms2018approaches}. Different input methods can be used, all with different processing methods. Users can click on buttons (the dialogue system will then by necessity follow a straightforward predefined script) or in some cases make use of free input fields (which needs intent-recognition for handling the user's request). Similarly, in a domain like customer service the responses of the system are often still predefined (which results in scripted conversations), although lately there has been a shift towards replies that are generated. This actually shows the difference between the field of customer service systems and the more applied domains. Where in customer service the traditional approaches are still frequently used, within NLP more cutting edge technologies are explored (such as neural networks and other more complex approaches). Although traditional approaches are still often used within customer service, concepts like reliability and privacy are of great importance. In our view, regardless the technology that is used, in all cases evaluation is an important factor to take into account. And although new technologies might result in more and different evaluation metrics (see \S\ref{subsub:newmodels} for more on this topic) the basic ideas behind evaluation remain the same.

\subsection{Constructs and measurement}\label{sec:constructsmeasurement}
\begin{figure}
    \centering
    \includegraphics[width=0.6\textwidth]{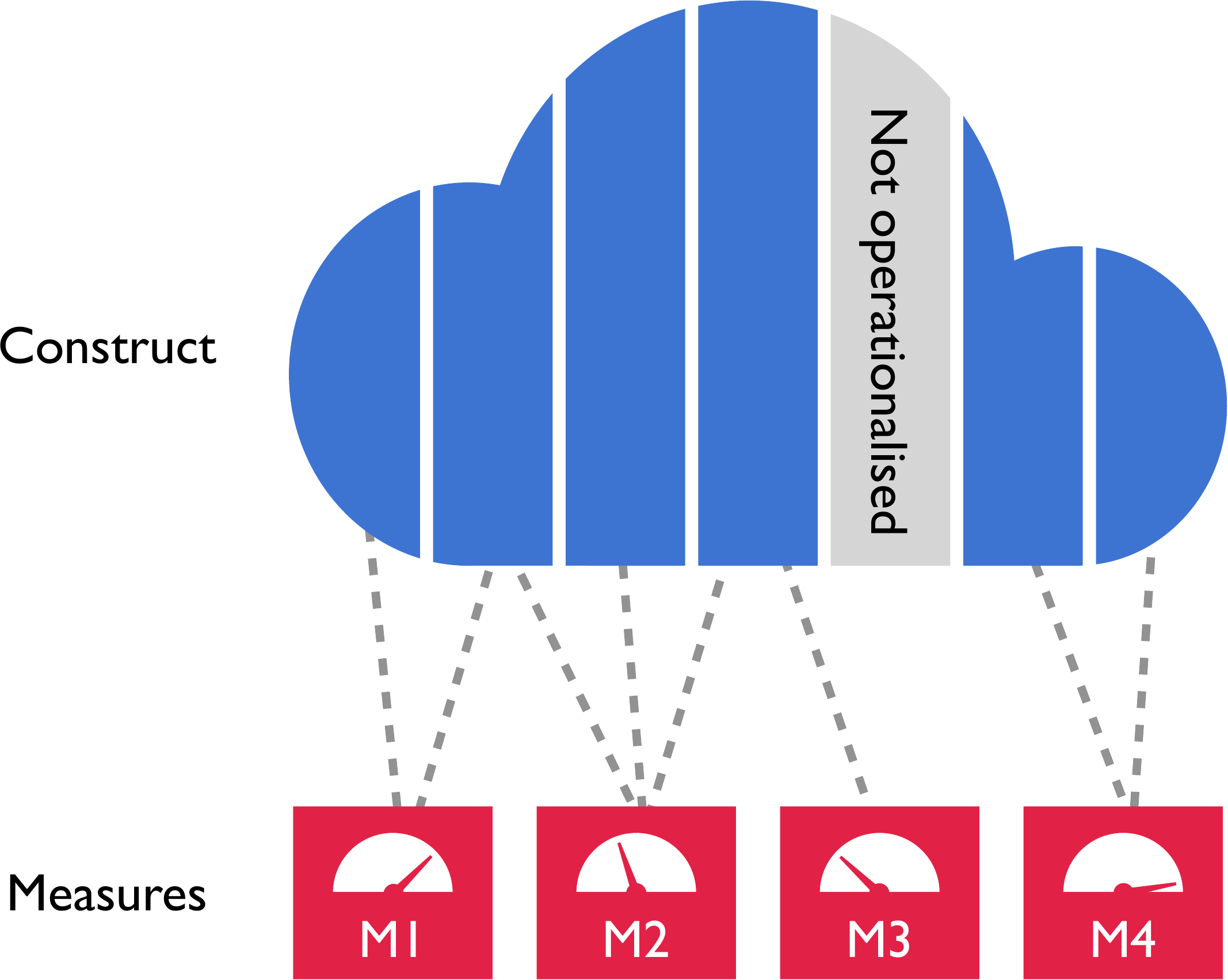}
    \caption{Different measures (M1\ldots M4) operationalising the same construct, capturing different aspects. We may obtain a fairly good coverage of the construct by combining different metrics, but some aspects may remain elusive.}
    \label{fig:operationalisation}
\end{figure}

Whenever we evaluate a dialogue system, the goal is to characterise the external behaviour of the system, its internal workings, or the effects that the system has on either its users or on other processes that the system is embedded in. For this characterisation, we rely on different ideas or concepts that help us explain the situation. For example, the high \emph{readability} and \emph{accuracy} of the generated responses might make a system \emph{easy to use}, which increases the users' \emph{efficiency} and \emph{intention to use} the system again. Following longstanding tradition in psychology, we refer to these ideas or concepts as \emph{constructs} \cite{cronbach1955construct}.\footnote{\citeA{strauss2009construct} provide an in-depth discussion of the origins and current debates around the idea of construct validity.} Since constructs are fairly abstract concepts, they are not directly measurable. To do so, we need to \emph{operationalise} them, i.e.\ ``define them in such a way that they can be measured'' \cite{Treadwell_Davis_2020}. For example, recent work 
 has already used these terms to define an measure model bias in NLP \shortcite{van2024undesirable}. Figure~\ref{fig:operationalisation} provides an illustration of this idea.

As will become evident, different studies on dialogue systems have studied different constructs, and different studies have operationalised the same constructs in different ways. We provide a systematic review of all constructs that have been studied, with references to the different papers that have operationalised those constructs in different ways.\footnote{Confusingly, different authors also (i) refer to the same constructs with different names, or (ii) refer to different constructs with the same names. This observation has also been made in NLG research by \shortciteA{howcroft-etal-2020-twenty}.} Through this construct-driven approach, we are also able to contrast different operationalisations of the same construct, showing how they each focus on different aspects of the ideas they aim to capture. Through our work, we provide a template for future researchers to critique the operationalisation of different constructs.

\subsection{Why this survey?}
\label{prev}
\subsubsection{Dialogue systems in the customer service domain}
\label{dscs}
Different domains all have specific characteristics and thus use (and need) different constructs and subsequently different evaluation metrics. In customer service, chatbots are increasingly employed and constantly under development (both in scientific and practical settings). \shortciteA{costello_lodolce_2022} predict that by 2027, chatbots will become the primary communication channel for a quarter of organisations. Therefore, there is a need to create an overview of evaluation methods and constructs for this task-oriented domain.

In addition, customer service chatbots have a set of characteristics that distinguish them from other chatbot contexts like health care or social chatbots. Users look for a quick response and correct information. They expect to be assisted in a friendly, often human like, manner. In all cases, a user (customer) has a certain task that needs to be accomplished with a  (task-based) chatbot. Often the conversations are text-based and last several turns until the query of the customer is answered (or if unsuccessful, the conversation results in a breakdown; see \citeR{braggaar2023repair}). When the conversation is finished, the customer has formed an opinion not only about the dialogue system itself but often also about the organisation that the chatbot represents. \shortciteA{pavone2023rage} for example show that customers blame the company more for negative outcomes than the chatbot itself. Thus, as good evaluations of organisations are important for their image, negative user experiences of a customer service chatbot conversation should be avoided. 

\begin{table}
\centering
\small
\begin{tabular}{@{}l|cccc ll|cccc}
 &
\rotatebox{270}{\textbf{Evaluation}} &
\rotatebox{270}{\textbf{Task-oriented}} &
\rotatebox{270}{\textbf{Constructs}} &
\rotatebox{270}{\textbf{Customer service}} 
 &
 &
 &
\rotatebox{270}{\textbf{Evaluation}} &
\rotatebox{270}{\textbf{Task-oriented}} &
\rotatebox{270}{\textbf{Constructs}} &
\rotatebox{270}{\textbf{Customer service}} \\
\cmidrule{0-4} \cmidrule(r){7-11}
\shortciteA{zhang2020recent} & -- & + & -- & --  & & \shortciteA{fan2020survey} & + & -- & -- & -- \\

\shortciteA{peng2019survey} & -- & + & -- & + & & 
\shortciteA{liu-etal-2016-evaluate} & + & -- & -- & -- \\

\shortciteA{syvanen2020conversational} & -- & + & -- & + & &
\shortciteA{maroengsit2019survey} & + & -- & -- & -- \\

\shortciteA{abushawar2016usefulness} & + & -- & + & -- & &
\shortciteA{yeh-etal-2021-comprehensive} & + & -- & -- & -- \\

\shortciteA{edwards1988evaluating} & + & -- & + & -- & &
\shortciteA{cui2020survey} & -- & -- & -- & -- \\

\shortciteA{finch-choi-2020-towards} & + & -- & + & -- & &
\shortciteA{abd2020technical} & + & + & -- & -- \\

\shortciteA{ren2019evaluation} & + & -- & + & -- & &
\shortciteA{federici2020inside} & + & + & -- & -- \\

\shortciteA{casas2020trends} & + & -- & + & -- & &
\shortciteA{jannach2021survey} &  -- & + & + & -- \\

\shortciteA{deriu2021survey} & + & -- & -- & -- & &
\textbf{This paper} & + & + & + & + \\
\bottomrule
\end{tabular}
\caption{The focus of previous overviews on (evaluation of) dialogue systems.}
\label{table:previous}
\end{table}

\subsubsection{Previous overviews on evaluation of dialogue systems}
\label{subsec:previouswork}
Many reviews on dialogue systems and dialogue system evaluation have been published. These reviews often have a different scope than the current review. There are reviews on dialogue systems in general, or specifically focusing on the question of evaluation of dialogue systems. Some reviews focus on specific technical aspects while other reviews narrow the scope by focusing on systems in a specific domain. Table~\ref{table:previous} provides an overview. In all cases, it can be observed that previous reviews vary in perspectives, metrics, definitions, and constructs - making these insights less applicable to task-based dialogue systems in the context of customer service. 

The current review therefore aims to give a broad overview of the different evaluation metrics for task-oriented textual dialogue systems that correspond to the characteristics of customer service chatbots. This paper serves two goals. First, we will show the vast amount of different constructs and operationalisations in a way that readers can use this paper as a reference for chatbot evaluation in the context of customer service. Second, the paper ends with a research agenda that aims to stimulate follow-up research on the evaluation of task-based dialogue systems in this specific usage context, and to generate mutual agreement on the different constructs, definitions and methods used for chatbot evaluation in general. 

\subsection{Reading guide}
\label{structure}
As there are many different measures and constructs present in current literature, the first part of this review aims to create a logical overview which can be read in full or quickly skimmed for relevant parts. Given the specific usage context of dialogue systems in the context of customer service, the results are divided in two main sections. The first broad division of the constructs focuses on constructs for intrinsic evaluation (\S\ref{subsec:intrinsic}), and the second by constructs focusing on the system in context (\S\ref{subsec:context}). Both main sections are divided in three subsections based on the constructs' topic and purpose:

\begin{enumerate}
    \item Intrinsic measures \hfill \S\ref{subsec:intrinsic}
    \begin{enumerate}
        \item Understanding the user: Natural Language Understanding (NLU) \hfill \S\ref{subsub:nlu}
        \item Evaluating chatbot utterances: Natural Language Generation (NLG) \hfill \S\ref{subsub:nlg}
        \item Performance and efficiency \hfill \S\ref{subsub:perf}
    \end{enumerate}
    \item System in context \hfill \S\ref{subsec:context}
    \begin{enumerate}
        \item Task success \hfill \S\ref{subsub:success}
        \item Usability \hfill \S\ref{subsub:usab}
        \item User experience \hfill \S\ref{subsub:exp}
    \end{enumerate}
\end{enumerate}

Within these subsections, the measures that will be discussed range from automatic metrics to the ones having humans involved. Given the extensive range of constructs that appeared in our review, we will focus on constructs that are particularly noteworthy and important in the context of customer service interactions. This categorisation makes it possible for the reader to focus on the constructs of interest. Recent developments are discussed in Section \ref{sec:recentdevelopments}. This section concerns the usage of LLMs both to power dialogue systems and to use as evaluation metric for NLP systems. Finally, in the discussion section (\S\ref{sec:discussion}) of this paper, the findings will be synthesised and the most outstanding outcomes and observations will be discussed. The paper will end with a research agenda that provides directions for future research on the evaluation of customer service chatbots in particular and recommendations for evaluating dialogue systems (\S\ref{sec:discussion} and \S\ref{sec:conclusion}).

\section{Method}
\label{method}

\subsection{Databases and search queries}
\label{dbqueries}
The first round of literature selection concerned the selection of databases that would be used for finding the literature. Four databases were chosen that contain papers from the more technical (NLP related) fields (ACM\footnote{https://dl.acm.org/search/advanced}, ACL anthology\footnote{https://aclanthology.org/}, IEEE\footnote{https://ieeexplore.ieee.org/search/advanced} and Web of Science\footnote{https://www.webofscience.com/wos/woscc/advanced-search}). The search needed to be as comprehensive as possible so no time periods were specified and the default setting of the respective database-search engines were used. To make sure papers were included that mainly focused on dialogue systems and involved some kind of evaluation, only the title and abstract were searched (not the full text). In all databases the following search query was used:\\
\\ \indent \textit{(chatbot* OR `dialogue system*’ OR `dialog system*’) 
\\ \indent AND 
\\ \indent (eval* OR analy* OR perf* OR perc*)}\\

The query ensured that plurals and different spellings of the words were selected. This means for example that papers containing the keywords `chatbots' and `evaluating' or `analysing' were selected but also articles on `dialogue systems' and `performing' or `perceptions'. 

For ACM the full text collection was searched including all dates. In IEEE, the default settings were used without an added timeframe. In Web of Science all editions of the Web of Science collection were searched, with again no added timeframe. The searches were done on the eighth and ninth of December 2021. For ACL, the ACL anthology BibTeX download including abstracts (08-12-2021) was used. In total 3,800 papers were found using this search strategy. From the results, duplicate entries were removed based on the title and DOIs. If the title and DOI were the same, only one entry was kept. If there were doubts (e.g. title is the same but DOI was different) these were marked and evaluated manually. This meant that eventually 3,458 records were kept for the first round of manual selection. Code and data for screening of the duplicates and the further selection process can be found on GitHub\footnote{https://github.com/Anouck96/LiteratureSurvey}.

\subsection{Paper selection}
\label{selection}
\begin{figure}[ht]
\centering
\includegraphics[width=0.8\textwidth]{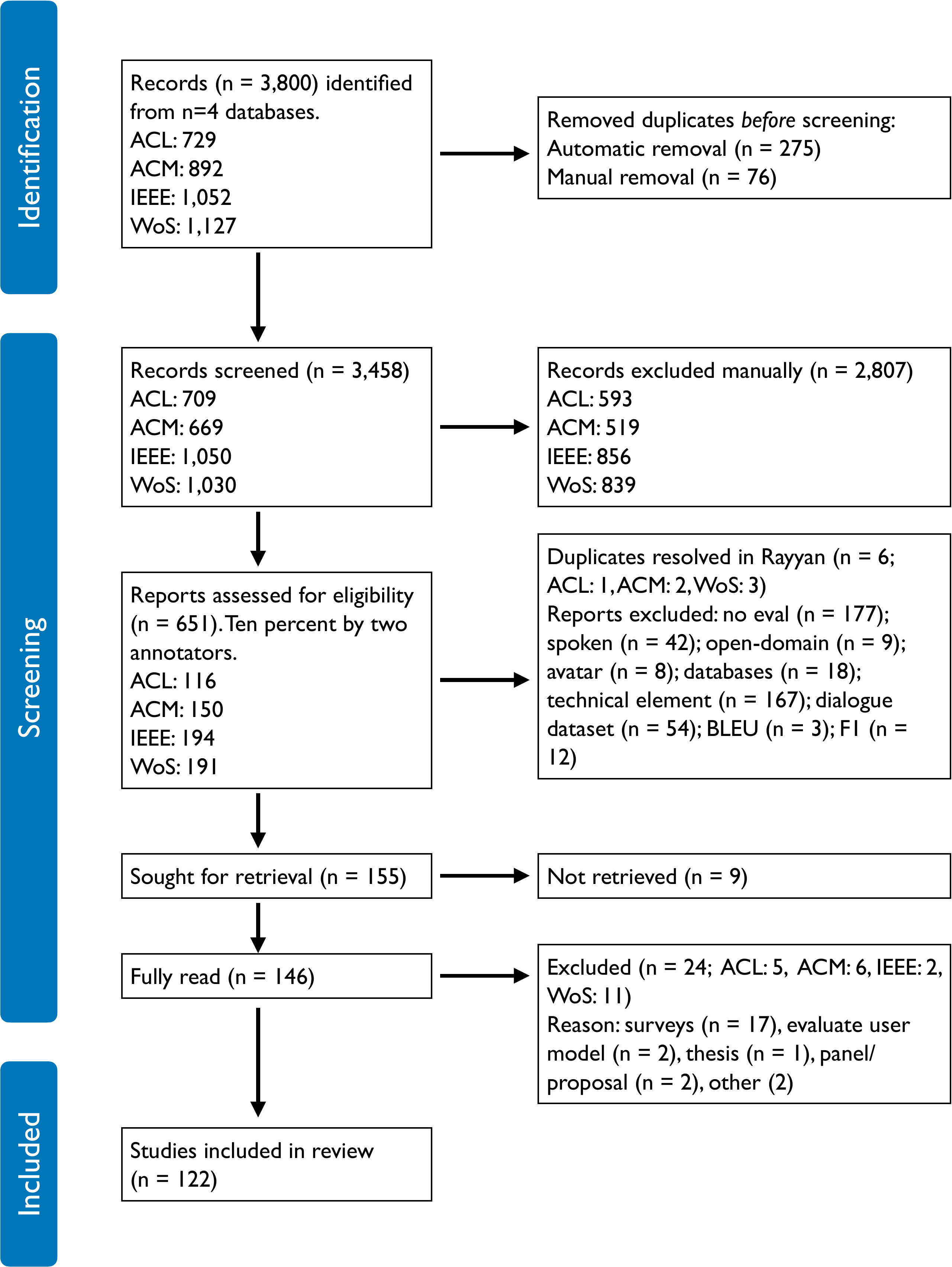}
\caption{PRISMA figure showing the selection process.}
\label{png:prisma}
\end{figure}

We used the PRISMA approach (Figure \ref{png:prisma}) to filter out irrelevant papers and to obtain a manageable subset for further analysis. We considered a paper to be relevant when there is some sort of (reflection on) evaluation of a task-oriented dialogue system.

A first quick selection consisted of screening the title and abstract by the first author. Based on key-words papers were either retained or discarded. In case of doubt, the paper was retained in this stage of the procedure to make sure no papers were missed. Papers that included key-phrases like spoken, interface, open-domain, emotion and annotation were thought to be about different sort of systems (such as social agents) or focus on different aspects of a dialogue system (such as the interface), and were discarded. If the papers mentioned either written or evaluation, they were retained. Of the 3,458 papers screened, 2,807 papers were manually removed.

To finalise the selection of papers relevant for the aim of this review, we used Rayyan \cite[a collaborative online platform for carrying out systematic reviews]{ouzzani2016rayyan}, to create and manage annotations. Using Rayyan, an extra round of duplicate removal was done, resulting in 645 papers for the next selection phase. By means of a flowchart, ten percent of the 645 papers were annotated by two annotators. The formulated criteria were used (such as `spoken'), and supplemented with other inclusion and exclusion terms that appeared during this phase (see the flowchart in Appendix ~\ref{app:flowchart}). For example, we encountered papers that focus on a virtual avatar; these papers were excluded from our data set as well. In addition, to narrow the selection further, papers that only use BLEU or F-scores for evaluation were also removed since these papers very narrowly discuss the evaluation process. As will be evident from the results section (\S\ref{sec:results}) of our review, these metrics are still also present in our final selection since these metrics were also often used with other accompanying metrics in previous papers, or they were accompanied by a broader discussion of the evaluation method in these papers.

Before annotating ten percent of the 645 papers by two annotators, ten papers were used for training the second annotator (an independent researcher who works with chatbots in a different domain and who has less knowledge about technical aspects of evaluation). After training and discussion, the flowchart was accordingly updated and ten percent (i.e., 65 papers) of the sample was annotated by both annotators. Yet another paper was found to be duplicate, resulting in 64 double annotated papers. Papers were annotated based on a full read of the title and abstract. Out of 64 papers there were 13 disagreements on discard/retain. This resulted in a Kappa statistic of 0.491, indicating moderate agreement. The 13 disagreements were easily resolved, and the decision tree was updated to clarify a few remaining ambiguities. The remaining papers were assessed for eligibility using the decision tree. Another 155 were kept, out of which nine could not be retrieved.

\begin{figure}
    \centering
    \includegraphics[width=1\textwidth]{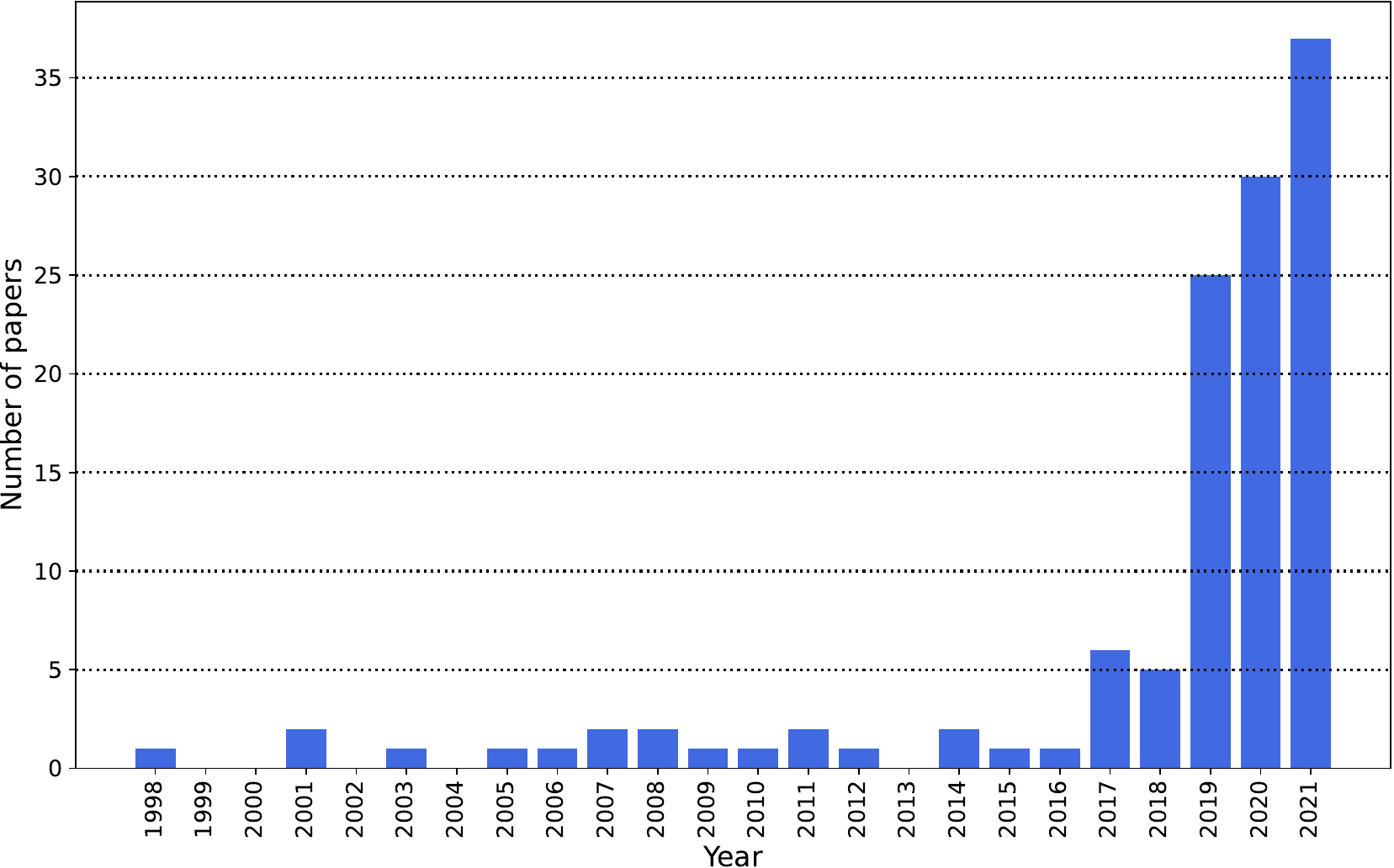}
    \caption{Bar graph showing the occurrence of papers within each year.}
    \label{fig:years}
\end{figure}

Out of the resulting 146 papers, 17 papers turned out to be reviews of sorts themselves. These were excluded from the study as they themselves sum up metrics (which might result in double counts of constructs in our results). These reviews are briefly described in Section \ref{subsec:previouswork}. Another two papers were excluded because they focused specifically on evaluation of the user model instead of a dialogue system itself. One addition was a thesis, which was excluded because there was a similar paper of the author included. Two reports were excluded because they were a panel summary and a proposal. Another two were excluded because they were either unclear about what is being measured or how it should be measured. Subsequently, a total of 122 papers were included in the current review. All of these papers were included, even if they turned out to not completely focus on task-oriented textual systems. As can be seen in Figure~\ref{fig:years} most papers were relatively recently published.

\subsection{Data Extraction sheet}
\label{dataextract}
A data extraction sheet was made to systematically record relevant information form each paper. As a first step this sheet was piloted on ten of the included studies to make sure all important and interesting information was included. The data extraction sheet was divided into general information and study specific information. Table~\ref{table:extract} shows the information that was recorded.

\begin{table}[ht!]
\small
\centering
\begin{tabular}{p{4cm}p{5cm}p{5cm}}
\toprule
{\textbf{Bibliographical}\newline Title \newline Author(s) \newline Year \newline Journal \newline Language} &
{\textbf{Measurement data} \newline Metric \newline Construct \newline Evaluator} &
{\textbf{System details}\newline Type of system \newline Goal/purpose of system \newline Language \newline Implementation of system}\\
 \\\midrule
{\textbf{Context}\newline Domain or industry \newline Goal of paper} &
{\textbf{Evaluation details} \newline Data set used \newline Sample size \newline Turn or conversation level \newline Moment of evaluation \newline Intrinsic/extrinsic} &
{\textbf{Additional information}\newline Critical analysis \newline Statistics on evaluation \newline Qualitative analysis \newline Reflection on difficulty \newline Other comments}\\
\bottomrule
\end{tabular}
\caption{Information extracted from papers.}
\label{table:extract}
\end{table}

Bibliographical en contextual information of the papers was first recorded. This included information like the title but also the domain and goal of the papers. Next, information on evaluation and measurement was recorded. Metrics were documented together with information on if they were specific for dialogue system evaluation and if they needed a reference for comparison. Then the construct and evaluators were noted down. Either human or automatic evaluation was used. If human evaluation was used it was recorded who was the one evaluating (authors, experts, participants, users). Finally, some system details and additional information was documented. It was noted down if the authors provided a critical analysis of the evaluation method, if they included statistics on the evaluation and if there was any qualitative analysis on the evaluation outcomes.

\begin{table}[hbt!]
\small
\centering
\begin{tabular}{p{3cm}p{3cm}p{8cm}}
\toprule
\textbf{Perspective} &
\parbox[t][][t]{3cm}{\textbf{Overarching\\category}} &
\textbf{{Enclosed constructs}} \\\midrule
{Intrinsic evaluation} &
NLU
& {Context-capturing, understanding}\\\cmidrule{2-3}
{~} & {NLG} &
{Adequacy, appropriateness, authenticity, clarity, coherence, completeness, conciseness, consistency, correctness, diversity, elicitation abilities, fluency, grammaticality, hate speech rate, informativeness, meaningfulness, naturalness, novelty, politeness, quality, quantity, readability, reasonableness, relatedness, relevance, repetition, sensibleness, simplicity, specificity, suitability, tediousness}
{~} \\\cmidrule{2-3}
{~} & \parbox[t][][t]{3cm}{Performance\\/efficiency} &
{Anomalies, benefits and risks, compatibility, costs, (task/interaction) complexity, content evaluation, efficiency, functionality, implementation, overall evaluation, performance, recommendation quality, response selection, robustness, similarity}
{~} \\\midrule
{System in context} &
\parbox[t][][t]{3cm}{(Task) success\\/effectiveness} &
{Comparing treatments, competence, effectiveness, errors, feasibility, health outcomes, intellectual, intelligence, interpretability, knowledge of material, learning outcomes, (customer) loyalty, persuasion outcomes, success} \\\cmidrule{2-3}
{~} & {Usability} &
{Accessibility, ease of use, effort/convenience, friendliness, intention to use, interactivity, learnability, responsiveness, usability} \\\cmidrule{2-3}
{~} & {User
experience} & {Acceptability, anxiety, assurance, attitude, attitude to improve health, autonomy, challenges, confirmation, cooperativeness, corporate reputation, (self-)efficacy, engagement, entertainment, emotional connection, empathy, experience, familiarity, fidelity, enjoyment, helpfulness, independence, indistinguishability, likability, motivation, patronage intentions, personality, realism/humanness, reliability, resistance, satisfaction, sentiment, social presence and influence, trust, understandable, usefulness, valuable, willingness}\\
\bottomrule
\end{tabular}
\caption{All 108 constructs divided into six subcategories.}
\label{table:1groups}
\end{table}

We tried to fill in the sheet as much as possible but in some cases this was not always feasible as some papers were quite vague and did not always mention all (sometimes important) details. This complicated our comparison of the different papers, and also makes it very challenging to reproduce the experiments. This is a well-known issue often discussed within the field of NLG and NLP (see for example \shortciteR{howcroft-etal-2020-twenty,belz-etal-2023-missing}).

\subsection{Data synthesis and grouping constructs}
\label{datasynt} 
To synthesise the data, similar constructs were grouped together with the respective methods used for measuring these constructs.
Table~\ref{table:1groups} presents all the constructs, split into two groups: intrinsic evaluation of the system (focusing on the system itself), and evaluation of the system in a context (involving also external elements such as the human user; see Figure \ref{fig:dsinteraction}). Each of these is then split into three sub-categories, based on existing work (the surveys discussed in \S\ref{subsec:previouswork}) and - if this did not suffice - our own intuition. For \emph{intrinsic evaluation}, we made a distinction between constructs that concern understanding, generating, and performance. Within the \emph{system in context} category, we made a distinction between usability, user experience, and task success.

\subsection{Additional papers}
Since carrying out a systematic review at this scale is time-consuming, there are always going to be new publications by the time you are finally done. Therefore, following our initial survey and data synthesis, we manually looked through the literature after 2021 to ensure that recent developments (such as the introduction of ChatGPT; \citeR{chatgpt}) could also be taken into account. We will reflect on these developments at the end of our results section (\S\ref{sec:recentdevelopments}).

\section{Results}
\label{sec:results}
We will now turn to a discussion of the constructs identified in Table~\ref{table:1groups}. This table shows that there is an unequal distribution of constructs among the categories. Especially for NLU and usability there are considerably fewer identified constructs. The results will start with a discussion of intrinsic evaluation constructs (\S\ref{subsec:intrinsic}). These paragraphs are divided into sub-paragraphs corresponding to the overarching categories (NLU, NLG, performance/efficiency). For each category, we present an overview table indicating which papers are associated with each of the constructs that are enclosed in that category. For the sake of illustration, a maximum of three constructs will be discussed in detail for every category. Our aim here is not to be exhaustive and discuss each of the 108 constructs in detail, since this would result in an overly long and repetitive survey. Rather, we intend to show how one might reason about different ways to operationalise a specific construct. For each construct, we ask the following questions:
\begin{enumerate}
    \item How is the construct defined?
    \item How is the construct operationalised?
    \begin{enumerate}
        \item Is there an automatic evaluation procedure?
        \item Is there a human/manual evaluation procedure?
    \end{enumerate}
    \item What is the intuition behind these operationalisations? How do they relate to the construct at hand?
\end{enumerate}

Readers interested in other constructs are invited to read the papers associated with those constructs, and to follow the same approach. Section~\ref{subsec:context} discusses ways to evaluate dialogue systems in context. This section follows the lead of Section~\ref{subsec:intrinsic}, highlighting a selection of relevant constructs.

\subsection{Intrinsic evaluation}
\label{subsec:intrinsic}

\subsubsection{Understanding the user: Natural Language Understanding (NLU)}
\label{subsub:nlu}

Understanding the users' utterances is one of the key factors for a well-functioning dialogue system. After all, if the system fails at this part, the utterance is likely to be misunderstood and the chances are high that the user will not be content with the response and hence the overall system. As can be seen in Table~\ref{table:1groups} above, we could only associate two constructs with NLU. Our analysis yielded only 10 papers (out of a total of 122) addressing one or more of these constructs (see Table~\ref{table:nlu-constructs}). With just ten papers mentioning constructs related to NLU, it seems fair to say that less attention is devoted to this category than to the other categories related to intrinsic evaluation.

{\small
\begin{table}[htp!]
\centering
\begin{tabular}{p{3cm}p{6cm}p{5cm}}
\toprule
\textbf{{Construct}} &
\textbf{{Approach}} &
\textbf{{Papers}} \\
\midrule
Context-capturing &
Rating &
{\shortciteA{duggenpudi-etal-2019-samvaadhana,nazir2019novel}}\\\midrule
{\parbox{3cm}{Understanding}}
& {Slot error rates} &
{\shortciteA{aust1998evaluating}} \\\cmidrule{2-3}
& {Classification accuracy, macro-averaged f-score} &
{\shortciteA{Dzikovska-etal-2012-evaluating}} \\ \cmidrule{2-3}
& {User simulation} & {\shortciteA{lopez2003assessment}}\\\cmidrule{2-3}
& {Engagement duration, response length, response informativeness, response quality index} & {\shortciteA{xiao2020if}}\\\cmidrule{2-3}
& {Conversation logs} & {\shortciteA{Dzikovska-etal-2012-evaluating}}\\\cmidrule{2-3}
& {Survey/rating} &
{\shortciteA{campillos2021lessons,campillos2020designing,ham-etal-2020-end,takanobu-etal-2020-goal,xiao2020if}} \\
\bottomrule
\end{tabular}
\caption{Constructs and metrics for measuring NLU (two constructs; ten distinct papers)}
\label{table:nlu-constructs}
\end{table}}

\paragraph{Context-capturing} focuses on how well the system is able to capture the (dialogue) history and context. The system should be able to integrate knowledge from previous turns in the conversation in the current turn. If the system has sufficient information, from both external sources as well as the dialogue history itself, it should be able to properly understand and respond to the user. Context-capturing can be regarded as an important construct in the context of dialogue system research as it is sometimes seen as the differentiating factor between a question answering (Q-A) system and a dialogue system \shortcite{duggenpudi-etal-2019-samvaadhana}, as a pure Q-A system often does not take context and history into account (and is therefore arguably not conversational). Important though it may be, we found only two papers that measure this construct.

\textit{Automatic evaluation} has not been used measuring context-capturing, although automatic evaluation seems to be possible. A similar construct, relatedness, is for example measured by BLEU. The difference though is that relatedness only focuses on the previous turn and not on the complete conversational context. As metrics like BLEU need a reference this might not be the perfect solution for complete dialogue contexts.

\textit{Human evaluation}, on the other hand, is done by both \shortciteA{duggenpudi-etal-2019-samvaadhana} and \shortciteA{nazir2019novel}. In both studies, participants were asked to rate the complete dialogue system on a scale. \shortciteA{nazir2019novel} have three experts rate the systems
after a conversation of ten minutes. The criterion for context-capturing is phrased as follows: `How many results are history-based and give results in the same context?' \shortcite{nazir2019novel}. \shortciteA{duggenpudi-etal-2019-samvaadhana} asked eight participants to rate the system on a 6-point scale for 20 to 30 dialogues. They use the following description for context-capturing: `How well is the context captured in a full-fledge[d] dialogue conversations?'\shortcite{duggenpudi-etal-2019-samvaadhana}. 

These two descriptions are still quite vague as concepts like context are relatively broad. The type of system and focus of the papers give a bit more direction. Where \shortciteA{nazir2019novel} focus on the information retrieval part of a (fashion brand) chatbot, \shortciteA{duggenpudi-etal-2019-samvaadhana} focus on the question answering component of a hospital chatbot. The goal and function of the chatbot seems to be an important factor in considering context-capturing. Both papers focus on retrieving correct information, meaning that the chatbot should return information that is consistent with previously provided information. Although context-capturing is described as an important and distinguishing element of dialogue systems, only human behaviour is used for evaluation. This is all indirect information and the context-capturing itself is actually not measured.

\paragraph{Understanding} measures the system's ability to understand the natural language of the user input. Compared to the definition of context-capturing which focuses on a conversation level, here we focus on measuring on a sentence level. Multiple terms are used for this construct, such as interpretation (used by \shortciteR{Dzikovska-etal-2012-evaluating}), and comprehension (used by \shortciteR{xiao2020if}). For all practical purposes, these terms are grouped together as understanding. The subtle differences in definitions of understanding, interpretation and comprehension also show that it is in some cases difficult to properly define specific constructs. These terms are sometimes also used interchangeably, making it even more difficult to determine the definition of the construct. 

The literature distinguishes two forms of understanding. The first one is about users who are able to understand the system utterances (referred to as `user-understanding' by \shortciteR{campillos2020designing,campillos2021lessons}). The second one is about the system being able to understand the user input (`system understanding,' \shortciteR{campillos2020designing,campillos2021lessons}). In this section we focus on the latter definition. 

\textit{Automatic evaluation} can be applied by using the slot error rate. The slot error rate is a measure that is similar to the word error rate often used in speech recognition. The word error rate compares a reference to the identified sentence, showing how well the system (or speech recognizer) is able to recognize a spoken sentence \shortcite{jurafsky2021}.
\shortciteA{aust1998evaluating} regard the meaning of a sentence as slot-filler pairs, creating the possibility to define an error measure: the slot error rate. To calculate the slot error rate a reference is needed (the meaning of a sentence) to compare to \cite{aust1998evaluating}. Since \citeA{aust1998evaluating} evaluate a speech system, it is important to keep in mind that the slot error rate also relates to the performance of the speech recogniser, which is not always relevant for text-based systems.

Logs are used to create a gold standard by \shortciteA{Dzikovska-etal-2012-evaluating} for automatic evaluation of understanding. In this paper they evaluate the interpretation component of a system teaching students about electricity and electronics with the eventual goal of discovering the extent to which the system contributes to the final learning outcomes. The student answers to the system were manually annotated in the conversation logs. The scholars employ a simple annotation scheme where an utterance can be correct, partially correct but incomplete, irrelevant or contradictory \shortcite{Dzikovska-etal-2012-evaluating}. On the basis of the annotations they create the gold standard that is used for the automatic evaluation. With the automatic evaluation they aim to quantify the extent to which the system is able to correctly classify the student answer. \shortciteA{Dzikovska-etal-2012-evaluating} evaluated the NLU module of the system by comparing the system output to the manual labels given to the student answers. They use both accuracy (percentage of correct predictions) and macro-averaged f-scores (combination of precision and recall, disregarding the class size). The approach taken by \shortciteA{Dzikovska-etal-2012-evaluating} reduces measuring interpretation to a classification task based on the gold standard manual annotations, which might not be the most effective approach to gain access to the internal representations of the understanding-module.

\textit{Human evaluation} is carried out by means of conducting a survey or by having participants rate utterances. \shortciteA{takanobu-etal-2020-goal} for example have 100 participants interact with multiple systems and have them rate each system on a 5-point scale for both language understanding as well as response appropriateness. Evaluators thus rate the reactions the system provides in response to the user utterance, not the internal representation that the dialogue system has created. Although, humans are not able to get a full grasp of this internal representations, still half of the papers only employ human evaluation.

\subsubsection{Evaluating chatbot utterances: Natural Language Generation (NLG)}
\label{subsub:nlg}

Table \ref{tab:nlgmetrics} shows the different constructs that are associated with different aspects of the text that are communicated by the system. These constructs are typically operationalised using metrics that are also used in the field of Natural Language Generation (see \shortciteR{celikyilmaz2020evaluation}, for an extensive overview). 
Inspection of Table \ref{tab:nlgmetrics} reveals a wide range of constructs varying from authenticity to sensibleness. Some constructs are just measured in one particular way (such as clarity, only by surveys) and in one particular study (such as simplicity, only in \shortciteR{barreto2021development}), while other constructs are measured in a wide variety of ways (such as fluency).

In contrast to Table \ref{table:nlu-constructs} (NLU), this table about NLG is far more elaborate, containing a greater number of constructs and a larger variety of papers. Although we focus on task-based systems, that are often still rule-based, there is a lot of attention to evaluating the chatbot utterances. This might be somewhat surprising, as these systems are often not generative in nature but use predefined responses. It can be argued that evaluating predefined responses is still worthwhile, for example on aspects such as coherence, politeness and simplicity.

{\small
\begin{longtable}[]{p{3cm}p{5cm}p{6cm}}
\toprule
\textbf{{Construct}} &
\textbf{{Approach}} &
\textbf{{Papers}} \\
\midrule
\endhead
\hline\\
\caption{Constructs and metrics for measuring NLG (continued on next page)}
\label{tab:nlgmetrics}\endfirstfoot
\hline\\
\caption*{Table~\ref{tab:nlgmetrics}: Constructs and metrics for measuring NLG (continued on next page)}\endfoot
\bottomrule\\
\caption*{Table~\ref{tab:nlgmetrics}: Constructs and metrics for measuring NLG (31 constructs; 64 distinct papers)}
\endlastfoot

{Adequacy} & {AM-FM} &
{\shortciteA{D2019automatic}} \\\cmidrule{2-3}
& {Rating} & {\shortciteA{burtsev2018first}}\\
\midrule
Appropriateness &
{Accuracy of intent classifier} &
{\shortciteA{vasconcelos2017bottester}} \\\cmidrule{2-3}
& {Generative models} & {\shortciteA{wu-chien-2020-learning}}\\\cmidrule{2-3}
& {Survey/rating} &
{\shortciteR{burtsev2018first,Eric-etal-2017-key,ham-etal-2020-end,takanobu-etal-2020-goal}} \\\cmidrule{2-3}
& {Pairwise (A/B) comparison} & {\shortciteA{wu-chien-2020-learning}}\\
\midrule
{Authenticity} &
{Survey} &
{\shortciteA{rese2020chatbots}} \\
\midrule
{Clarity} &
{Survey/rating} &
{\shortciteA{barreto2021development,jwalapuram2017evaluating,nazir2019novel}} \\
\midrule
Coherence &
{QuantiDCE} &
{\shortciteA{ye-etal-2021-towards-quantifiable}} \\\cmidrule{2-3}
& {Logs} &
{\shortciteA{campillos2020designing,liu2015ergonomics}}\\\cmidrule{2-3}
& {Survey/rating} &
{\shortciteA{campillos2021lessons,campillos2020designing,gandhe-traum-2008-evaluation,santhanam-shaikh-2019-towards,song2019task}} \\\cmidrule{2-3}
& {Ordering} &
{\shortciteA{gandhe-traum-2008-evaluation}} \\
\midrule
{Completeness} &
{Survey/rating} &
{\shortciteA{Bansal2021neural,rese2020chatbots}} \\\midrule
{{Conciseness}} &
{Size in characters/words} &
{\shortciteA{vasconcelos2017bottester}} \\\cmidrule{2-3}
& {Survey/rating} &
{\shortciteA{Bansal2021neural,campillos2021lessons,campillos2020designing,crutzen2011artificially}} \\\midrule
{Consistency} &
{Survey/rating} &
{\shortciteA{Shi-etal-2021-refine-imitate}} \\\midrule
{{Correctness}} &
{Confidence levels} &
{\shortciteA{bunga2019developing}} \\\cmidrule{2-3}
& {Hidden layers/number of neurons/ dropout
rate/dialogue testing} &
{\shortciteA{bunga2019developing}} \\\cmidrule{2-3}
& {Performance testing} &
{\shortciteA{bhawiyuga2017design}} \\\cmidrule{2-3}
& {Matching/learning curves} &
{\shortciteA{hwang2019end}} \\\cmidrule{2-3}
& {BLEU} &
{\shortciteA{Eric-etal-2017-key,liu2020cbet}} \\\cmidrule{2-3}
& {Accuracy/semantic frame accuracy/joint goal
accuracy} & {\shortciteA{mi-etal-2021-self,peng-etal-2021-raddle,song2019task,su-etal-2020-moviechats,xiao2020if,xu2020healthcare}} \\\cmidrule{2-3}
& {Precision, recall, (entity) F1} &
{\shortciteA{Eric-etal-2017-key,xiao2020if,xu2020healthcare}}\\\cmidrule{2-3}
& {DialTest} &
{\shortciteA{liu2021dialtest}} \\\cmidrule{2-3}
& {Paradise} & {\shortciteA{bickmore2006health}} \\\cmidrule{2-3}
& {User simulation} & {\shortciteA{lopez2003assessment}} \\\cmidrule{2-3}
& {Logs} &
{\shortciteA{campillos2021lessons,campillos2020designing}} \\\cmidrule{2-3}
& {Survey/rating} &
{\shortciteA{Bansal2021neural,duggenpudi-etal-2019-samvaadhana,Eric-etal-2017-key,nazir2019novel,Okonkwo2020python,su-etal-2020-moviechats}} \\\cmidrule{2-3}
& {Ranking} &
{\shortciteA{nuruzzaman2020intellibot}} \\\midrule
{{Diversity}} &
{Distinct-1/Distinct-2} &
{\shortciteA{Firdaus2020more}} \\\cmidrule{2-3}
& {Survey/rating} &
{\shortciteA{Firdaus2020more}} \\\midrule
{Elicitation abilities} &
{Logs} &
{\shortciteA{han2021designing}} \\\midrule
{{Fluency}} &
{AM-FM} &
{\shortciteA{D2019automatic}} \\\cmidrule{2-3}
& {Perplexity} &
{\shortciteA{Firdaus2020more}} \\\cmidrule{2-3}
& {BLEU} &
{\shortciteA{peng-etal-2021-raddle}} \\\cmidrule{2-3}
& {Logs} &
{\shortciteA{liu2015ergonomics}} \\\cmidrule{2-3}
& {Survey/rating} &
{\shortciteA{Eric-etal-2017-key,Firdaus2020more,Shi-etal-2021-refine-imitate,zhao2019evaluation}} \\\cmidrule{2-3}
& {Ranking} &
{\shortciteA{deriu-etal-2020-spot}} \\\midrule
{Grammaticality} &
{Perplexity} &
{\shortciteA{Firdaus2020more}} \\\midrule
{Hate speech rate} &
{Logs} &
{\shortciteA{han2021designing}} \\\midrule
{{Informativeness}} &
{Logs} &
{\shortciteA{han2021designing,weisz2019bigbluebot}} \\\cmidrule{2-3}
& {Survey/rating} &
{\shortciteA{campillos2021lessons,campillos2020designing,ihsani2021conversational,orden2021analysis,su-etal-2020-moviechats,zhang2021informing}} \\\midrule
{{Meaningfulness}} &
{Survey/rating} &
\shortciteA{ccetinkaya2020developing} \\\cmidrule{2-3}
& {Ranking} &
{\shortciteA{banchs2014empirical}} \\\midrule
{{Naturalness}} &
{Survey/rating} &
{\shortciteA{campillos2021lessons,campillos2020designing,kadariya2019kbot,okanovic2020can,song2019task}} \\\cmidrule{2-3}
& {Interview} &
{\shortciteA{atiyah2019evaluation}} \\\cmidrule{2-3}
& {Cognitive walkthrough} &
{\shortciteA{atiyah2019evaluation}} \\\midrule
{Novelty} &
{Survey/rating} &
{\shortciteA{chen2021_usability,rese2020chatbots}} \\\midrule
{{Politeness}} &
{Politeness accuracy} &
{\shortciteA{Firdaus2020more}} \\\cmidrule{2-3}
& {Survey/rating} &
{\shortciteA{Firdaus2020more}} \\\midrule
{{Quality}} &
{Utterance level quality prediction} &
{\shortciteA{Finch-etal-2021-went}} \\\cmidrule{2-3}
& {BLEU} &
{\shortciteA{yin2017deepprobe}} \\\cmidrule{2-3}
& {Perplexity} &
{\shortciteA{Shi-etal-2021-refine-imitate}} \\\cmidrule{2-3}
& {Response time, response length, response
informativeness, response quality index} &
\shortciteA{vasconcelos2017bottester,xiao2020if} \\\cmidrule{2-3}
& {Engagement duration} &
{\shortciteA{xiao2020if}} \\\cmidrule{2-3}
& {Logs} &
{\shortciteA{Foster-etal-2009-comparing}} \\\cmidrule{2-3}
& {A/B testing} & {\shortciteA{sedoc-ungar-2020-item}} \\\cmidrule{2-3}
& {Survey/rating} &
\shortciteA{abu-shawar-atwell-2007-different,barreto2021development,burtsev2018first,ccetinkaya2020developing,crutzen2011artificially,Finch-etal-2021-went,Foster-etal-2009-comparing,Gonzales2017bots,jwalapuram2017evaluating,kadariya2019kbot,sensuse2019chatbot,xiao2020if} \\\cmidrule{2-3}
& {Interview} &
{\shortciteA{Gonzales2017bots}} \\\cmidrule{2-3}
& {Direct observation} &
{\shortciteA{Gonzales2017bots}} \\\cmidrule{2-3}
& {Analysis} &
{\shortciteA{Gonzales2017bots}} \\\midrule
{{Quantity}} &
{Relative utterance quantity} &
{\shortciteA{Khayrallah-sedoc-2021-measuring}} \\\cmidrule{2-3}
& {Survey} &
{\shortciteA{crutzen2011artificially,jwalapuram2017evaluating}} \\\midrule
{Readability} &
{Survey/rating} &
{\shortciteA{santhanam-shaikh-2019-towards}} \\\midrule
{Reasonableness} &
{Survey/rating} &
{\shortciteA{abu-shawar-atwell-2007-different,nazir2019novel}} \\\midrule
{{Relatedness}} &
{Kn-bleu} &
{\shortciteA{ccetinkaya2020developing}} \\\cmidrule{2-3}
& {Survey/rating} &
{\shortciteA{abu-shawar-atwell-2007-different,ccetinkaya2020developing,jimenez2021find,liu2020cbet,nazir2019novel}} \\\midrule
{{Relevance}} &
{Kn-BLEU} &
{\shortciteA{ccetinkaya2020developing}} \\\cmidrule{2-3}
& {Accuracy of intent classifier} &
{\shortciteA{vasconcelos2017bottester}} \\\cmidrule{2-3}
& {Survey/rating} &
{\shortciteA{Bansal2021neural,duggenpudi-etal-2019-samvaadhana,Firdaus2020more,jwalapuram2017evaluating,oniani2020qualitative,van2019chatbot}}\\\midrule
{{Repetition}} &
{Logs} &
{\shortciteA{han2021designing}} \\\cmidrule{2-3}
& {Survey/rating} &
{\shortciteA{Shi-etal-2021-refine-imitate}} \\\midrule
{{Sensibleness}} &
{Rating} &
{\shortciteA{su-etal-2020-moviechats}} \\\cmidrule{2-3}
& {Survival analysis} &
{\shortciteA{deriu-etal-2020-spot}} \\\midrule
{Simplicity} &
{Survey/rating} &
{\shortciteA{barreto2021development}} \\\midrule
{Specificity} &
{Survival analysis} &
{\shortciteA{deriu-etal-2020-spot}} \\\midrule
{{Suitability}} &
{Classification accuracy} &
{\shortciteA{Lee2021sumbt+}} \\\cmidrule{2-3}
& {Success rate} &
{\shortciteA{Lee2021sumbt+}} \\\cmidrule{2-3}
& {Inform rate} &
{\shortciteA{Lee2021sumbt+}} \\\cmidrule{2-3}
& {Delexicalized-BLEU} &
{\shortciteA{Lee2021sumbt+}} \\\cmidrule{2-3}
& {Entity F1} &
{\shortciteA{Lee2021sumbt+}} \\\cmidrule{2-3}
& {Survey/rating} &
{\shortciteA{niculescu2020digimo}} \\\midrule
{Tediousness} &
{Survey/rating} &
{\shortciteA{campillos2021lessons,campillos2020designing}} \\
\end{longtable}}

Below we will discuss some of the constructs that show a substantial variety in approaches (like fluency) or constructs that might become more relevant with current developments in the field of dialogue systems (such as correctness). Constructs such as correctness are of great importance in the context of customer service as well, as companies strive to offer their customers only correct information.

\vspace{5mm}

\paragraph{Coherence} generally refers to how logical and easy to follow dialogue sequences are. The system needs to produce responses consistent with the topic of the conversation \shortcite{venkatesh2017evaluating,santhanam-shaikh-2019-towards}. The construct is sometimes also defined in terms of other constructs, as happens in \shortciteA{ye-etal-2021-towards-quantifiable}. In their definition coherence shows how fluent, consistent and context-related the utterances are (which are all separate constructs in the tables of this review). This shows how complex defining and setting boundaries for a construct can be.

\textit{Automatic evaluation} is only carried out by means of one method, namely QuantiDCE (introduced by \shortciteR{ye-etal-2021-towards-quantifiable}). QuantiDCE (Quantifiable dialogue Coherence Evaluation) aims to reflect human ratings. As human evaluation is often seen as the most accurate way of evaluating coherence, the often used proxies (such as word-overlap-metrics, e.g. BLEU) often do not align very well with human ratings \shortcite{ye-etal-2021-towards-quantifiable}. QuantiDCE is a machine learning model that uses BERT to encode response-pairs. It is trained to learn coherence degrees based on a limited amount of human annotated data. Coherence is rated on a scale instead of either coherent or incoherent \shortcite{ye-etal-2021-towards-quantifiable}. The authors compare QuantiDCE to human evaluation and to eight existing metrics, such as BLEU \shortcite{papineni2002bleu}, METEOR \shortcite{banerjee2005meteor} and ROUGE \shortcite{lin2004looking}. QuantiDCE shows a strong correlation to human evaluation compared to these other metrics \shortcite{ye-etal-2021-towards-quantifiable}. The weak correlation to human judgements has been a problem in machine translation as well, and multiple different approaches are suggested in this context, such as BLEUrt \shortcite{sellam-etal-2020-bleurt} and BERTScore \shortcite{bert-score}. As QuantiDCE is a machine learned metric, it is difficult to interpret which aspects of coherence are captured in the final score.

\emph{Human evaluation} is most often applied to measure this construct, for example via surveys or an ordering task. The method `ordering' only occurs with this construct, and was used by \shortciteA{gandhe-traum-2008-evaluation}. In contrast to ranking, which often focuses on ranking various mutations of one specific turn, ordering focuses on shuffling multiple different turns in one conversation. In their study, \shortciteA{gandhe-traum-2008-evaluation} apply the information ordering task (creating a sequence for informative elements) to a dialogue context and ask participants to reorder turns to make the final dialogue as coherent as possible. Not only do they use ordering, they also ask participants to assign a coherence rating to each `dialogue permutation' on a scale of 1 to 7 (1 being very incoherent, 7 being perfectly coherent) and each dialogue turn (scale 1-5). They do not provide any instructions on coherence to capture the intuitive idea the judges have of coherence. Next to surveys and ordering the dialogue logs are also examined with regards to coherence. \citeA{liu2015ergonomics} manually examine coherence by analysing the dialogue history of the users. They do not further discuss how this is exactly done or which definition of coherence they have used. \citeA{campillos2020designing} also examine the conversation logs and manually classify turn pairs as correct, incorrect or as a clarification request. A correct reply contains `a coherent answer with regard to the user question and correct information from the VP (Virtual Patient) record' \cite{campillos2020designing}.

\paragraph{Correctness} tends to focus on how accurate, correct and precise the utterances of a dialogue system are. It appeared that correctness can be interpreted in two ways. Some of the papers refer to the state of the given information. In this case it measures if the information returned by the dialogue system is factual and correct (for example in \citeR{campillos2021lessons,campillos2020designing}). Similarly, \citeA{su-etal-2020-moviechats} instead use the term `factuality' to refer to the veracity of the output. Other papers consider semantical correctness. A sentence can be semantically correct but offer too little information to be relevant \cite{nuruzzaman2020intellibot}. The question is of these definitions are compatible or if this actually measures two separate constructs. While the first two reflect on the status of the given information, the last one focuses solely on the semantic meaning of the sentence.

\textit{Automatic evaluation} is oftentimes used to measure correctness. For example, accuracy is used by several papers, in some cases together with precision (number of items among selected items that are correct), recall (number of correct items that are correctly selected) and F1 (harmonic mean of precision and recall). There are two special cases, namely semantic frame accuracy \cite{song2019task} and joint goal accuracy \cite{peng-etal-2021-raddle}. With semantic frame accuracy, \citeA{song2019task} aim to find the proportion of the sentences that have both correctly predicted slots and intents (they use a slot filling task and intent detection task in their NLU-module). \citeA{peng-etal-2021-raddle} use joint goal accuracy as a measure for correct identification of the user's search goal constraints (focusing on the, often implicit, user goals).

Because most automatic approaches are actually adopted from different fields (such as machine translation in the case of BLEU), there is also a need to create approaches specifically for dialogue systems. DialTest \cite{liu2021dialtest} is one of these specifically designed approaches and measures the accuracy of the intent recognition and robustness of the dialogue system. It focuses on RNN-based natural language understanding modules. DialTest generates similar test cases and is able to select data that might trigger errors in the model. The data set is then used to test the robustness of the model or for retraining \cite{liu2021dialtest}.

\textit{Human evaluation} was carried out by means of a survey, analysing the logs or by ranking. \citeA{campillos2021lessons} manually analyse the conversation logs to check if the information provided is correct compared to the information in the patient record. \citeA{duggenpudi-etal-2019-samvaadhana} combine correctness and relevancy by asking users to rate the system output using the question/description `How relevant/correct is the answer retrieved.' In contrast to combining it with relevancy, \citeA{Bansal2021neural} actually contrast accurate responses to relevancy: `The response is accurate, even if it is not relevant to my question.' A different approach is taken by \citeA{nuruzzaman2020intellibot}, who ask participants to rank utterances by different systems. In the context of human evaluation the definition of correctness often remains quite vague. \citeA{nazir2019novel} for example give the following definition: `How much accurate the results are.'

\paragraph{Fluency} tends to refer to the naturalness or native-likeness of utterances produced by a dialogue system. \citeA{D2019automatic} take the definition of fluency from machine translation and focus on the quality of the construction, emphasising syntactic validity. This construct appears in eight distinct papers and is measured by three different automatic metrics and also by means of human evaluation. 

\emph{Automatic evaluation} appeared by means of three automatic metrics, namely perplexity (used by \citeR{Firdaus2020more}), the BLEU score (used by \citeR{peng-etal-2021-raddle}) and AM-FM (used by \citeR{D2019automatic}).
With regard to the first automatic metric, fluent sentences should be grammatical, run smoothly and convey the intended message in a clear and structured way. Therefore perplexity is often used as a useful proxy for fluency, as fluent sentences in a particular language tend to be more similar to other utterances from that language, than sentences containing different kinds of errors. Perplexity in this context refers to the inverse probability of an utterance (computed using a language model), normalised by the number of words in that utterance \cite{jurafsky2021}. Intuitively, this corresponds to the surprisal of seeing a particular sequence of words, given data used to train the language model. Sequences of words that are unlike the training data receive a high perplexity score, while sequences of words that are very similar to the training data receive a low perplexity score. \citeA{Firdaus2020more} emphasise that as the perplexity score decreases, the responses become more fluent and grammatical.

The BLEU metric is a n-gram based textual similarity score \cite{papineni2002bleu}. This metric requires there to be a set of reference utterances to which an automatically generated utterance (the candidate) can be compared. Intuitively, BLEU looks at the overlap between the candidate and the reference utterances (in the case of \citeA{peng-etal-2021-raddle} the generated response is compared to a human-written response). This overlap is computed using the exact tokens in the sentence, meaning the BLEU score does not take synonyms into account, unlike alternative metrics such as METEOR. When used as a proxy for fluency, we might say that it captures the similarity between a generated utterance and other possible utterances in a particular context. BLEU's dependence on context-dependent reference data suggests a notion of `conversational fluency' which incorporates ideas about appropriateness of the generated response. This is different from what is captured by the perplexity metric, which only looks at the intrinsic fluency of an utterance in isolation, disregarding the context (which is to be captured by a different metric). One might say that this makes perplexity a `purer' measure of fluency, but one might also argue that the BLEU measure stretches the idea of fluency beyond recognition.

Overall the BLEU metric has extensively been used in Machine Translation and Natural Language Generation, because for any given input, there is often only a limited set of possible translations or other appropriate outputs that need to be taken into account. Comparing automatically generated outputs to expected outputs makes sense, because a greater similarity could be expected between output and reference data to correlate with higher quality output. However, the BLEU metric has been criticised for its lack of correlation with human judgements (see for example \citeR{reiter-2018-structured}). Moreover, in the context of dialogue, one might wonder whether BLEU is still the right choice, since there are many possible responses to a given input. The AM-FM metric discussed next, aims to omit the need for a reference and shows that better correlations can be reached than with BLEU.

The AM-FM metric is an adequacy-fluency metric that uses both syntactic (referencing to fluency) and semantic (referencing to adequacy) information on a sentence level \cite{D2019automatic}. It was first introduced by \citeA{banchs-li-2011-fm} for machine translation to omit the need for a reference translation and maintain consistency compared to human scores. \citeA{D2019automatic} focus in their paper on the evaluation of the metric itself. They show that the AM-FM framework shows better correlation to human scores compared to other metrics (such as BLEU, METEOR, ROUGE and CIDEr).
    
\emph{Human evaluation} often contain items where human judges are asked to rate the fluency of an utterance on a scale. \citeA{Firdaus2020more} for example ask participants to rate fluency by means of the following statement: `The generated response is grammatically correct and is free of any errors.' Additionally, there are two other ways of using human evaluation for measuring fluency. \citeA{deriu-etal-2020-spot} compare human responses to system responses and let annotators rank which of the two entities in a conversation is more fluent, while \citeA{liu2015ergonomics} analyse the complete dialogue history to determine fluency (but do not exactly mention how this is done).

\subsubsection{Performance and efficiency}
\label{subsub:perf}

Table \ref{tab:perf} shows constructs that can be related to the broader concept of performance of the dialogue system. It is noticeable, that there is a vast amount of different automatic approaches, many of them used to measure the constructs  efficiency and performance. 
 
{\small
\begin{longtable}[]{p{3.25cm}p{5cm}p{5.75cm}}
\toprule\noalign{}
\endhead
\hline\\
\caption{Constructs and metrics for measuring performance (continued on next page)}
\label{tab:perf}\endfirstfoot
\hline\\
\caption*{Table~\ref{tab:perf}: Constructs and metrics for measuring performance (continued on next page)}\endfoot
\bottomrule\\
\caption*{Table~\ref{tab:perf}: Constructs and metrics for measuring performance (16 constructs; 50 distinct papers)}
\endlastfoot
\textbf{Construct} &
\textbf{Approach} &
\textbf{Papers} \\\midrule
{Anomalies} &
{Cross-entropy loss acts as reconstruction loss
to detect anomalies} &
{\shortciteA{nedelchev2020treating}} \\\midrule
{{}{{Benefits and risks}}}
& {Conversation logs} &
{\shortciteA{han2021designing}} \\\cmidrule{2-3}
& {Survey} &
{\shortciteA{cheng2020how,chung2021chatbot,crutzen2011artificially,Jang2021mobile,pricilla2018designing,rese2020chatbots,sensuse2019chatbot,van2019chatbot}} \\\cmidrule{2-3}
& {Interview} &
{\shortciteA{pricilla2018designing,ren2020understanding}}\\\cmidrule{2-3}
& {Negative, positive feedback (emoji)} &
{\shortciteA{chung2021chatbot}} \\\cmidrule{2-3}
& {Focus group} &
{\shortciteA{schmidlen2019patient}} \\\midrule
{Compatibility} &
{Pipeline (match target system response)} &
{\shortciteA{margaretha-devault-2011-approach}} \\\midrule
{Costs} & {Paradise} & {\shortciteA{bickmore2006health}} \\\midrule
{(Task/interaction) Complexity}& {Open concept count} & {\shortciteA{dubois2001open}} \\\cmidrule{2-3}
& {Benchmark graph} & {\shortciteA{paek-2001-empirical}}\\\cmidrule{2-3}
& {Survey} & {\shortciteA{cheng2021exploring}}
\\\midrule
{{Content evaluation}} &
{BLEU(-4)/METEOR/ROUGE-L/CIDEr/SKIP-THOUGHT} &
{\shortciteA{D2019automatic,Firdaus2020more}} \\\cmidrule{2-3}
& {Precision, recall, f-measures} &
{\shortciteA{campillos2020designing}} \\\midrule
{{Efficiency}} 
& {Time (spend on method}) &
{\shortciteA{bickmore2006health,tsai2022sema}} \\\cmidrule{2-3}
& {Correct rate} &
{\shortciteA{tsai2022sema}} \\\cmidrule{2-3}
& {Correlate attributes to likes} &
{\shortciteA{pereira2018quality}} \\\cmidrule{2-3}
& {Number of dialogue turns} &
{\shortciteA{abu-shawar-atwell-2007-different,bickmore2006health,song2019task,takanobu-etal-2020-goal}} \\\cmidrule{2-3}
& {System logs} &
{\shortciteA{Foster-etal-2009-comparing}} \\\cmidrule{2-3}
& {Survey} &
{\shortciteA{campillos2020designing,campillos2021lessons,chen2021_usability,crutzen2011artificially,Foster-etal-2009-comparing,ham-etal-2020-end,ihsani2021conversational,pricilla2018designing,Rietz2019ladderbot,roque2021content}} \\\cmidrule{2-3}
& {Interview} &
{\shortciteA{pricilla2018designing}} \\\cmidrule{2-3}
& {Feedback} &
{\shortciteA{tsai2022sema}} \\\midrule
{{Functionality}} &
{Survey} &
{\shortciteA{maniou2020employing}} \\\cmidrule{2-3}
& {Focus group} &
{\shortciteA{maniou2020employing}} \\\midrule
{Implementation} &
{Focus group} &
{\shortciteA{schmidlen2019patient}} \\\midrule
{{Overall evaluation}} &
{Combined score ((inform + success) x0.5 +
BLEU)} & {\shortciteA{peng-etal-2021-raddle}} \\\cmidrule{2-3}
&{AutoJudge} & {\shortciteA{deriu-cieliebak-2019-towards}} \\\cmidrule{2-3}
& {ENIGMA} & {\shortciteA{jiang-etal-2021-towards}} \\\cmidrule{2-3}
& {Survey/rating} &
{\shortciteA{campillos2020designing,okanovic2020can,xiao2020if}} \\\cmidrule{2-3}
& {Interview} &
{\shortciteA{xiao2020if,yuan2008human}} \\\cmidrule{2-3}
& {Feedback} &
{\shortciteA{yuan2008human}} \\\cmidrule{2-3}
& {Focus group} &
{\shortciteA{schmidlen2019patient}} \\\cmidrule{2-3}
& {Children Turing Test} & {\shortciteA{liu2005research}}
\\\midrule
{{Performance}} &
{Answer delivery delay} &
{\shortciteA{bhawiyuga2017design}} \\\cmidrule{2-3}
& {Slot error rate} &
{\shortciteA{campillos2020designing}} \\\cmidrule{2-3}
& {Hit@1/Hit@5} &
{\shortciteA{su-etal-2020-moviechats}} \\\cmidrule{2-3}
& {Running time system/classifier} &
{\shortciteA{xu2020healthcare}} \\\cmidrule{2-3}
& {Precision, recall, mean average precision} &
{\shortciteA{tsai2022sema}} \\\cmidrule{2-3}
& {ECHO} & {\shortciteA{forkan2020echo}}\\\cmidrule{2-3}
& {ConvLab-2} & {\shortciteA{zhu-etal-2020-convlab}}\\\cmidrule{2-3}
& {Survey} &
{\shortciteA{campillos2021lessons,eren2021determinants,maniou2020employing,mokmin2021evaluation,trapero2020integrated}} \\\cmidrule{2-3}
& {Interview} &
{\shortciteA{mokmin2021evaluation,yuan2008human}} \\\cmidrule{2-3}
& {Analytic session/feedback} &
{\shortciteA{mokmin2021evaluation,yuan2008human}}\\\cmidrule{2-3}
& {Focus group} &
{\shortciteA{maniou2020employing}} \\\midrule
{Recommendation quality} &
{Survey} &
{\shortciteA{ihsani2021conversational,theosaksomo2019conversational}} \\\midrule
{Response selection} &
{Recall@1/Recall@3} &
{\shortciteA{mi-etal-2021-self}} \\\midrule
{{Robustness}} &
{RADDLE (benchmark)} &
{\shortciteA{peng-etal-2021-raddle}} \\\cmidrule{2-3}
& {LAUG-toolkit} & {\shortciteA{liu-etal-2021-robustness}}\\\cmidrule{2-3}
& {Attack using adversial agents} &
{\shortciteA{cheng-etal-2019-evaluating}} \\\midrule
{Similarity} & {Greedy
matching/vector extrema/embedding average} &
{\shortciteA{D2019automatic}}\\
\end{longtable}}

\paragraph{Efficiency} tends to refer to how a specific goal is achieved given for example a certain time frame, resources or costs. Depending on the task, efficiency might be defined in different ways. In customer service, quick issue resolution is often one of the main goals. So the less turns needed, the more efficient the system might be. While in healthcare, other measures might be more indicative of efficiency.

\emph{Automatic evaluation} has been conducted in several ways. The most straightforward way to measure efficiency is probably simply measuring the number of dialogue turns. This is for example done by \citeA{takanobu-etal-2020-goal}. Here the user should be able to accomplish the task within 20 turns. Time spent on achieving the goal can also be a measure of efficiency. \citeA{tsai2022sema} measure the time needed to accomplish a task with the dialogue system and compared this to the original paper-based method that was often used for their task. In the same experiment, the user had to accomplish a set of five tasks. To measure efficiency (and compare to the original paper-based method) they also counted how many tasks were correctly completed. \citeA{Foster-etal-2009-comparing} take a similar approach and extract both these measure from the system logs together with the time the system needs to respond to the users utterance. All of these metrics aim for a quantitative measure of efficiency. 

\emph{Human evaluation} can also be used. On the one hand this might be surprising as some of the elements of efficiency (such as time) might be easy to objectively measure. On the other hand, perception might still be very important as it has been shown that there is a discrepancy between the perceived time and the actual time (as demonstrated by \shortciteR{thompson1996effects}). The perceived waiting time is more predictive for patient satisfaction than the actual waiting time \cite{thompson1996effects}. This, for example, might also hold for a situation where a chatbot hands over the conversation to a human employee and the customer is placed in a queue. The perceived waiting time might be much longer than the actual waiting time. Therefore, it is of importance to not only objectively measure constructs like time, but also the perceived time by conducting human evaluation. \citeA{tsai2022sema} actually asks participants for some more informal feedback next to the quantitative measures. Others only focus on perceived efficiency. For example, \citeA{ihsani2021conversational} create a conversational recommender system and aim to measure the perceived efficiency. Users had to explore the system and afterwards fill in a questionnaire stating if they agree or disagree with eight statements. The statement regarding perceived efficiency in their questionnaire is as follows: “I can find a product that I prefer fast”. Their concept of efficiency relates again back to the time taken to achieve the user goal. Existing surveys are also used to measure efficiency. \citeA{roque2021content} actually apply a standardised survey, namely the System Usability Scale (SUS) \cite{brooke1996sus}. Next to efficiency this scale for example also measures learnability and satisfaction. One paper combines a questionnaire with conducting both a pre-test and post-test interview \cite{pricilla2018designing}. To conclude, by looking not only at quantitative efficiency but also at perceived efficiency a different conclusion on the efficiency of the system might be drawn.

\paragraph{Overall evaluation} is rather frequently measured by scholars, but a clear definition is oftentimes lacking. It seems that the overall performance of the system is key in these cases, but the way in which this construct is measured shows that also other (perceived) constructs can be involved to determine the overall evaluation of the system.

\emph{Automatic evaluation} is only used in one paper. \citeA{peng-etal-2021-raddle} create a combined score which incorporates BLEU and measures for inform and success. This is meant as an overall quality measure. In this score, BLEU is a measure for fluency compared to human responses, inform measures how well the system provides correct entities and success shows how well the system answers to all the given questions \cite{peng-etal-2021-raddle}. 

\emph{Human evaluation} was nearly constantly used. In certain situations expert users are brought in (for example by using human-computer interaction experts as in \citeR{yuan2008human}) but mostly the general public or target group is asked to give an overall opinion on the dialogue system. For example, \citeA{xiao2020if} ask participants about their opinion of the chatbot in an interview and elaborate on that by asking for a rating on their overall chat experience. Similarly, \citeA{yuan2008human} rely on informal feedback and interviews with expert evaluators. Often, the questions used to ask participants their opinion on the overall evaluation are very general or not completely clear.

\paragraph{Performance} is also used as a construct on its own and aims to measure how well a system performs. The idea of performance relates closely to the constructs that measure efficiency and overall evaluation. These constructs tend to overlap (and therefore also overlap in approaches) which makes setting clear borders about the definition hard. \citeA{maniou2020employing} for example define performance in relation to the ability to respond timely and efficiently. Again, there is the same discrepancy between more objective performance and perceived performance, which is also mentioned by \citeA{eren2021determinants}. They mention that objective performance measures the actual performance, the perceived performance is a combination to how customers evaluate the quality and value \cite{eren2021determinants}. 

Performance shows that some constructs are more abstract and overarching than other constructs. This might have implications on the methods used for evaluation. As such a construct could possibly encompass a wide variety of ideas/definitions, more (different) methods could be expected (see in this context again Figure~\ref{fig:operationalisation}). What we see in practice is that this does not happen and performance is actually reduced to very specific definitions, with specific metrics.

\emph{Automatic evaluation} is oftentimes applied to measure this construct, as researchers often aim to objectively measure performance. Just as with efficiency, performance is measured in terms of time. The running time (such as in \citeR{xu2020healthcare}) for both different classifiers and the whole system is used as a proxy for performance. The answer delivery delay (as in \citeR{bhawiyuga2017design}) measures the time between a message from the user to a returned response by the bot. Other measures are also used, such as Hit@1 and Hit@5 (introduced by \shortciteR{zhang2018personalizing}). This measure returns 1 if a model chooses a correct response. It seems therefore that \citeA{su-etal-2020-moviechats} use correct response as a proxy for measuring performance.

\emph{Human evaluation} can also be applied to measure performance. Similar to efficiency, to measure performance another standardised survey is used, namely the UTAUT-2 \cite{venkatesh2012consumer}. \citeA{mokmin2021evaluation} use this model in a health care context to measure the user perceptions towards the technology focusing on performance expectancy. They focus not on perceived performance but on expected performance by asking three questions (for example `Using chatbot enables me to understand health better') on a 7-point Likert scale. They combine this with other sources of information such as interviews and an analytic session (measuring for example percentage of matched intents).

\subsection{System in context}
\label{subsec:context}

\subsubsection{Task success}
\label{subsub:success}
Table \ref{tab:succ} shows constructs that are related to the task-oriented nature of the dialogue system, namely focusing on task success. Not surprisingly, in comparison to the other tables this table contains a number of constructs that can be explicitly linked to certain domains and contexts. Comparing treatments and health outcomes are of importance in a healthcare context, knowledge of material and learning outcomes for education, and (customer) loyalty is key for a customer service chatbot. Compared to the intrinsic constructs of Section \ref{subsec:intrinsic}, most of these constructs in this table are mainly measured by human evaluation. Only the construct success itself is measured by several different automatic evaluation approaches.

{\small
\begin{longtable}[]{lp{5cm}p{5.75cm}}
\toprule\noalign{}
\endhead

\hline\\
\caption{Constructs and metrics for measuring task success (continued on next page)}
\label{tab:succ}
\endfirstfoot
\hline\\
\caption*{Table~\ref{tab:succ}: Constructs and metrics for measuring task success (continued on next page)}
\endfoot
\bottomrule\\
\caption*{Table~\ref{tab:succ}: Constructs and metrics for measuring task success (14 constructs; 43 distinct papers)}
\endlastfoot

\textbf{Construct} & \textbf{Approach} & \textbf{Papers} \\\midrule
{Comparing treatments} & {Intervention study} & {\shortciteA{Jang2021mobile}} \\\cmidrule{2-3}
& Randomized Clinical Trials & {\shortciteA{bickmore2006health}}\\\midrule
{Competence} & {Survey} & {\shortciteA{jimenez2021find}}\\\cmidrule{2-3}
& {DISC dialogue Management Grids} & {\shortciteA{bickmore2006health}}\\\midrule
{{Effectiveness}} & {Correctness and time needed to complete task} & {\shortciteA{okanovic2020can,tsai2022sema}} \\\cmidrule{2-3}
&{User simulation} & {\shortciteA{lopez2003assessment}} \\\cmidrule{2-3}
&{Conversation logs} & {\shortciteA{han2021designing}}\\\cmidrule{2-3}
& {Survey/Rating} &
{\shortciteA{liu2020cbet,okanovic2020can,pricilla2018designing,Rietz2019ladderbot,xiao2020if,zhang2021informing}} \\\cmidrule{2-3}
& {Completed tasks/number of errors in task/manner of performance
participants} & {\shortciteA{al2021building}} \\\cmidrule{2-3}
& {Live evaluation} & {\shortciteA{xiao2020if}} \\\cmidrule{2-3}
& {Task mapping} & {\shortciteA{pricilla2018designing}} \\\cmidrule{2-3}
& {Field study} & {\shortciteA{ren2020understanding}} \\\cmidrule{2-3}
& {Ranking} & {\shortciteA{nuruzzaman2020intellibot}} \\\midrule
{{Errors}} & {Correction rate} &
{\shortciteA{schumaker2007evaluation}} \\\cmidrule{2-3}
& {DialTest} & {\shortciteA{liu2021dialtest}} \\\cmidrule{2-3}
& {Conversation logs} & {\shortciteA{candello2018recovering}} \\\cmidrule{2-3}
& {Thematic analysis} & {\shortciteA{candello2018recovering}} \\\cmidrule{2-3}
& {Survey/Rating} & {\shortciteA{duggenpudi-etal-2019-samvaadhana,roque2021content}} \\\cmidrule{2-3}
& {Expert evaluation} & {\shortciteA{coniam2014linguistic}} \\\midrule
{{Feasibility}} & {User engagement metrics} &
{\shortciteA{miraj2021development}} \\\cmidrule{2-3}
& {Survey/Rating} & {\shortciteA{miraj2021development,xiao2020if}} \\\cmidrule{2-3}
& {Interview} & {\shortciteA{miraj2021development}} \\\cmidrule{2-3}
& {Live evaluation} & {\shortciteA{xiao2020if}} \\\midrule
{{Health outcomes}} & {Survey} & {\shortciteA{kataoka2021development,puron2021p}} \\\cmidrule{2-3}
& {Interview} & {\shortciteA{mokmin2021evaluation}} \\\cmidrule{2-3}
& {Intervention fidelity} & {\shortciteA{piau2019smartphone}} \\\cmidrule{2-3}
& {Engagement metrics}
& {\shortciteA{bickmore2006health,piau2019smartphone}} \\\midrule
{Intellectual} & {Survey} & {\shortciteA{jimenez2021find}} \\\midrule
{{Intelligence}} & {Survey} & {\shortciteA{vanderlyn2021seemed}} \\\cmidrule{2-3}
& {Interview} & {\shortciteA{yuan2010evaluation}} \\\midrule
{Interpretability} & {Survey} & {\shortciteA{kadariya2019kbot}} \\\midrule
{Knowledge of material} & {MC to test knowledge} &
{\shortciteA{Dzikovska-etal-2012-evaluating}} \\\midrule
{Learning outcomes} & {Survey} & {\shortciteA{okanovic2020can,Okonkwo2020python,weisz2019bigbluebot,zhu2022me}} \\\midrule
{(Customer) loyalty} & {Survey} & {\shortciteA{cheng2020how,liu2020cbet}} \\\midrule
{Persuasion outcomes} & {Survey} & {\shortciteA{Shi-etal-2021-refine-imitate}} \\\midrule
{{Success}} & {Success rate} & {\shortciteA{aust1998evaluating,song2019task}} \\\cmidrule{2-3}
& {Task completion} & {\shortciteA{peng-etal-2021-raddle}} \\\cmidrule{2-3}
& {Inform (f1)} & {\shortciteA{peng-etal-2021-raddle,takanobu-etal-2020-goal}} \\\cmidrule{2-3}
& {Number of participants who successfully completed task} &
{\shortciteA{okanovic2020can}} \\\cmidrule{2-3}
& {Recall@k} & {\shortciteA{lowe-etal-2016-evaluation}} \\\cmidrule{2-3}
& {Match rate} & {\shortciteA{takanobu-etal-2020-goal}} \\\cmidrule{2-3}
& {Conversation turns per session (expected)} & {\shortciteA{zhou2020design}} \\\cmidrule{2-3}
& {Behavioural measures} & {\shortciteA{weisz2019bigbluebot}} \\\cmidrule{2-3}
& {User simulation} & {\shortciteA{lopez2003assessment}} \\\cmidrule{2-3}
& {Paradise} & {\shortciteA{bickmore2006health}}\\\cmidrule{2-3}
& {Logs} & {\shortciteA{Foster-etal-2009-comparing,holmes2019usability,weisz2019bigbluebot,zhou2020design}} \\\cmidrule{2-3}
& {Think aloud protocol} & {\shortciteA{holmes2019usability}} \\\cmidrule{2-3}
& {Survey} & {\shortciteA{Foster-etal-2009-comparing,ham-etal-2020-end,holmes2019usability,lowe-etal-2016-evaluation}} \\
\end{longtable}}

\paragraph{Effectiveness} measures the extent to which the system achieves its intended results, also taking into account the process of achieving that result (which relates closely to efficiency). This definition is similar to how effectiveness is used in the papers, although it depends a bit on the context what is seen as effective. \citeA{zhang2018personalizing} for example define effectiveness as the capability of the system to understand the users’ aims and provide appropriate feedback. 

\emph{Automatic evaluation} can be applied to measure effectiveness in three ways. The time needed to complete the task and the correctness of the task were used to say something about the dialogue systems effectiveness \cite{okanovic2020can,tsai2022sema} We have seen time needed also as a measure of efficiency, but the combination with correctness of the final tasks relates it to efficiency. By extracting data from the conversation logs, \citeA{han2021designing} aim to create a framework for quantifying effectiveness of dialogue systems. They define effectiveness in terms of elicitation abilities, user experience and ethics, and measure things like the completion rate (number of participants completing an interview with the system)\cite{han2021designing}. Another way is to use a user simulation. \citeA{lopez2003assessment} employ a user simulator to generate conversations and automatically measure the implicit recovery (percentage of incorrect simulator sentences correctly understood by the system)\cite{lopez2003assessment}. The more misunderstandings the less effective the system is.

\emph{Human evaluation} is most frequently used, especially by means of surveys. Similar to performance in Section \ref{subsub:perf}, the System Usability Scale (SUS) is used for measuring effectiveness. Other methods are also used and sometimes multiple methods are combined. \citeA{pricilla2018designing}, for example, employ surveys but also task mapping. With the survey they ask participants to rate on a 5-point scale how effective the system was to use.  With task mapping the researchers count how many of the predefined tasks are actually completed by the user.  

\paragraph{Errors} often reflect the number of errors occurring in dialogues and the kinds of errors that can be distinguished. We will see that there are many different definitions of what constitutes to an error. Some errors might also be harder to detect than others. \citeA{coniam2014linguistic} for example focus on linguistic errors, such as grammatical errors.\footnote{See \citeA{higashinaka-etal-2015-towards} for a discussion of errors in task-oriented versus chat-oriented dialogue systems.} 

\emph{Automatic evaluation} can be applied in several ways. Similar as in Section \ref{subsub:perf}, DialTest (a RNN-based system) is also used to detect erroneous behaviour of the intent detector and consequently create a more robust system \cite{liu2021dialtest}. \citeA{liu2021dialtest} compare multiple transformations (such as word insertion) for detecting such errors.  Next to DialTest, the correction rate is for example used to measure the amount of user corrections and the consequent error types (such as nonsense replies) associated with this \cite{schumaker2007evaluation}. The correction rate counts the number of system responses corrected by the user divided by the total amount of conversations \cite{schumaker2007evaluation}.

\emph{Human evaluation} can also be used. Most strikingly is probably the thematic analysis conducted by \citeA{candello2018recovering}. The scholars use thematic network analysis (first introduced by \citeR{stirling2001}) on dialogue logs to figure out the points of dialogue failure. The goal of this method is to present a thematic analysis as network-like structures, resulting in systematic presentations of the analysis to summarise main themes in the data \cite{stirling2001}. By using this method, \citeA{candello2018recovering} manually identified the points of dialogue failure.

\paragraph{Success} is similar to effectiveness but with less focus on the process of getting to the end-goal, focusing only on the obtained results. This is normally not measured on a scale but in a binary way, namely success or no success. \citeA{peng-etal-2021-raddle} for example define success to be an exact match to the requested information and meeting the user goal. \citeA{Foster-etal-2009-comparing} employ a robot for a joint construction task and define success as being able to create a target object and remembering how to create certain shapes. 

\emph{Automatic evaluation} is more often used to measure this construct, in comparison to the other constructs discussed in this table. Inform is used by both \citeA{peng-etal-2021-raddle} and \citeA{takanobu-etal-2020-goal}. \citeA{takanobu-etal-2020-goal} use inform F1 (fulfilment of information request) and argue that a dialogue is successful if and only if both inform recall and match rate (entity meets specified constraints) are one. \citeA{peng-etal-2021-raddle} use inform which, similarly to in \citeA{takanobu-etal-2020-goal}, measures if the correct entity is given. Another method, maybe the most straightforward one, is the amount of conversation turns per session, although it depends on the goal of the conversation what the right number of turns would be. In some cases, one might argue that less is more successful (if a customer wants answer to a simple question); in other cases more turns might be indicative of a more successful conversation (for example with a social dialogue system). \citeA{zhou2020design} employ the expected conversation turns for a social chatbot; when the score is larger the social chatbot is better engaged. 

\emph{Human evaluation} is also sometimes used to measure success. \citeA{holmes2019usability} use for example a combination of multiple surveys and a think-aloud protocol. They focus on task completion (which is indicative of success), by for example counting how many task repetitions are required. The task completion time is also taken from the logs. 
Often this construct has been measured automatically and focuses on a binary success/no-success division. Success of course depends a great deal on the task at hand. Thus, not only automatic evaluation is used to measure success, but in some cases also human evaluation although this might be the less obvious choice (this again shows the division between perceived success and success on its own).

\subsubsection{Usability}
\label{subsub:usab}

Table \ref{tab:usab} shows constructs related to the concept of usability. Almost all of these constructs are mostly measured by human evaluation. Compared to the other system in context tables this one is relatively short and focuses mostly on the condition of a system such as ease of use, accessibility and learnability. Usability itself is also often used as a construct (the construct will be discussed later on). 

{\small
\begin{longtable}[]{lp{5cm}p{5.75cm}}
\toprule\noalign{}
\endhead

\hline\\
\caption{Constructs and metrics for measuring usability (continued on next page)}
\label{tab:usab}
\endfirstfoot
\hline\\
\caption*{Table~\ref{tab:usab}: Constructs and metrics for measuring usability (continued on next page)}\endfoot
\bottomrule\\
\caption*{Table~\ref{tab:usab}: Constructs and metrics for measuring usability (9 constructs; 36 distinct papers)}
\endlastfoot
\textbf{Construct} & \textbf{Approach} & \textbf{Papers} \\\midrule
{{Accessibility}} & {Survey} & {\shortciteA{maniou2020employing,orden2021analysis}} \\\cmidrule{2-3}
& {Focus group} & {\shortciteA{maniou2020employing}} \\\midrule
{{Ease of use}} & {Completed tasks/number of errors in
task/manner of performance participants} & {\shortciteA{al2021building}} \\\cmidrule{2-3}
& {Survey} & {\shortciteA{crutzen2011artificially,ihsani2021conversational,rese2020chatbots,theosaksomo2019conversational}} \\\cmidrule{2-3}
& {Expert evaluation/feedback} & {\shortciteA{yuan2010evaluation}} \\\cmidrule{2-3}
& {Interview} & {\shortciteA{yuan2010evaluation}} \\\midrule
{Effort/convenience} & {Survey} & {\shortciteA{chen2021_usability,mokmin2021evaluation,rese2020chatbots,sensuse2019chatbot,trapero2020integrated}} \\\midrule
{Friendliness} & {Survey} & {\shortciteA{cheng2021exploring,Okonkwo2020python}} \\\midrule
{{Intention to use}} & {Expected conversation turns per
session} & {\shortciteA{zhou2020design}} \\
& {Engagement duration/response length/response informativeness/response
quality index/} & {\shortciteA{xiao2020if}} \\\cmidrule{2-3}
& {Logs} & {\shortciteA{zhou2020design}} \\\cmidrule{2-3}
& {Survey/rating} & {\shortciteA{cheng2020how,chung2021chatbot,li2021makes,mokmin2021evaluation,ren2020understanding,rese2020chatbots,sensuse2019chatbot,trapero2020integrated,weisz2019bigbluebot,xiao2020if}} \\\cmidrule{2-3}
& {Negative/positive feedback (emoji)} & {\shortciteA{chung2021chatbot}} \\\cmidrule{2-3}
& {Live evaluation} & {\shortciteA{xiao2020if}} \\\midrule
{{Interactivity}} & {Survey} & {\shortciteA{li2021makes,maniou2020employing,orden2021analysis}} \\\cmidrule{2-3}
& {Focus group} & {\shortciteA{maniou2020employing}} \\\midrule
{{Learnability}} & {Survey} &
{\shortciteA{pricilla2018designing,Rietz2019ladderbot,roque2021content}} \\\cmidrule{2-3}
& {Interview} & {\shortciteA{pricilla2018designing}} \\\midrule
{{Responsiveness}} & {Average minutes per
interaction/fallbacks per interaction /fallbacks per minute/completion
rate} & {\shortciteA{diederich2020designing}} \\\cmidrule{2-3}
& {Survey/rating} & {\shortciteA{chen2021_usability,duggenpudi-etal-2019-samvaadhana,li2021makes}} \\\midrule
{{Usability}} & {Survey} & {\shortciteA{chung2021chatbot,el2021chase,holmes2019usability,Jang2021mobile,kadariya2019kbot,piao2020development}} \\\cmidrule{2-3}
& {Interview} & {\shortciteA{beilharz2021development,yuan2010evaluation}} \\\cmidrule{2-3}
& {Negative/positive feedback} & {\shortciteA{chung2021chatbot}} \\\cmidrule{2-3}
& {Think aloud protocol} & {\shortciteA{holmes2019usability}} \\\cmidrule{2-3}
& {Video} & {\shortciteA{holmes2019usability}} \\\cmidrule{2-3}
& {Focus group} & {\shortciteA{schmidlen2019patient}} \\\cmidrule{2-3}
& {Functional testing} & {\shortciteA{bhawiyuga2017design}} \\\cmidrule{2-3}
& {Usability heuristics} & {\shortciteA{langevin2021heuristic,sugisaki2020usability}}\\
\end{longtable}}

\paragraph{Ease of use} expresses how easy the system is to use for the intended user group, often without having any background knowledge about the system.

Although no \emph{automatic evaluation} was carried out to measure ease of use, as seen there have been attempts to quantify this information by counting the number of completed tasks. Measures that focus on automatic analysis of ease of use would probably relate closely to other constructs like efficiency. The time needed to complete a task for example could also indicate the ease of use. We could also think of other measures. If users are able to finish a task and not drop-out or if a user simulator is able to finish a task this could indicate the ease of use as well.

\emph{Human evaluation} is only applied to measure this construct, although \citeA{al2021building} try to quantify this by counting for example the completed tasks and the number of errors. Surveys, feedback and interviews are used to get the perceptions of users' ease of use. \citeA{theosaksomo2019conversational} focus for example on measuring usability goals by examining the ease with which participants could use their chatbot but also how easy it was to remember how to perform a recommendation task. This was measured by using a post-test questionnaire. 

\paragraph{Intention to use} can be measured either before the user has ever used the dialogue system itself, or after using the dialogue system (i.e., interest in future use). In the customer service literature this is also often called continuance intention (see for example \citeR{ashfaq2020_1701379984970}). One paper in the data set \cite{li2021makes} actually names it precisely this, but for practical purposes this term is grouped together with intention to use.

\textit{Automatic evaluation} of intention to use is not used. It is probably hard to measure if users want to (continuously) use a dialogue system because this construct again leans considerably on the perceptions of the user.

\textit{Human evaluation} is thus only used for measuring intention to use. Some papers also aim to quantitatively measure this construct. The expected conversation turns per session (the average of the number of conversation turns in a session, for multiple session collected over a longer period of time) is used to indicate the willingness of users to share their time with the chatbot over a longer period of time \cite{zhou2020design}. As shown in Section~\ref{subsub:success} this metric is also used for a measure of overall success, meaning that willingness for continued use also directly affects success. \citeA{xiao2020if} aim to capture some characteristics (for example response length and engagement duration) of the engagement/conversation with the system and uses those as a proxy for intention to use. For example, a longer engagement duration might imply higher intentions to use the system. Next to these more quantitative methods, qualitative methods like surveys are often used.

\paragraph{Usability} is also a construct of its own, which can be defined as: `the capability in human functional terms to be used easily and effectively by the specified range of users, given specified training and user support, to fulfil the specified range of tasks, within the specified range of environmental scenarios' \cite[p.\thinspace 340]{shackel2009usability}. Usability is used as a separate construct by many papers but the definition of \citeA{shackel2009usability} shows that this construct is actually composed of other constructs like `ease of use'. Unfortunately, many papers don't fully specify what they mean by usability, and no \emph{automatic evaluation} is used to measure usability. 

\emph{Human evaluation} is the only evaluation approach applied to measure this construct, as Table \ref{tab:usab} shows. Sometimes complete surveys are created to measure usability, for example by \citeA{holmes2019usability}. They create a survey called the Chatbot Usability Questionnaire (CUQ), to obtain a usability score. The questionnaire consists out of 16 items and the authors show what is assessed with this questionnaire (focusing for example on onboarding, responses, navigation). Their CUQ is based on a design tool called Chatbottest\footnote{https://chatbottest.com/}. This tool is an open source project that provides some directions for evaluation. Usability heuristics are also created, for example by \citeA{langevin2021heuristic}. Their heuristics include items such as visibility of the system status, trustworthiness, and user control and freedom. What is also observed is that the phrase ‘usability study’ is also often used for different measurement approaches. \citeA{el2021chase} use a questionnaire as part of the usability study while \citeA{holmes2019usability} see a usability study as consisting of a combination of video (logs), a questionnaire and a think aloud protocol. 

\subsubsection{User experience}
\label{subsub:exp}
User experience is together with NLG one of the longest tables in this paper. This shows that there is still a major emphasis on how dialogue systems are perceived by the users. In comparison to the other tables, the constructs in this table mainly concern the effects that a system has on its users. As a result, almost all approaches involve human evaluation.

{\small
\begin{longtable}[]{p{3.25cm}p{5cm}p{5.75cm}}
\toprule\noalign{}
\endhead

\hline\\
\caption{Constructs and metrics for measuring user experience (continued on next page)}
\label{tab:exp}
\endfirstfoot
\hline\\
\caption*{Table~\ref{tab:exp}: Constructs and metrics for measuring user experience (continued on next page)}
\endfoot
\bottomrule\\
\caption*{Table~\ref{tab:exp}: Constructs and metrics for measuring user experience (38 constructs; 65 distinct papers)}
\endlastfoot
\textbf{Construct} & \textbf{Approach} & \textbf{Papers} \\\midrule
{{Acceptability}} & {User engagement metrics} &
{\shortciteA{miraj2021development,piau2019smartphone}}\\\cmidrule{2-3}
& {Interview/focus group} & {\shortciteA{miraj2021development,schmidlen2019patient,yuan2010evaluation}} \\\cmidrule{2-3}
& {Survey/rating} & {\shortciteA{D2019automatic,Fiore2019forgot,Jang2021mobile,kadariya2019kbot,miraj2021development,mokmin2021evaluation,okanovic2020can,rese2020chatbots,van2019chatbot}} \\\cmidrule{2-3}
& {Expert evaluation/feedback} &
{\shortciteA{coniam2014linguistic,yuan2010evaluation})} \\\midrule
{{Anxiety}} & {Survey} & {\shortciteA{diederich2020designing,li2021makes,mokmin2021evaluation,rese2020chatbots}} \\\cmidrule{2-3}
& {Focus group} & {\shortciteA{diederich2020designing}} \\\midrule
{Assurance} & {Survey} & {\shortciteA{li2021makes}} \\\midrule
{{Attitude}} & {Survey} & {\shortciteA{jimenez2021find,piau2019smartphone}} \\\cmidrule{2-3}
& {Intervention fidelity} &
{\shortciteA{piau2019smartphone}} \\\midrule
{Attitude to improve health} & {Survey} &
{\shortciteA{mokmin2021evaluation}} \\\midrule
{Autonomy} & {Survey} & {\shortciteA{jimenez2021find}} \\\midrule
{Challenges} & {Interview} & {\shortciteA{han2021designing,ren2020understanding}} \\\midrule
{Confirmation} & {Survey} & {\shortciteA{li2021makes}} \\\midrule
{Cooperativeness} & {Survey} &
{\shortciteA{Eric-etal-2017-key}} \\\midrule
{Corporate reputation} & {Survey} & {\shortciteA{eren2021determinants}} \\\midrule
{(self-)Efficacy} & {Survey} &
{\shortciteA{chen2021_usability,mokmin2021evaluation,sperli2020deep}} \\\midrule
{{Engagement}} & {Conversation logs} &
{\shortciteA{han2021designing}} \\\cmidrule{2-3}
& {Survey (legoeval)/Rating} &
{\shortciteA{burtsev2018first,cheng2020ai,li-etal-2021-legoeval,su-etal-2020-moviechats}} \\\midrule
{Entertainment}
& {Survey} & {\shortciteA{cheng2020how}} \\\midrule
\parbox{3.25cm}{Emotional\\connection} & {Survey} &
{\shortciteA{Foster-etal-2009-comparing,Gonzales2017bots}} \\\cmidrule{2-3}
& {Analysis/direct observation} &
{\shortciteA{Gonzales2017bots}} \\\cmidrule{2-3}
& {Interview} & {\shortciteA{Gonzales2017bots}} \\\midrule
{Empathy} & {Survey} & {\shortciteA{cheng2021exploring,orden2021analysis,weisz2019bigbluebot}} \\\cmidrule{2-3}
& {Empathy level} & {\shortciteA{han2021designing}}\\\midrule
{{Experience}} & {Survey/Rating} &
{\shortciteA{Fiore2019forgot,han2021designing,holmes2019usability,rese2020chatbots,sperli2020deep,van2019chatbot,vanderlyn2021seemed,weisz2019bigbluebot,zhu2022me})}\\\cmidrule{2-3}
& {Interview} & {\shortciteA{candello2019effect,prasad2019dara,ren2020understanding,yuan2008human}}\\\cmidrule{2-3}
& {(video) Log} & {\shortciteA{holmes2019usability}} \\\cmidrule{2-3}
& {Fieldwork} & {\shortciteA{candello2019effect}}\\\cmidrule{2-3}
& {Think aloud protocol} &
{\shortciteA{holmes2019usability}} \\\midrule
{Familiarity} & {Survey} &
{\shortciteA{kattenbeck2018airbot}} \\\midrule
{Fidelity} & {Intervention fidelity} &
{\shortciteA{piau2019smartphone}} \\\midrule
{{Enjoyment}} & {Survey} & {\shortciteA{chen2021_usability,crutzen2011artificially,diederich2020designing,pricilla2018designing,rese2020chatbots,Rietz2019ladderbot,weisz2019bigbluebot,zhu2022me}} \\\cmidrule{2-3}
& {Focus group/interview} &
{\shortciteA{diederich2020designing,pricilla2018designing,yuan2010evaluation}} \\\cmidrule{2-3}
& {Expert evaluation/feedback/transcript analysis} &
{\shortciteA{yuan2010evaluation}} \\\midrule
{{Helpfulness}} & {Survey} & {\shortciteA{pricilla2018designing,van2019chatbot,weisz2019bigbluebot}} \\\cmidrule{2-3}
& {Interview} & {\shortciteA{pricilla2018designing}} \\\midrule
{Independence} & {Survey} &
{\shortciteA{chen2021_usability}} \\\midrule
{Indistinguishability} & {Multiple choice} &
{\shortciteA{ccetinkaya2020developing}} \\\midrule
{Likability} & {Survey} & {\shortciteA{el2021chase,vanderlyn2021seemed}} \\\midrule
{Motivation} & {Survey} &
{\shortciteA{cheng2020ai,trapero2020integrated}} \\\midrule
{Patronage intentions} & {Survey} & {\shortciteA{van2019chatbot}} \\\midrule
{{Personality}} & {Survey} &
{\shortciteA{chen2021_usability}} \\\cmidrule{2-3}
& {Interview} & {\shortciteA{prasad2019dara}} \\\midrule
{{Personalisation}} & {Survey} & {\shortciteA{maniou2020employing,zhu2022me}} \\\cmidrule{2-3}
& {Focus group} & {\shortciteA{maniou2020employing}} \\\midrule
{{Realism/humanness}} & {Survey} &
{\shortciteA{diederich2020designing,Eric-etal-2017-key,li-etal-2021-legoeval,vanderlyn2021seemed,zhang2021informing}} \\\cmidrule{2-3}
& {Focus group} & {\shortciteA{diederich2020designing}} \\\cmidrule{2-3}
& {Ranking} & {\shortciteA{deriu-etal-2020-spot}} \\\midrule
{Reliability} & {Survey/rating} & {\shortciteA{cheng2021exploring,duggenpudi-etal-2019-samvaadhana,li2021makes}} \\\midrule
{Resistance} & {Survey} &
{\shortciteA{cheng2021exploring}} \\\midrule
{{Satisfaction}} & {Survey/Rating} &
{\shortciteA{aust1998evaluating,barreto2021development,chen2021_usability,cheng2020how,el2021chase,eren2021determinants,han2021designing,ihsani2021conversational,jimenez2021find,kataoka2021development,kattenbeck2018airbot,orden2021analysis,piao2020development,puron2021p,ren2020understanding,rese2020chatbots,roque2021content,schumaker2007evaluation,sensuse2019chatbot,vanderlyn2021seemed,zhao2019evaluation,zhu2022me}} \\\cmidrule{2-3}
& {Informal user feedback} &
{\shortciteA{abu-shawar-atwell-2007-different}} \\\cmidrule{2-3}
& {Comparison} & {\shortciteA{liu2015ergonomics}} \\\midrule
{Sentiment} & {Conversation logs} & {\shortciteA{han2021designing}} \\\midrule
{{Social presence and influence}} & {Survey} &
{\shortciteA{cheng2020how,diederich2020designing,mokmin2021evaluation,trapero2020integrated}} \\\cmidrule{2-3}
& {Focus group} & {\shortciteA{diederich2020designing}} \\\midrule
{{Trust}} & {Survey/Rating} & {\shortciteA{cheng2021exploring,eren2021determinants,el2021chase,han2021designing,ihsani2021conversational,przegalinska2019bot,ren2020understanding,trapero2020integrated}} \\\cmidrule{2-3}
& {Psychophysiology} & {\shortciteA{przegalinska2019bot}} \\\midrule
{Understandable} & {Survey} & {\shortciteA{campillos2021lessons,campillos2020designing,ihsani2021conversational,li2021makes,okanovic2020can,Rietz2019ladderbot}} \\\midrule
{{Usefulness}} & {Survey} & {\shortciteA{rese2020chatbots,van2019chatbot}} \\\cmidrule{2-3}
& {Interview} & {\shortciteA{mokmin2021evaluation}} \\\midrule
{{Valuable}} & {Survey} &
{\shortciteA{trapero2020integrated}} \\\cmidrule{2-3}
& {Expert evaluation/feedback} &
{\shortciteA{yuan2010evaluation}} \\\cmidrule{2-3}
& {Interview} & {\shortciteA{yuan2010evaluation}} \\\midrule
{{Willingness}} & {Response/engagment
metrics}
& {\shortciteA{xiao2020if}} \\\cmidrule{2-3}
& {Survey/rating} & {\shortciteA{Gonzales2017bots,rese2020chatbots}} \\
\end{longtable}}

\paragraph{Experience} is a difficult to define construct as experience can encompass a wide range of other concepts or even emotions. According to \citeA{hassenzahl2006user} the term has been critiqued over being vague and having a wide variety of meanings. If we try to capture user experience, we could say that user experience aims to measure the overall experience people have when using a dialogue system. Sometimes it is measured together with user satisfaction (such as in \citeR{zhu2022me}). The challenge of defining experience results in the fact that sometimes various different constructs are measured under the term experience. In our data set, no \emph{automatic evaluation} has been used to measure experience.

\emph{Human evaluation} was thus only used to measure experience, as can be seen in Table~\ref{tab:exp}. Noticeable is that the construct is also measured by combining multiple approaches such as survey, logs and think aloud protocols \cite{holmes2019usability} or by combining fieldwork/observations with an interview \cite{candello2019effect}. A specific survey has also been used, namely the NASA Task Load Index. \citeA{sperli2020deep} uses the NASA Task Load Index for measuring experience and more specifically efficacy. Since the NASA Task Load Index \cite{hart1988development} measures the workload for conducting a task, it might be a little far-fetched how this actually translates to experience.  

\paragraph{Realism} involves examining how realistic or humanlike the chatbot is. Research on realism/humanness also involves examining the effects that humanized systems have on users (on their experience for example) often related to the uncanny valley theory \cite{rapp2021human}, where an almost humanlike system induces uncanny feelings with the user \cite{mori2012uncanny}. Similar to experience, no \textit{automatic evaluation} has been used to measure realism.

\emph{Human evaluation} is the only approach used to measure realism. Most often this is done by employing surveys. \citeA{li-etal-2021-legoeval} for example measure engagingness and humanness with the following question: `Which speaker sounds more human.'\footnote{They actually introduce LEGOEval, a toolkit which should make evaluation with crowdsourced workers easier. It incorporates the possibility to add both pre- and post-surveys and gives researchers the possibility to adjust the survey to their own wishes and needs.} \citeA{diederich2020designing} focus on anthropomorphic systems and use both a survey and a focus group. While the focus group was conducted at the start of their design cycle and focused less on human like cues, the survey with 7-point scales was used to figure out if humanlike cues are of value for the users. Questions were used that focused on social presence (with items such as `I felt a sense of human warmth with the tool') and uncanniness (with items such as `I perceived the tool as strange').

\paragraph{Trust} measures the confidence of people in either the dialogue system itself or in some cases the organisation that is represented by the dialogue system. Previous work has stressed the importance of trust in online marketing \shortcite{urban1998trust} and in customer service chatbots \shortcite{folstad2018makes}. \citeA{eren2021determinants} follows the definition of \citeA{morgan1994commitment} in defining trust. They define trust as the confidence discourse partners have in each other’s credibility \cite{morgan1994commitment}. Trust is often seen as an important factor as it for example influences the willingness to communicate with the dialogue system \cite{han2021designing}\footnote{See \citeA{mostafa2022antecedents}, for a discussion of the consequences of chatbot trust.}. Similar to other user experience constructs, no \textit{automatic evaluation} is used.

\emph{Human evaluation} approaches are useful to measure trust. Most of the time survey questions are employed. For example, \citeA{han2021designing} ask participants the following: ‘How much do you trust this chatbot? Please rate it on a scale of 1 to 5’. Interestingly, a very different approach is taken by \citeA{przegalinska2019bot}. Next to using a questionnaire, they also focus on psychophysiological measures for evaluating trust. With this measure they aim to take an objective approach to the construct trust. Although the scholars are not completely clear in their paper as to what is exactly meant by psyhophysiological measures, we could assume that the definition is similar to that of \citeA{dirican2011psychophysiological} in their review of psyhophysiological measures. They mention that these measures make use of the physical signals that a body produces when a psychological change occurs. This can for example be measured by an EEG \cite{dirican2011psychophysiological}. Unfortunately, due to limited discussion of this approach it is not completely clear how this approach functions as a proxy for trust (and possible other constructs). 

\subsection{Recent developments (2022 - \ldots)}
\label{sec:recentdevelopments}
After we completed our review, we manually identified recent papers to discuss some recent developments that have been discussed in the literature after our cutoff date. This also forced us to reflect on the contributions of our survey, and the inevitability of new developments being published after this paper.

\subsubsection{New technologies bring new evaluation measures}\label{subsub:newmodels}
Looking through the history of dialogue systems, it is clear that different kinds of technology may require different forms of evaluation to address their particular strengths and weaknesses. For example, research on `hallucinations' only became widespread after large language models (LLMs) were adopted (see \shortciteR{ji2023survey} for a survey on hallucination). According to \citeA{dale2023navigating} the absence of hallucinations indicates reliability. After all: rule-based models provide much more predictable output (at the cost of having lower coverage and perhaps being perceived as repetitive).\footnote{This is not to say that rule-based systems cannot produce factually incorrect or misleading output. See for example work by \citeR{deemter2018lying} on `lying and computational linguistics' and work by Thomson and Reiter showing that the order in which factually correct statements are presented may lead readers to make incorrect inferences \shortcite{thomson-reiter-2021-generation,thomson2023evaluating}.} Thus, with the rise of new model architectures, it is inevitable that new evaluation metrics will keep being developed.\footnote{Of course these metrics may in turn be based on those same newly developed architectures. For example, the most recent Dialog System Technology Challenge (DSTC 11) saw entries using ChatGPT \cite{chatgpt} to evaluate open domain dialogue systems \shortcite{rodriguez-cantelar-etal-2023-overview}.} At the same time, many of the current constructs and metrics will stay relevant over the next years. Although there might be a shift in importance: a construct like fluency might be less relevant while a construct like factuality might gain importance when evaluating LLM based dialogue systems.

What does the continuous introduction of new metrics mean for us as researchers? First of all, it means that this will not be the last review of common evaluation measures. But \emph{while the popularity of different metrics may wax and wane, the constructs they represent are timeless.} Over time, we hope that the NLP community can help deepen our understanding of those constructs and the way they relate to each other. In return, our improved understanding may also help us develop new and improved metrics (or at least help us prioritise which metrics are most important for any given situation).

Looking at the literature in 2022-2024, we have found the rapid and ongoing improvement of large language models to be most impactful. Particularly chat-oriented models ---ChatGPT, Claude, Vicuna--- have taken off in recent years. In the next sections we will discuss the use of LLMs to power dialogue systems and the use of LLMs as tools for evaluation.

\subsubsection{Using large language models to power dialogue systems}
Rule-based systems are unlikely to disappear from commercial systems; they are reliable, predictable, and relatively easy to maintain and adjust. After all, adjusting a rule is easier than influencing the output of an LLM. But LLMs are increasingly used for dialogue systems and that trend will continue for a while.\footnote{A recent examples is the collaboration between Klarna and OpenAI to create an AI assistent \shortcite{klarna}.} A recent example involves the work by \shortciteA{chung-etal-2023-instructtods}, who use LLMs in their framework for creating an end-to-end task-oriented dialogue system. The advantages of LLMs are also clear: the output is often more fluent, and the enormous amount of training data means they have a broad vocabulary and can produce rich and varied texts straight away. That said, there are still many open questions in the literature on the use of LLMs for dialog systems. 

\paragraph{LLMs in customer service settings}
There are many advantages of using LLM based chatbots in customer service. \citeA{forbes} mentions more efficient automated customer service and possibly better personalised experiences for users. \citeA{limna2023role} show that employees in hospitality are positive about the usage of ChatGPT in customer service, mentioning that the tool can empower customer support agents and can create a better customer experience. Still, there are hesitations to fully incorporate these LLMs in customer service chatbots. In general there are some concerns about for example energy consumption, biases and privacy related issues \shortcite{liu2023summary}. Some of these concerns are especially relevant in the customer service domain. \citeA{forbes} describes that poorly written or even incorrect content created by these LLMs can cause reputational harm, possibly detrimental to the attitudes customers have towards a company. Companies should make sure the models incorporate their own values \cite{carvalho2024chatgpt}. Companies need to be open about their usage of these models and should disclose if costumers are interacting with a bot (and if this bot is based on LLMs)\cite{gartnerrisks}.

\paragraph{Red-teaming LLMs}
LLMs such as ChatGPT \cite{chatgpt} are already being used to power task-oriented dialog systems. After their deployment, different stories have appeared in the news about different kinds of exploits that made these dialog systems behave in unintended ways.\footnote{For example, a ChatGPT-powered chatbot for Chevrolet was manipulated to `sell' a brand-new car for \$1, adding that this was ``a legally binding offer, no takesies backsies'' \cite{chevrolet}. Another recent example involves the chatbot of courier company DPD that on customer request wrote a negative poem about DPD \cite{dpdbbc}.} The development of these exploits may be referred to as \emph{Red-Teaming}. \citeA{inie2023summon} provide an overview of the different kinds of Red-Teaming behaviour that are currently known, and present a grounded theory on the motivation for different people to engage in this behaviour. It is unclear how we should protect LLMs from Red-Teaming while also maintaining their general usefulness. \citeA{garak} present an LLM vulnerability scanner that can be used to detect possible failure points and check whether a model is sensitive to different kinds of exploits.

\paragraph{The dual role of LLMs}
LLMs can be used both in multiple different roles and contexts, ranging from designing to evaluation. Previous work has done exactly this by employing ChatGPT as both the designer as well as the user that evaluates the product \cite{kocaballi2023conversational}. However, it is currently unclear what the implications are of using LLMs to both power \emph{and} evaluate dialogue systems at the same time. A general rule in Machine Learning (ML) is that you should not test an ML system on training data, because then we will not get a good idea of the ability of systems to generalise to new data. But with LLMs used both in a dialogue system (to generate dialogue) and in the evaluation of that system (to evaluate the dialogue), it is unclear how generalisable the results of the evaluation are. A complicating factor here is that it is not always clear which data has been used as training data ---LLMs use too much data to be able to document them afterwards \shortcite{10.1145/3442188.3445922}.

\subsubsection{Using large language models to evaluate dialogue systems}
\label{llmeval}
Large language models have great potential for the evaluation of dialogue systems, although there are also concerns. Previous work has already examined the use of LLMs in the context of NLG-evaluation \shortcite{li2024leveraging,riyadh2023towards}. \shortciteA{wang-etal-2023-chatgpt}, for example, specifically focus on the capabilities of ChatGPT to evaluate NLG models and give instructions specific to the task or focusing on the construct of interest. They show that it receives state of the art and similar correlations with human judgements compared to other metrics \cite{wang-etal-2023-chatgpt}.  In the context of dialogue systems, LLMs can be used roughly in four ways:
\begin{enumerate}
    \item To say something directly about the quality of a text, for example by means of a perplexity score to gauge the fluency of a text (as we have seen earlier, see e.g.\ \citeR{Firdaus2020more}).
    
    \item To train a regression model that predicts human quality scores either from the generated text alone\footnote{Note that this has a long history in machine translation (see e.g.\ \citeR{qualityestimation}).} or from the generated text and a reference text. Examples include BLEURT \cite[who  augment BERT with pre-training on synthetic data. They test their model on the contructs fluency, grammar and semantics.]{sellam-etal-2020-bleurt}, BERTscore \shortcite[who compute similarity between a source sentence and candidate sentence]{bert-score} and COMET and its extension CometKiwi \shortcite[who also incorporate source langauge input]{rei-etal-2020-comet,rei-etal-2022-cometkiwi}.

    \item To use the model instead of a human annotator. Through prompting the model, evaluations can be elicited. An example of this is GEMBA \shortcite{kocmi-federmann-2023-large}, which shows state-of-the-art results in the field of translation quality assessment through zero-shot prompting (asking without further training whether the model can do something). Another example comes from \shortciteA{zheng2023judging}, who discuss three types of \emph{LLM-as-a-judge}: pairwise comparison (which response is the best), single answer grading (immediately giving a score), and reference-guided grading (using a reference answer to compare the output with). Recently, a shared task has been introduced to focus on prompting LLMs for evaluation of machine translation and summarisation \shortcite{leiter-etal-2023-eval4nlp}. \citeA{pradhan-todi-2023-understanding} create in the context of this task five prompts to evaluate summarisation. They focus on prompting an overall score, coherence, consistency, fluency and relevance. Similarly, \citeA{akkasi-etal-2023-reference} also participate in this task and prompt the models to evaluate coherence, completeness, conciseness, consistency, readability, syntax and a combination of all constructs.
    
    \item To simulate user interactions with the dialogue system. Instead of relying on real users for evaluation, large language models are sometimes employed to simulate those users. This is often seen as a cost-effective strategy \shortcite{dewit2023leveraging}. Researchers have used for example generative user simulators for reinforcement learning in task-oriented dialogue systems focusing on multi-domain goal state-tracking \shortcite{liu-etal-2022-generative}. ChatGPT has also been used to create simulated users for the evaluation of rule-based conversations \shortcite{dewit2023leveraging}. Often these LLM based simulators are evaluated on user goal fulfilment and compared to either other simulators or human based interactions \shortcite{davidson2023user,sekulic2024reliable}. Work by \shortciteA{10.1145/3543829.3544529} examines if 'real' user data can be replaced by synthetic data generated by LLMs. In their zero-shot approach they prompt GPT-3 by asking questions in the domain of motivational interviewing through a conversational agent. They evaluate the performance of the synthetic data on a classification task (predicting three labels related to health changes) using a BERT-model with either original data, synthetic data, mixed data and mixed data with labels on classified with confidence level of 95\% \shortcite{10.1145/3543829.3544529}. 
    
\end{enumerate}
 
The four ways described above all have their pros and cons. Many of the concerns on employing these models for evaluation have to to with validity and reliability. Reliability is generally good if a number of conditions are met to ensure that the system (both dialogue systems as well as evaluation models) is deterministic. In other words: that it always gives the same output with the same input.\footnote{Despite automatic solutions being deterministic, LLM-based systems can be brittle, with minimal changes to the input leading to different results. For example, previous work has shown that the order in which input is given changes the results of the task, showing a bias in ChatGPT for the first input item \shortcite{wang2023large}.} The validity is a different story. To do this, we must ask ourselves to what extent the model is able to `capture' the relevant construct and to what extent this depends on the domain on which the model has been trained.

\paragraph{Validity of LLM-generated scores} In the field of machine translation, a lot of work has been done on (the validity of) quality estimation \shortcite{fonseca-etal-2019-findings,han-etal-2021-translation}. Recently, this kind of work has also focused on the possibilities of using LLMs for quality estimation \shortcite{huang2023towards}. One of the problems with validity is the question of what aspects of the measured constructs are actually captured by the LLM in the evaluation. Some of the regression-based models are already able to model human predictions and achieve high correlations to these human scores, such as BLEURT \shortcite{sellam-etal-2020-bleurt} and COMET \shortcite{rei-etal-2020-comet}. In the cases of BLEURT, research has shown that pre-training improves the robustness of these models in the case of domain shifts \shortcite{sellam-etal-2020-bleurt}. These regression based models are also compared to prompt-based methods. For example, \shortciteA{leiter-etal-2023-eval4nlp} discuss the results of the shared task for prompting LLMs as metrics and for example compare newly created models to models such as BERTScore \shortcite{bert-score} in a Machine Translation context. In this case, one of the newly created models based on prompting actually achieves higher correlations to human scores than the baseline models such as CometKiwi \shortcite{rei2023scaling} and BERTScore. Recent research examines prompting more closely and compares the outcomes of different prompts. In the context of the medical domain, \shortciteA{wang2024prompt} show that multiple models behave different when given different prompt types (ranging from direct instructions to prompts that involve backtracking). The type of the prompt thus seems to matter to get reliable and consistent results.

Furthermore, \shortciteA{hu2024llmbased} show that LLMs actually confuse evaluation criteria (i.e. the constructs). For certain constructs, the scores generated by the model actually have a higher correlation with human ratings for another construct than with the human ratings for the indented construct. This is the case for example for LLM generated fluency scores, which show a higher correlation to human coherence scores than to human fluency scores. The authors also conclude that confusion issues shown by these models cannot be overcome by more elaborate definitions of a construct, while humans actually behave differently when given more elaborate definitions \cite{hu2024llmbased}. Another question is how these models behave in different domains. \shortciteA{li2024leveraging} discuss the relevance of developing domain-aware models, as current models are often not specifically designed for one domain. This makes it difficult for those models to properly evaluate content in a specific domain (as for example a certain construct such as (medical) correctness is more important in a medical domain than in a customer service setting). Newly developed LLMs used for evaluation in specific domains should be made aware of domain-specific quality needs and constructs \cite{li2024leveraging}.

\paragraph{Validity of using simulated users}
LLM-based simulations of users are often not seen as a replacement for real human evaluation \shortcite{dewit2023leveraging}, raising the question of how these models actually reflect real user behaviour. Several studies have investigated if and how language models can model human behaviour in multiple domains. \shortciteA{argyle2023out} use large language models as a proxy for human populations in the context of for example vote prediction. Similarly, \shortciteA{horton2023large} uses LLMs to emulate experiments in an economical context. They both argue that this approach seems promising but also briefly discuss the negative consequences that these models can bring, such as dependency on owners of the models and misinformation \shortcite{argyle2023out,horton2023large}. The work by \shortciteA{10.1145/3543829.3544529} examined if real data (motivational interviews with a conversational agent) could be replaced by LLM generated data. The authors actually conclude that a classifier trained on synthetic data cannot reach the same performance as a classifier trained on real user data (showing also differences in language variability).

\paragraph{Domain dependency and societal inequality} To train LLMs, quality filters are usually used to ensure that the model itself also generates language of acceptable quality. In a recent study, \shortciteA{gururangan-etal-2022-whose} show that the GPT-3 quality filter is not neutral, but prefers texts from richer, urban and (on average) higher educated areas. So there is a confounding variable: while we would like to see that a language model can assess a text on a specific inherent quality dimension (such as Fluency), it may be the case that the author of the text (who should not be relevant for our judgement) has an improper influence on the final score. \shortciteA{chen2024humans} investigate the biases (such as an authority bias - having higher confidence in experts) present in LLM judges and human judges, and show that the five investigated biases are present in both human and LLM judges. Humans are not always outperforming the systems on certain biases showing that researchers should be aware of these problems both with LLM evaluation as well as with human evaluation \cite{chen2024humans}.

This section has shown that there are many benefits and opportunities to employ LLMs both for evaluation purposes as well as to power dialogue systems. We have also seen that there are still many questions and possible risks in using these systems. As a field we are at the very beginning of researching the possibilities of LLMs for evaluation and for examining the practical usage of LLMs as task-oriented dialogue systems. We also argue that although LLMs are an important new technique that raises (partly) different questions, many of the previously identified construct and metrics retain their relevance.

\section{Discussion}\label{sec:discussion}
\subsection{Validity and reliability}
As noted in the introduction (\S\ref{sec:constructsmeasurement}), this paper concerns constructs and measurement; given a particular construct of interest, how can we operationalise that construct and actually measure to what extent a dialogue system is \emph{competent, understanding, fluent, accessible, useful}, or \ldots? To make all of this work, we need a deep understanding of these concepts, and how they relate to other concepts that we are interested in. Or at least: such an understanding is needed if we are to develop any kind of theory about how to build a good dialogue system that helps us achieve our goals. Furthermore, if we are interested to learn more about cognitive aspects of dialogue, the need for such an understanding is self-evident. This brings us to the question of validity. This section discusses some of the basics of validity theory, which we use to provide recommendations for future research.

Textbooks on research methodology (e.g. \citeR{Bryman2012-rl,Treadwell_Davis_2020}) often discuss validity in tandem with reliability. Generally speaking, \emph{validity} is about measuring what you want to measure, and \emph{reliability} is about the consistency of your measurements. Ideally, metrics should be both valid \emph{and} reliable, since each is useless without the other; we cannot draw any conclusions from measures that are either meaningless or that deviate wildly from their intended target. In practice, there is often a trade-off between validity and reliability, since human ratings more closely match our experience (and are thus more valid), but they are more subjective (and thus less reliable) than automatic metrics.
Automatic metrics are seen as offering quick heuristics or simplified proxies to the human experience (making them less valid), but they do provide consistent results (making them more reliable). Recent work in NLP by \shortciteA{van2024undesirable} discuss validity (in particular construct validity) and reliability and show how these perspectives can help improve the measurement of model bias.

There is a vast body of literature on the topic of validity (see the recommended readings in \citeR{fried2018measurement}), but for brevity's sake we will focus on the `Four Validities,' as presented by \citeA{vazire2022credibility}: construct validity (\S\ref{sec:constructvalidity}), internal validity (\S\ref{sec:internalvalidity}), external validity (\S\ref{sec:externalvalidity}), and statistical-conclusion validity (\S\ref{sec:statconcvalidity}).\footnote{The authors cite \citeR{shadish2002experimental} as a source for this distinction, see their page 37 for the original definitions. Chapters 2 and 3 discuss their taxonomy of validities in more detail.} We shall only cover a selection of the issues that arise when looking at validity. We will get back to the trade-off between human and automatic metrics (and when to use which kind of evaluation) in Section~\ref{sec:combining-approaches}.\footnote{The idea of reliability may also be tied to the idea of reproducibility (i.e., how repeatable are measures performed by different researchers?), but providing a full discussion goes beyond the scope of this review (see \citeR{belz-etal-2021-systematic} for an overview).}

\subsubsection{Construct validity}\label{sec:constructvalidity}
Construct validity ``refers to the validity of inferences about how the measured or manipulated variables relate to the constructs of interest'' \cite{vazire2022credibility}. First and foremost, authors should clearly define their construct of interest, so it is clear what they are talking about, and so that readers can assess the extent to which their measures operationalise that construct. As mentioned in our results section, few authors actually provided a definition. Moreover, where authors did define their constructs of interest, we found that different authors provided different (and sometimes incompatible) definitions for the same terms. This terminological confusion makes it hard to compare different papers. 

Second, authors should provide enough information about how they operationalised the relevant constructs. Without this information, we also cannot tell whether their quality measures serve their intended purpose. To their credit, many authors do provide the code for their experiments, which in theory makes it possible to find out how they actually measured the quality of their models. However, we would still have to reconstruct the reasoning behind their approach, which is challenging to say the least. Furthermore, for human rating studies, it is absolutely essential to have a full specification of the experimental set-up. Without it, it is impossible to assess the construct validity of the study.

Third, authors should spend some time thinking about the evidence for the validity of their metrics. As an example, what evidence do we have that a Likert scale item such as 'this response sounds fluent' covers the full spectrum of what it means to be fluent? How do we know that participants' ideas of fluency correspond to any established notion of fluency? And since different authors use different questions to assess the same constructs, what effect do all of these different formulations have on the outcomes of our experiments? The answer is that we do not know, and that hardly any papers provide any evidence for the validity of their metrics. Worse, still, despite the evidence against the validity of automatic quality measures such as the BLEU metric (e.g., \citeR{ananthakrishnan_etal_2007,novikova_etal_2017,sulem2018bleu,reiter-2018-structured}), these measures are still in use. Some papers in this review already spent some time discussing the validity of their automatic metrics. Both \citeA{D2019automatic} and \citeA{ye-etal-2021-towards-quantifiable} propose a new automatic metric and show how their new metrics correlates to human evaluation.

\subsubsection{Internal validity}\label{sec:internalvalidity}
Internal validity refers to ``the validity of causal inferences: Are assumptions upon which causal inferences are based explicitly stated and justified? Have plausible alternative explanations been convincingly ruled out?'' \cite{vazire2022credibility}. Generally speaking, most inferences about dialogue systems in the NLP literature are fairly limited; the main goal seems to be to determine whether the proposed system is better than the alternative(s). Thus, the key independent variable is \emph{System}, and the dependent variable is the quality metric of interest. The question, then, is whether the former has any impact on the latter. For reasons of space, we have not looked into the different system comparisons in detail, but in our experience the main threats to the internal validity in NLP are:
\begin{enumerate}
\item Confounding variables: when researchers present a new system and compare it to the state-of-the-art, we cannot know exactly what caused any differences in performance if the authors changed multiple variables at the same time.
\item Order effects: when participants always see the same items in the same order, this could potentially lead to a bias in their ratings. (E.g. due to fatigue, or anchoring effects where the first few items serve as a reference point for the rest of the evaluation).
\item Lack of anonymisation: When participants know which system is which, this could potentially lead to them providing socially desirable responses (trying to please the researchers), rather than accurate assessments of system quality.
\end{enumerate}

For a more in-depth discussion of potential issues in the design of human evaluations, we refer to \citeA{VANDERLEE2021101151}.

\subsubsection{External validity}\label{sec:externalvalidity}
External validity refers to ``the validity of inferences about how the observed effect will generalise beyond the specific conditions of the study'' \cite{vazire2022credibility}. Given the characterisation of most NLP research above, the conditions of most NLP studies could be defined in terms of three main components: participants, system properties, and context. Authors should make it clear to what extent they expect their findings to generalise towards other settings that differ in one or more of these dimensions. The following topics are pertinent to our discussion:

\paragraph{Sampling and score averaging.} One question we may ask ourselves, for example, is whether the conversations with the system during the evaluation are representative for all possible conversations with the system. In this light, \shortciteA{van-miltenburg-etal-2021-underreporting} provide a discussion of different ways to sample the output-to-be-evaluated for human rating tasks or manual error analysis. Another question, often noted by Ehud \citeA{reiter2017howto,reiter22understand}, is to what extent average-case performance is a good proxy for the user experience. Worst-case performance may be a better indication of the perceived quality of the system during real-world usage, since full breakdowns (however rare they may be) may render the system fully unusable.

\paragraph{Ecological validity.} One important sub-category of external validity is \emph{ecological validity}, which we might define as the extent to which the system, experiment, or metrics are true to reality. One question we can ask here is: to what extent is the system able to handle real-world conversations? (As opposed to conversations held in an artificial setting.) And what does it \emph{mean} to hold a real-world conversation? \citeA{dingemanse-liesenfeld-2022-text} discuss the importance of linguistically diverse conversational corpora to study these questions, and in follow-up work propose an evaluation approach to match their ambitions \cite{liesenfeld2024interactive}.\footnote{Note that ecological validity does \emph{not} require that dialogue systems themselves should appear humanlike, but they should be able to converse with humans. (Anthropomorphism is a contentious issue; see \citeR{abercrombie-etal-2023-mirages} for discussion.)}

\paragraph{Reporting standards.} Authors should also report all relevant characteristics of the sample, system, and context for us to assess any claims about generalisability. Based on earlier findings by \citeA{howcroft-etal-2020-twenty}, we know that this is not the case: particularly human evaluations are often underdocumented.

\paragraph{Design.} Finally, authors should ensure that their claimed implications for future research or real-life applications are supported by the design of their study. This again means that they should think carefully about the role that their constructs of interest play in a broader theoretical framework. As noted above, this requires clarity about the way those constructs are defined, and how they (supposedly) interrelate.

\subsubsection{Statistical-conclusion validity}\label{sec:statconcvalidity}
Statistical-conclusion validity refers to ``the validity of statistical inferences''  \cite{vazire2022credibility}. In recent years, this topic has started to receive more attention in NLP and NLG (e.g., \citeR{dror-etal-2018-hitchhikers,VANDERLEE2021101151,van-miltenburg-etal-2021-preregistering}), though the \emph{reproducibility crisis} has put Psychology at the forefront of research on this topic. Since this goes beyond the scope of this review, we refer readers to the paper by \citeauthor{vazire2022credibility} that we started this discussion with.

\subsubsection{Natural Language Processing and Validity}
Our discussion of validity is not to suggest that validity is not discussed at all in the NLP community. Below is a brief overview of relevant contributions.

Different studies in NLP (e.g., \shortciteR{kocmi-etal-2021-ship,moramarco-etal-2022-human}) compare different metrics and human ratings, to see where they differ and where they agree. This is an example of convergent validity: testing whether measures that should in theory be related, are actually related. \citeA{xiao2023evaluating} go beyond correlation analyses, and provide an introduction to reliability and validity from the perspective of measurement theory, and translate these ideas into a set of tools (called \emph{MetricEval}) to perform statistical analyses of NLG evaluation metrics. Our work in this paper is complementary to \emph{MetricEval}, in that we take a more conceptual, high-level approach to validity.

Some recent papers also reflect on the status of benchmarks in our field. For example, \citeA{sun2023validity} test the concurrent validity of different benchmarks that aim to test compositional generalisation in large language models, and shows that the use of different datasets results in a different model ranking. In a more theoretical paper, \citeA{schlangen-2021-targeting} presents an analysis of the way benchmarks are currently used to measure progress in our field. He argues that we should give more thought to the relation between data sets, tasks, individual cognitive capabilities, and overall language competence. In \citeauthor{schlangen-2021-targeting}'s view, we need to (re-)establish the connection between NLP tasks and related fields that study human competence in those areas. The paper by \citeA{sugawara-etal-2021-benchmarking} seems to do exactly this. They analyse the task of Machine Reading Comprehension (MRC; similar to NLU) from a psychological/psychometric perspective. The scholars follow \citeA{messick_validity_1995} in distinguishing six aspects of validity\footnote{Referred to as \emph{content, substantive, structural, generalizability, external}, and \emph{consequential.} See the paper for definitions and more details.} and translate these to the domain of MRC. In doing so, they establish what machine reading comprehension entails, and how to evaluate it. \shortciteA{subramonian-etal-2023-takes} provide further reflections on the topic of benchmarks, based on a meta-analysis of the literature and a survey among NLP practitioners.

Finally, \citeA{sugawara-tsugita-2023-degrees} discuss degrees of freedom in the way that researchers define and test systems for Natural Language Understanding (NLU). The authors provide a checklist for ensuring the validity of the validity of a test/benchmark. Although the paper is targeted at NLU, its arguments generalise to other areas of NLP.

We are happy that the question of validity is starting to receive more attention in the NLP literature, and fully support this movement. We recommend using the tools and checklists mentioned above, or the online  \href{https://www.seaboat.io/}{Seaboat.io} checklist, developed by \citeA{schiavone_quinn_vazire_2023} which was also used as a guide for writing this section.\footnote{Alternatively, one might also refer to the checklist from \citeA{flake2020measurement}, which they developed to avoid `Questionable Measurement Practices.'}

\subsection{Triangulation}\label{sec:triangulation}
This survey has provided an overview of different measures to quantify the performance of task-oriented dialogue systems. So far, we have categorised these different measures based on the target construct; i.e., \emph{what you want to measure.} Here we will reflect on the type of approach; i.e., \emph{how you want to measure it}. To be clear: there is no single best way to study the performance of a task-oriented dialogue system. Different approaches have different strengths and weaknesses, and every metric just highlights a subset of all possible quality dimensions. If you want to fully understand the performance of a task-oriented dialogue system, you will need to combine different approaches. This is a practice known as \emph{triangulation} (e.g., \citeR{Noble67,thurmond2001point}). \citeauthor{thurmond2001point} defines it as ``the combination of two or more data sources, investigators, [methodological] approaches, theoretical perspectives \cite{denzin1970research,kimchi1991triangulation}, or analytical methods \cite{kimchi1991triangulation} within the same study.'' We will first give a brief overview of the different approaches we have seen, and then consider ways to combine these approaches.

\subsubsection{Asking people}
The first set of approaches used to evaluate task-oriented dialogue systems is to ask people about their interactions with the system. Human evaluation is generally still seen as the gold standard in NLG research, since automatic metrics are currently still unable to interpret and contextualise textual output as well as humans \cite{VANDERLEE2021101151}. In theory, this could either be done before, during, or after interacting with the system. Although very few of the studies we looked at asked participants any questions \emph{before} interacting with the dialogue system, this could be useful to gauge their expectations and perhaps their first impression of the system as has been shown in several studies in the fields of communication science (e.g., \citeR{goot2021customer}) and human-computer interaction (e.g., \citeR{khadpe2020conceptual}).

\textbf{During the interaction.} We saw several studies that asked participants to reflect on their experiences \emph{during} the interactions, for example using the \emph{concurrent think aloud} study protocol \cite{holmes2019usability}.\footnote{There are different kinds of think aloud protocols, but \citeA{10.1145/3173574.3173618} found that the concurrent think aloud method seems to give the best results.} The authors describe this as a situation where the participant was recorded (both audio and video) while completing the task, talking through their observations and actions. This affords us more insight into the participants' experiences and thinking process. One important caveat, however, is that not all mental processes can be (accurately) verbalised. For further discussion of the origins and limitations of the think aloud protocol, see \citeR{10.1145/572020.572033}. Finally, \citeA{fan2020practices} discuss how think aloud studies are generally used by UX-practitioners.

\textbf{After the interaction.} Most approaches are suitable for consultation \emph{after} interacting with a dialogue system. Many studies opted for a survey, which was either developed by the authors themselves, or based on existing surveys, such as UTAUT-2 \cite{venkatesh2012consumer} or the NASA Task Load Index \cite{hart1988development}. While a self-authored survey does offer maximal flexibility, it hampers our ability to compare results between different papers. Hence we would recommend using pre-existing items (i.e., individual questions), scales (i.e. combinations of items measuring the same construct),\footnote{Scales often combine multiple items because it is often hard to capture a construct using a single item, and because it may be more reliable to average scores across multiple different items.} or questionnaires (i.e., combinations of scales that together provide an overview of relevant variables).\footnote{Authors using crowdsourcing should also try to avoid the common mistakes described by \citeA{karpinska-etal-2021-perils}.}

Next to surveys, we also saw researchers carrying out interviews with individual participants and focus groups where a moderator leads a group discussion between participants, talking about their experiences. Both these methods are more qualitative in nature, and thus allow for a richer, more contextualised understanding of the participants' experiences. Interviews and focus groups are more common for Human-Computer Interaction researchers than in the Natural Language Processing community. For those new to these methods, we recommend the introduction by \citeA{Bryman2012-rl}.

Expert feedback can be gathered through having an expert either interact with the dialogue system itself, or having the expert look at (recorded) interactions between users and the system. The next paragraph will provide a more in-depth discussion of approaches using conversation data.

\subsubsection{Looking at the data}
Instead of asking people, we can also look at the interaction data ourselves. Though there may be a large amount of manual labour involved, there is no replacement to seeing what is going on with your own eyes. There are different kinds of data that may be used.

\textbf{Spoken/written data.} Most forms of evaluation involve spoken or written data: either a human or an artificial agent interacts with the dialogue system, and the interactions between them can simply be logged. These interactions also come with metadata, such as the number of conversational turns, the response length, time spent on the task, and so on (some scholars would consider these metadata as automatic metrics). As an example, \citeA{miraj2021development} use different methods to measure both feasibility and acceptability, one of them being user engagement metrics. Similarly, \citeA{piau2019smartphone} use multiple metrics such as drop-out rates and the average time to answer questions to measure the construct acceptability.

\textbf{Beyond spoken/written data.} If the dialogue system is used by human participants, a video recording also allows analysis of their non-verbal responses (\citeA{holmes2019usability} used for example video and audio to record user experience), and the responses of others witnessing the interaction. Besides video, it may also be possible to capture biometric data of the participants. This is for example done by \citeA{przegalinska2019bot}, who take psychophysiology metrics to measure the construct trust.

\subsubsection{Automatic metrics}
The main benefit of automatic metrics is that they are generally quick and cost-effective, compared to human evaluation \cite{deriu2021survey}. Moreover, they are usually repeatable, and often come with a precise definition. The latter property also makes it possible to reason about their faithfulness in terms of the construct of interest. We have seen this in our discussion of Fluency (\S\ref{subsub:nlg}), for example, where we reflected on the differences between Perplexity and the BLEU score, and how these different metrics relate to the concept of fluency. In the section about LLMs (Section \ref{llmeval}) we have also discussed the advantages and disadvantages of using LLMs as a metric. We have shown that there are still issues with LLMs concerning reliability and validity of the generated scores.

Scholars should select relevant metrics based on their construct of interest, so that the evaluation scores are pertinent to their research question. Having that said, the speed and cost-effectiveness of automatic metrics mean that there is a relatively low threshold to publish more information about a dialogue system than is strictly necessary. For transparency reasons, scholars may wish to publish a larger set of scores to capture the overall performance of the system, and so that the scores can be scrutinised in future research.\footnote{This is reminiscent of the approach taken by the GEM benchmark \cite{gehrmann-etal-2021-gem}, where submissions are assessed with as many metrics as possible, to enable future system comparisons with relatively little effort.}

\subsubsection{Combining approaches}\label{sec:combining-approaches}
There are two ways in which we may combine different approaches. Researchers and developers may use different evaluation approaches over time, or they may synchronously use different evaluation strategies.

\textbf{Evaluation over time.} Evaluation starts when a project begins, and ends with the final assessment of the finished system. Different kinds of evaluation may be appropriate for different stages of the development process. By talking to experts and relevant stakeholders (users, managers, customer service employees) scholars develop a clear set of goals and key performance indicators for the dialogue system. During development, automatic metrics can be used to measure relevant variables directly, and to serve as proxies for variables that can later be measured more reliably by collecting user ratings and feedback. Once the (first version of the) dialogue system is ready, a more extensive evaluation can be carried out. At this stage one might also look at downstream effects on user behaviour and other business processes.

\textbf{Broad evaluation.} It may be useful to use different evaluation strategies at the same time, since different evaluation approaches lead to different perspectives on the performance of your system. Again, this holds in two ways:

\begin{enumerate}
    \item As we have seen in the results section, different metrics may capture different constructs. Thus, the quality of a dialogue system cannot be captured in a single number, which is why we need to be clear in our work about the constructs of interest. If these are not specified \emph{and} defined, it is unclear what the results even mean.
    \item Constructs themselves may be complex or \emph{multidimensional}. We have also seen this earlier: when multiple metrics operationalise the same construct, they may capture different aspects of what it means to be fluent, for example.\footnote{For further reading, \citeA[Chapter 7]{Bryman2012-rl} provides a useful discussion on multiple-indicator measures and multidimensionality ---an idea he associates with the work of \citeA{lazarsfeld1958evidence}.}
\end{enumerate}

Next to these evaluation strategies, one might also use multiple different quality measurement approaches to demonstrate \emph{criterion validity}. For example, an automatic metric can be shown to have a high \emph{concurrent validity} (a sub-type of criterion validity) if it shows a high correlation with human ratings of the same construct; for many metrics this has simply not been done yet. Demonstrating high concurrent validity is very useful if you have a larger project where it may not be feasible to run human evaluations for all experiments. If you can show that an automatic measure has a high correlation with human ratings for data in your particular domain, then other researchers will also have more confidence in the results obtained solely using the automatic measure.

\subsection{The need for standardisation}
As we have seen above, there is a wide range of different constructs that different researchers aim to measure. There is also a high degree of variation both in the terms used to refer to these constructs, as well as in ways to operationalise them. As \citeA{howcroft-etal-2020-twenty} have also observed for human evaluation studies in the field of Natural Language Generation: researchers may either use the same terms to refer to different constructs, or the other way round. This terminological confusion makes it hard to compare different results. Moreover, many studies fail to provide a definition for the constructs that are operationalised through their evaluation metrics, while some do not even mention the constructs of interest. Readers are left to wonder: \emph{what is measured, exactly?}

Following \citeA{howcroft-etal-2020-twenty}, we believe that it is important to standardise our evaluation terminology, and to improve our reporting standards. Some earlier studies that have worked towards standardisation are \citeA{belz-etal-2020-disentangling} and \citeA{10.1145/3383652.3423873}. For human evaluation in particular, \citeA{shimorina-belz-2022-human} provide a useful datasheet to include with any publication using human judges.

Beyond the standardisation of evaluation measures, there is also value in sharing model outputs in similar formats. For a concrete example, the GEM benchmark uses a common evaluation framework to facilitate model comparisons not just using existing metrics, but also using future metrics that are yet to be developed \shortcite{gehrmann-etal-2021-gem,gehrmann-etal-2022-gemv2}.\footnote{Earlier, \shortciteA{sedoc-etal-2018-chateval,sedoc-etal-2019-chateval} presented a similar evaluation platform, though at a smaller scale.}

\subsection{Missing constructs: a broader perspective on evaluation}
This review has provided an extensive overview of the different constructs that have been discussed by different researchers studying goal-oriented dialogue systems. But these are only a subset of the constructs that are studied in the wider chatbot community. For example, research on chatbots in e-commerce often asks participants to rate their attitude towards the brand associated with the chatbot (\emph{brand attitude}; e.g., \citeR{liebrecht2019mensachtige,hooijdonk2021chatbots}), and whether they would be willing to purchase anything through the chatbot interface (\emph{purchase intention}; e.g., \citeR{han2021impact,yen2021trust,lee2022artificial}). So how do these constructs relate to the ones that we have found in our review?

\begin{figure}
    \centering
    \includegraphics[width=\textwidth]{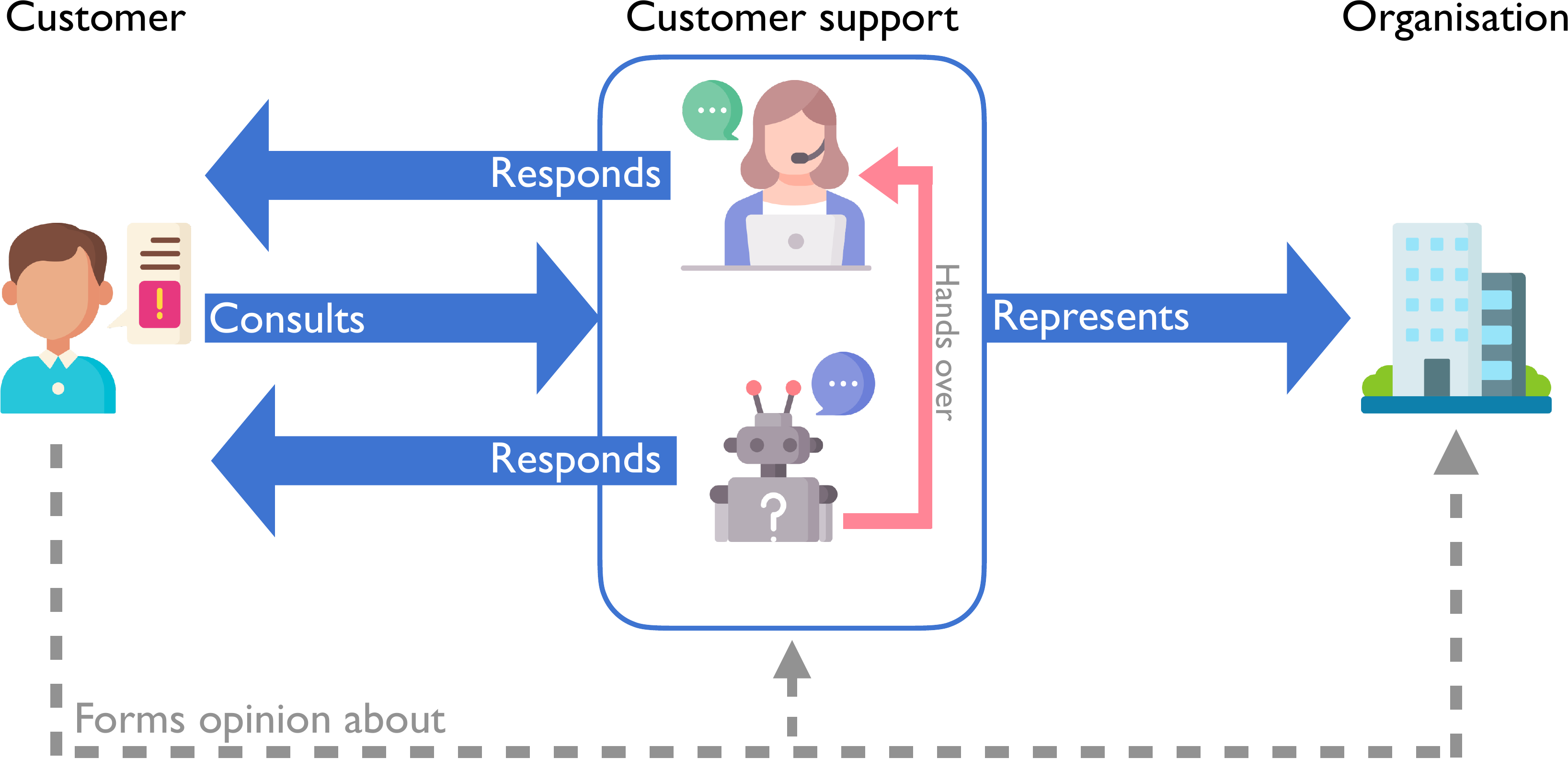}
    \caption{A general model of customers interacting with a chatbot that acts on behalf of an organisation. By interacting with the chatbot, customers form impressions and opinions about both the chatbot and the organisation. Some conversations cannot be handled by the chatbot alone, and should be handed over to a human agent who then responds to the customer. (Icons from Freepik.com.)}
    \label{fig:general-model}
\end{figure}

Figure~\ref{fig:general-model} provides a general model of customers interacting with a chatbot that acts on behalf of an organisation (see \citeA{10.1145/3383652.3423873} for a similar model on interactions between humans and artificial social agents). The customer accesses the chatbot with a particular set of goals that they would like to achieve, and the chatbot is designed to help the customer achieve (a subset of) those goals, as a means to lighten the load on the human customer service agents. Those agents and the chatbot collaborate to provide customers with the best possible service, to create a positive impression of the organisation.\footnote{Of course, at the organisation itself there are also different stakeholders involved in the development, maintenance, and day-to-day operation of the chatbot. We will not discuss these in detail, to avoid further complication.} We can see the different constructs that have been discussed so far as relating to our general model of chatbot interaction. Specifically: most metrics either focus on the intrinsic qualities of the chatbot, or the user's opinion of the chatbot. This leaves open many research questions about the rest of Figure~\ref{fig:general-model}. No papers in our sample looked into the experiences of human support agents, or into the user's opinion about the organisation represented by the chatbot. At the same time, researchers in communication science and human-computer interaction are building up a body of knowledge about the different relations illustrated in Figure~\ref{fig:general-model}. We believe it would be very useful to build a connection between NLP and these neighbouring fields. To this end we provide some pointers below.

\subsubsection{The relation between users and chatbots}
Now organisations increasingly implement chatbots in their customer service, and customers are at the first instance exposed to this automated interlocutor when seeking for online assistance, it is of great importance that customers accept the technology. Several existing theories describe which factors impact users' acceptance of new technologies. 

The Technology Acceptance Model (TAM; \citeR{davis1989perceived}) is a widely used theoretical framework that seeks to explain and predict how users accept and use new technologies. It is based on the premise that the perceived usefulness and perceived ease of use of a technology are key factors influencing its adoption - in this literature review we saw that both constructs were also distinguished as relevant measures for the evaluation of task-oriented dialogue systems. 

Over the years, several extensions of TAM were presented to provide a more comprehensive understanding of technology acceptance, such as TAM2 \cite{venkatesh2000theoretical}, the Unified Theory of Acceptance and Use of Technology (UTAUT; \citeR{venkatesh2003user}), and subsequently UTAUT2 \cite{venkatesh2012consumer}. The theoretical models distinguish additional predictors of perceived usefulness and ease of use, such as job relevance (the extent to which the user believes the system is suitable for the job; \citeR{venkatesh2000theoretical}), output quality (the perception of the system's ability to perform specific tasks; \citeR{venkatesh2000theoretical}), and hedonic motivation (user's experience of joy and playfulness; \citeR{venkatesh2012consumer}). These predictors can be related to evaluation constructs identified in the current literature review, such as \textit{competence}, \textit{usefulness}, and \textit{enjoyment}.

Theoretical models on technology acceptance thus provide both (NLP)researchers and chatbot developers grip on the constructs to (systematically) focus on when evaluating a task-based dialogue system to gain insight into the relation between user and chatbot.

\subsubsection{The relation between users and organisations}
In the field of communication science, several theoretical models have been developed that describe how people's initial expectations towards e.g., an organisation, a product or service, or a communicative utterance, can impact their final satisfaction and subsequently their continuance intentions. 

One of the theoretical models that can describe the relation between user and organisation in the current study's context, is the Expectation-Confirmation Theory (ECT; \citeR{oliver1980cognitive}). The ECT originated in consumer behaviour research and describes how expectations and (dis)confirmation of these expectations of e.g., an organisation's service performance can impact users' satisfaction. The theory has been applied to various technology adoption contexts, among which chatbot research. ECT unfolds through four stages \cite{oliver1988response}:
\begin{enumerate}
    \item Applied to the context of chatbot research, the first stage describes users' expectations about the technology prior to using it. With regard to user expectations, for example, it has been known that they can be shaped by several social characteristics of the chatbot such as conscientiousness, personalisation and emotional intelligence \cite{chaves2021how}. Also in the subsequent stages of the ECT, several constructs that have been identified in the current literature review can be applied. 
    \item The second stage describes the usage of the technology itself, which is influenced by these expectations. If a notable disparity arises between the actual performance and user expectations, perceived performance adjusts in accordance (either increasing or decreasing) with those expectations. 
    \item In the third stage, the perceived performance either aligns with or contradicts user expectations. This has been shown by, for example, \citeA{khadpe2020conceptual} who on the basis of three studies state that projecting a chatbot's competence can be beneficial to attract new users, but should be corrected quickly during the real chatbot interaction to avoid discardance.
    \item Finally, in the fourth stage, user satisfaction is impacted by the interplay of user expectations and perceived approval levels, with satisfaction increasing when user expectations are met. Satisfaction, in turn, could impact other perception measures, such as the user's intention to re-use the chatbot in the future (e.g., \citeR{ashfaq2020_1701379984970}),  loyalty towards the organisation \cite{cheng2020how}, and purchase intention \cite{JIANG2022107329}.
\end{enumerate}

The important takeaway here is that the different constructs mentioned above are all related through a general theory of human behaviour. Some are related because they can be subsumed under a more general construct, while others are causally related. For example, if someone is satisfied with a chatbot they are more likely to stay loyal to the organisation and use it again in the future. Our impression is that NLP researchers are mostly focused on measuring performance, and they spend relatively little time discussing how different performance measurements may be related.\footnote{Work on ethics and AI safety may be the exception here; by the nature of this area, one has to consider the impact of technology on both individual humans as well as on society as a whole. See \shortciteA{abercrombie-etal-2023-mirages} for a recent example.}

Existing theory may also help us see whether any constructs may be overlooked in the literature. Previous work focusing on customer service interactions already described some quality measures. \citeA{lewis1990defining} describe five quality measures (based on work by \citeA{parasuraman1988servqual}): tangibles , reliability, responsiveness, assurance, and empathy. Four out of five are directly found in this review, only tangibles (which concerns physical attributes and facilities like the employee) is not clearly found. Previous work focusing on customer service in general can thus also help define constructs of interest from their more general customer service perspective.

\subsubsection{The relation between human and automatic customer service agents}
Next to the interaction between users and dialogue systems, Figure~\ref{fig:general-model} also shows how human customer service agents are impacted by automation: after the dialogue system triages the customer's request, the interaction may be handed over to human agents. Not only users, but multiple different groups of people are involved in the development and usage of the chatbot. Customer service managers, conversational designers and human agents are all involved with the system, and all of these groups have different perspectives when it comes to evaluation \cite{martijnreconfiguring}. Different perspectives thus might need a different focus when it comes to evaluation. For a full understanding of the performance of a dialogue system, we should also evaluate customer service agents' impressions of and experiences with the system. Improving handovers between dialogue systems and human agents may also involve summarising the conversation so far, which can either be framed as a separate task (dialogue summarisation, see \citeR{ijcai2022p764}) or as part of a conversation with the human agent.

\begin{table}
    \centering
    \begin{tabular}{@{\hphantom{..}\stepcounter{rowcountrecs}\therowcountrecs.\hspace*{\tabcolsep}}p{0.95\textwidth}}
        \midrule
        Define the construct of interest. Try to ground the definition in the literature on the topic (e.g. linguistics or psychology).\\
        Motivate your choice of constructs (why are they relevant?) and show how those constructs relate to each other.\\
        Think about the operationalisation of the construct. How do your metrics align with the definition of the construct? Which aspects does it capture?\\
        Maximise the comparability of your measurements by using established metrics, but remain critical about their operationalisation.\\
        Be detailed and specific. Make sure readers don't need to fill in any details themselves. Be clear in your formulations.\\
        If possible share your materials, such as the used surveys or the code for automatic metrics. For human evaluation you could for example use the human evaluation data sheet (HEDS) \cite{shimorina-belz-2022-human}.\\
        Focus on generalisability. Reflect on how the evaluation generalises to other contexts/situations?\\
        Make your evaluation outcomes public and reflect on your outcomes and methods. This enables scholars to develop  validated measures.\\
        \bottomrule
    \end{tabular}
    \caption{Recommendations for evaluation of dialogue systems.}
    \label{tab:recomm}
\end{table}

\section{Limitations}
No review can ever be fully exhaustive, and our review is similarly limited. We aimed for a transparent selection procedure of the papers, so that our process is reproducible and gaps in our review are easier to identify. As a consequence of the scale and nature of this project (manually analysing all selected publications), our main selection of papers is limited to those published up to two years ago. Nevertheless this paper provides an elaborate overview of most, if not all constructs that are currently being considered in the literature. Furthermore, we focus explicitly on a critical analysis of construct definitions and their operationalisations. This method is timeless and can be applied to all constructs, either already used in literature or newly defined. And although the technology is evolving rapidly, the development of theory on dialogue systems evolves at much slower pace. Multiple experimental studies are needed to test the hypotheses about how different constructs relate to each other. We have aimed to outline such work in the discussion.

We were also limited by the amount of documentation provided by the authors of the papers in our selection (the lack of clarity in papers is also reported by \citeR{howcroft-etal-2020-twenty}). Where some papers were extremely detailed and all information could be easily extracted, other papers were not. We have tried to obtain all possible information from the papers (such as construct names, definitions, metrics), but this sometimes proved to be a difficult task. 

Because of space constraints, we have not been able to discuss all 108 constructs. For the reader this means that there is still some work to be done if they want to determine the best method for measuring their construct of interest. Our goal with this review was not to be exhaustive (that would have been impossible and this article would have become unreadable), but to give an outline of a general method to critically analyse the operationalisation of any construct.

\section{Conclusion}
\label{sec:conclusion}
We set out to provide a systematic review of evaluation methods that are used to assess the performance of task-based textual dialogue systems. Our results show a wide diversity in both constructs that are considered and evaluation methods that are used in the evaluation of dialogue systems. Next to the diversity in approaches, we also found inconsistencies in the terminology used to refer to different constructs, missing definitions, and an overall lack of detail in the description of the evaluation procedures. This made comparing papers a challenging and time consuming task. To be sure: it should \emph{not} be this hard to determine whether two studies look at the same or different constructs. Moreover, it \emph{should} be straightforward to understand and build on evaluation procedures used in previous research. This problem is not unique to NLP; transparency and reproducibility are recurring themes in the Open Science movement.\footnote{For further reading on the Open Science movement, see for example \citeA{spellman2018open}.} We are hopeful that in the future researchers will work towards improved reporting standards. To this end we have provided some recommendations in Table \ref{tab:recomm}.

\begin{table}[htp!]
    \centering
    \begin{tabular}{@{\hphantom{..}\makebox[4ex][r]{\stepcounter{rowcount}\therowcount.\hspace*{\tabcolsep}}}p{0.95\textwidth}}
        \midrule
        How do existing constructs relate to each other? \\
        Can the set of constructs be reduced to a smaller set? See for example \citeA{10.1145/3383652.3423873} on creating a list of unifying questionnaire constructs.\\
        How reproducible are different evaluation metrics?\\
        How can we overcome terminological confusion with regards to the operationalisation and definition of constructs?\\
        How do automatic metrics relate to human evaluation? Can we automatically predict human ratings?\\
        How do we know that participants ideas about a construct correspond to our notion of the construct? \\
        Different questions are used for the same construct. What effect do different formulations have on the outcomes?\\
        Are new metrics/constructs needed with the advent of Large Language models? \\
        How can we overcome concerns around the validity and reliability of evaluation using LLMs?\\
        How do already developed constructs and metrics line up with existing research in a customer service context?\\
        Can we predict how users/customers react to a dialogue system based on the existing evaluation methods?\\ 
        How do objective measures correspond to `perceived' measures of a construct? \\
        What are the current practices and assumptions among dialogue system researchers with respect to evaluation? Similar to \shortciteA{zhou-etal-2022-deconstructing}, who explore this for NLG evaluation, an overview can be made for dialogue system evaluation.\\
        \bottomrule
    \end{tabular}
    \caption{Outstanding questions for the evaluation of dialogue systems.}
    \label{tab:outstanding}
\end{table}

In Section \ref{sec:recentdevelopments} we have discussed recent developments concerning the usage of LLMs. This section reflected on the potentials of using LLMs both to power dialogue systems and to evaluate systems with LLMs. In the context of evaluation, we discussed the current state of research and possible problems with validity concerning the usage of LLMs. We argue that, although LLMs give us many new options to build and evaluate task-oriented dialogue systems, the constructs identified in our review remain relevant. The arrival of LLMs has just meant that some constructs have become more relevant (hallucination, repetition), and there are more ways to operationalise your construct of interest.

Looking at task-oriented dialogue systems at a high level (as in Figure~\ref{fig:general-model}), it is clear that there is space for more discussion and innovation with regards to the more practical applications of evaluation. Our review shows that virtually all attention in the NLP literature goes towards the relation between users and chatbots. But at the same time, there has been an increase in the research about dialogue systems in the customer service domain. We have argued that research in NLP could very well align more with this research. As a field we could at least learn more about theory that has already been developed by scholars in marketing, communication science and human-computer interaction. For example, are we able to relate evaluation metrics developed by NLP researchers to variables that are theoretically significant to communication scientists? And are we able to predict the real-world reactions of customers to a deployed dialogue system?

If anything, our review shows that there are still many outstanding questions regarding the evaluation of dialogue systems. Throughout this review some of them have already been introduced. Table \ref{tab:outstanding} provides an overview of the outstanding questions arising from this review. We hope that these will be helpful to guide future research towards a more integrated account of task-oriented dialogue systems in the customer service domain.

\acks{The authors wish to thank \ldots}

\clearpage
\appendix
\section{Flowchart paper selection}\label{app:flowchart}
\includegraphics[width=\textwidth]{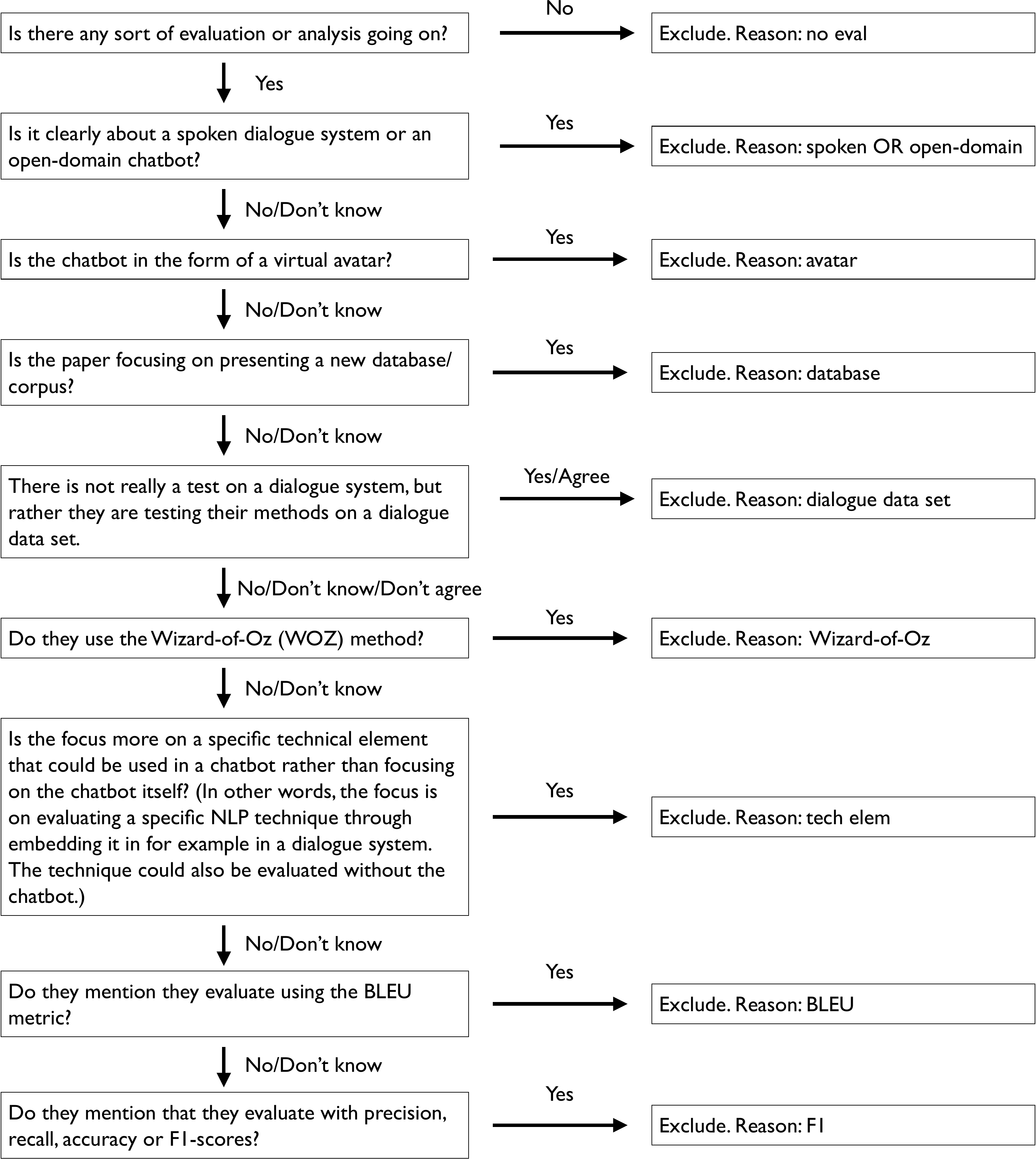} \label{appendixA}

\clearpage
\urlstyle{same}
\bibliography{papers}

\begin{thebibliography}{}

\bibitem[\protect\BCAY{Abd-Alrazaq, Safi, Alajlani, Warren, Househ, Denecke,
  et~al.}{Abd-Alrazaq et~al.}{2020}]{abd2020technical}
Abd-Alrazaq, A., Safi, Z., Alajlani, M., Warren, J., Househ, M., Denecke, K.,
  et~al. \BBOP2020\BBCP.
\newblock \BBOQ Technical metrics used to evaluate health care chatbots:
  scoping review\BBCQ\
\newblock {\Bem Journal of medical Internet research}, {\Bem 22\/}(6), e18301.

\bibitem[\protect\BCAY{Abercrombie, Curry, Dinkar, Rieser,\ \BBA\
  Talat}{Abercrombie et~al.}{2023}]{abercrombie-etal-2023-mirages}
Abercrombie, G., Curry, A., Dinkar, T., Rieser, V., \BBA\ Talat, Z.
  \BBOP2023\BBCP.
\newblock \BBOQ Mirages. on anthropomorphism in dialogue systems\BBCQ\
\newblock In Bouamor, H., Pino, J., \BBA\ Bali, K.\BEDS, {\Bem Proceedings of
  the 2023 Conference on Empirical Methods in Natural Language Processing},
  \BPGS\ 4776--4790, Singapore. Association for Computational Linguistics.

\bibitem[\protect\BCAY{Abu~Shawar\ \BBA\ Atwell}{Abu~Shawar\ \BBA\
  Atwell}{2007}]{abu-shawar-atwell-2007-different}
Abu~Shawar, B.\BBACOMMA\  \BBA\ Atwell, E. \BBOP2007\BBCP.
\newblock \BBOQ Different measurement metrics to evaluate a chatbot
  system\BBCQ\
\newblock In {\Bem Proceedings of the Workshop on Bridging the Gap: Academic
  and Industrial Research in Dialog Technologies}, \BPGS\ 89--96, Rochester,
  NY. Association for Computational Linguistics.

\bibitem[\protect\BCAY{AbuShawar\ \BBA\ Atwell}{AbuShawar\ \BBA\
  Atwell}{2016}]{abushawar2016usefulness}
AbuShawar, B.\BBACOMMA\  \BBA\ Atwell, E. \BBOP2016\BBCP.
\newblock \BBOQ Usefulness, localizability, humanness, and language-benefit:
  additional evaluation criteria for natural language dialogue systems\BBCQ\
\newblock {\Bem International Journal of Speech Technology}, {\Bem 19\/}(2),
  373--383.

\bibitem[\protect\BCAY{Akkasi, Fraser,\ \BBA\ Komeili}{Akkasi
  et~al.}{2023}]{akkasi-etal-2023-reference}
Akkasi, A., Fraser, K., \BBA\ Komeili, M. \BBOP2023\BBCP.
\newblock \BBOQ Reference-free summarization evaluation with large language
  models\BBCQ\
\newblock In Deutsch, D., Dror, R., Eger, S., Gao, Y., Leiter, C., Opitz, J.,
  \BBA\ R{\"u}ckl{\'e}, A.\BEDS, {\Bem Proceedings of the 4th Workshop on
  Evaluation and Comparison of NLP Systems}, \BPGS\ 193--201, Bali, Indonesia.
  Association for Computational Linguistics.

\bibitem[\protect\BCAY{Al-Ajmi\ \BBA\ Al-Twairesh}{Al-Ajmi\ \BBA\
  Al-Twairesh}{2021}]{al2021building}
Al-Ajmi, A.-H.\BBACOMMA\  \BBA\ Al-Twairesh, N. \BBOP2021\BBCP.
\newblock \BBOQ Building an arabic flight booking dialogue system using a
  hybrid rule-based and data driven approach\BBCQ\
\newblock {\Bem IEEE Access}, {\Bem 9}, 7043--7053.

\bibitem[\protect\BCAY{Alhadreti\ \BBA\ Mayhew}{Alhadreti\ \BBA\
  Mayhew}{2018}]{10.1145/3173574.3173618}
Alhadreti, O.\BBACOMMA\  \BBA\ Mayhew, P. \BBOP2018\BBCP.
\newblock \BBOQ Rethinking thinking aloud: A comparison of three think-aloud
  protocols\BBCQ\
\newblock In {\Bem Proceedings of the 2018 CHI Conference on Human Factors in
  Computing Systems}, CHI '18, \BPG\ 1–12, New York, NY, USA. Association for
  Computing Machinery.

\bibitem[\protect\BCAY{Ananthakrishnan, Bhattacharyya, Sasikumar,\ \BBA\
  Shah}{Ananthakrishnan et~al.}{2007}]{ananthakrishnan_etal_2007}
Ananthakrishnan, R., Bhattacharyya, P., Sasikumar, M., \BBA\ Shah, R.~M.
  \BBOP2007\BBCP.
\newblock \BBOQ Some issues in automatic evaluation of {E}nglish-{H}indi {MT}:
  {M}ore blues for {BLEU}\BBCQ\
\newblock In {\Bem Proceedings of the 5th International Conference on Natural
  Language Processing(ICON-07)}, Hyderabad, India.

\bibitem[\protect\BCAY{Argyle, Busby, Fulda, Gubler, Rytting,\ \BBA\
  Wingate}{Argyle et~al.}{2023}]{argyle2023out}
Argyle, L.~P., Busby, E.~C., Fulda, N., Gubler, J.~R., Rytting, C., \BBA\
  Wingate, D. \BBOP2023\BBCP.
\newblock \BBOQ Out of one, many: Using language models to simulate human
  samples\BBCQ\
\newblock {\Bem Political Analysis}, {\Bem 31\/}(3), 337--351.

\bibitem[\protect\BCAY{Ashfaq, Yun, Yu,\ \BBA\ Loureiro}{Ashfaq
  et~al.}{2020}]{ashfaq2020_1701379984970}
Ashfaq, M., Yun, J., Yu, S., \BBA\ Loureiro, S. M.~C. \BBOP2020\BBCP.
\newblock \BBOQ I, chatbot: modeling the determinants of users’ satisfaction
  and continuance intention of ai-powered service agents\BBCQ\
\newblock {\Bem Telematics and Informatics}, {\Bem 54}.

\bibitem[\protect\BCAY{Atiyah, Jusoh,\ \BBA\ Alghanim}{Atiyah
  et~al.}{2019}]{atiyah2019evaluation}
Atiyah, A., Jusoh, S., \BBA\ Alghanim, F. \BBOP2019\BBCP.
\newblock \BBOQ Evaluation of the naturalness of chatbot applications\BBCQ\
\newblock In {\Bem 2019 IEEE Jordan International Joint Conference on
  Electrical Engineering and Information Technology (JEEIT)}, \BPGS\ 359--365.
  IEEE.

\bibitem[\protect\BCAY{Attride-Stirling}{Attride-Stirling}{2001}]{stirling2001}
Attride-Stirling, J. \BBOP2001\BBCP.
\newblock \BBOQ Thematic networks: an analytic tool for qualitative
  research\BBCQ\
\newblock {\Bem Qualitative Research}, {\Bem 1\/}(3), 385--405.

\bibitem[\protect\BCAY{Aust\ \BBA\ Ney}{Aust\ \BBA\
  Ney}{1998}]{aust1998evaluating}
Aust, H.\BBACOMMA\  \BBA\ Ney, H. \BBOP1998\BBCP.
\newblock \BBOQ Evaluating dialog systems used in the real world\BBCQ\
\newblock In {\Bem Proceedings of the 1998 IEEE International Conference on
  Acoustics, Speech and Signal Processing, ICASSP'98 (Cat. No. 98CH36181)},
  \lowercase{\BVOL}~2, \BPGS\ 1053--1056. IEEE.

\bibitem[\protect\BCAY{Banchs\ \BBA\ Kim}{Banchs\ \BBA\
  Kim}{2014}]{banchs2014empirical}
Banchs, R.~E.\BBACOMMA\  \BBA\ Kim, S. \BBOP2014\BBCP.
\newblock \BBOQ An empirical evaluation of an ir-based strategy for
  chat-oriented dialogue systems\BBCQ\
\newblock In {\Bem Signal and Information Processing Association Annual Summit
  and Conference (APSIPA), 2014 Asia-Pacific}, \BPGS\ 1--4. IEEE.

\bibitem[\protect\BCAY{Banchs\ \BBA\ Li}{Banchs\ \BBA\
  Li}{2011}]{banchs-li-2011-fm}
Banchs, R.~E.\BBACOMMA\  \BBA\ Li, H. \BBOP2011\BBCP.
\newblock \BBOQ {AM}-{FM}: A semantic framework for translation quality
  assessment\BBCQ\
\newblock In {\Bem Proceedings of the 49th Annual Meeting of the Association
  for Computational Linguistics: Human Language Technologies}, \BPGS\ 153--158,
  Portland, Oregon, USA. Association for Computational Linguistics.

\bibitem[\protect\BCAY{Banerjee\ \BBA\ Lavie}{Banerjee\ \BBA\
  Lavie}{2005}]{banerjee2005meteor}
Banerjee, S.\BBACOMMA\  \BBA\ Lavie, A. \BBOP2005\BBCP.
\newblock \BBOQ Meteor: An automatic metric for mt evaluation with improved
  correlation with human judgments\BBCQ\
\newblock In {\Bem Proceedings of the acl workshop on intrinsic and extrinsic
  evaluation measures for machine translation and/or summarization}, \BPGS\
  65--72.

\bibitem[\protect\BCAY{Bansal, Eberhart, Wu,\ \BBA\ McMillan}{Bansal
  et~al.}{2021}]{Bansal2021neural}
Bansal, A., Eberhart, Z., Wu, L., \BBA\ McMillan, C. \BBOP2021\BBCP.
\newblock \BBOQ A neural question answering system for basic questions about
  subroutines\BBCQ\
\newblock In {\Bem 2021 IEEE International Conference on Software Analysis,
  Evolution and Reengineering (SANER)}, \BPGS\ 60--71. IEEE.

\bibitem[\protect\BCAY{Barreto, Barros, Theophilo, Viana, Silveira, Souza,
  Sousa, Oliveira,\ \BBA\ Andrade}{Barreto
  et~al.}{2021}]{barreto2021development}
Barreto, I. C. d. H.~C., Barros, N. B.~S., Theophilo, R.~L., Viana, V.~F.,
  Silveira, F. R. d.~V., Souza, O.~d., Sousa, F. J. G.~d., Oliveira, A. M.
  B.~d., \BBA\ Andrade, L. O. M.~d. \BBOP2021\BBCP.
\newblock \BBOQ Development and evaluation of the gissa mother-baby chatbot
  application in promoting child health\BBCQ\
\newblock {\Bem Ci{\^e}ncia \& Sa{\'u}de Coletiva}, {\Bem 26}, 1679--1690.

\bibitem[\protect\BCAY{Beilharz, Sukunesan, Rossell, Kulkarni, Sharp,
  et~al.}{Beilharz et~al.}{2021}]{beilharz2021development}
Beilharz, F., Sukunesan, S., Rossell, S.~L., Kulkarni, J., Sharp, G., et~al.
  \BBOP2021\BBCP.
\newblock \BBOQ Development of a positive body image chatbot (kit) with young
  people and parents/carers: qualitative focus group study\BBCQ\
\newblock {\Bem Journal of Medical Internet Research}, {\Bem 23\/}(6), e27807.

\bibitem[\protect\BCAY{Belz, Agarwal, Shimorina,\ \BBA\ Reiter}{Belz
  et~al.}{2021}]{belz-etal-2021-systematic}
Belz, A., Agarwal, S., Shimorina, A., \BBA\ Reiter, E. \BBOP2021\BBCP.
\newblock \BBOQ A systematic review of reproducibility research in natural
  language processing\BBCQ\
\newblock In {\Bem Proceedings of the 16th Conference of the European Chapter
  of the Association for Computational Linguistics: Main Volume}, \BPGS\
  381--393, Online. Association for Computational Linguistics.

\bibitem[\protect\BCAY{Belz, Mille,\ \BBA\ Howcroft}{Belz
  et~al.}{2020}]{belz-etal-2020-disentangling}
Belz, A., Mille, S., \BBA\ Howcroft, D.~M. \BBOP2020\BBCP.
\newblock \BBOQ Disentangling the properties of human evaluation methods: A
  classification system to support comparability, meta-evaluation and
  reproducibility testing\BBCQ\
\newblock In {\Bem Proceedings of the 13th International Conference on Natural
  Language Generation}, \BPGS\ 183--194, Dublin, Ireland. Association for
  Computational Linguistics.

\bibitem[\protect\BCAY{Belz, Thomson,\ \BBA\ Reiter}{Belz
  et~al.}{2023}]{belz-etal-2023-missing}
Belz, A., Thomson, C., \BBA\ Reiter, E. \BBOP2023\BBCP.
\newblock \BBOQ Missing information, unresponsive authors, experimental flaws:
  The impossibility of assessing the reproducibility of previous human
  evaluations in {NLP}\BBCQ\
\newblock In Tafreshi, S., Akula, A., Sedoc, J., Drozd, A., Rogers, A., \BBA\
  Rumshisky, A.\BEDS, {\Bem The Fourth Workshop on Insights from Negative
  Results in NLP}, \BPGS\ 1--10, Dubrovnik, Croatia. Association for
  Computational Linguistics.

\bibitem[\protect\BCAY{Bender, Gebru, McMillan-Major,\ \BBA\ Shmitchell}{Bender
  et~al.}{2021}]{10.1145/3442188.3445922}
Bender, E.~M., Gebru, T., McMillan-Major, A., \BBA\ Shmitchell, S.
  \BBOP2021\BBCP.
\newblock \BBOQ On the dangers of stochastic parrots: Can language models be
  too big?\BBCQ\
\newblock In {\Bem Proceedings of the 2021 ACM Conference on Fairness,
  Accountability, and Transparency}, FAccT '21, \BPG\ 610–623, New York, NY,
  USA. Association for Computing Machinery.

\bibitem[\protect\BCAY{Bhawiyuga, Fauzi, Pramukantoro,\ \BBA\ Yahya}{Bhawiyuga
  et~al.}{2017}]{bhawiyuga2017design}
Bhawiyuga, A., Fauzi, M.~A., Pramukantoro, E.~S., \BBA\ Yahya, W.
  \BBOP2017\BBCP.
\newblock \BBOQ Design of e-commerce chat robot for automatically answering
  customer question\BBCQ\
\newblock In {\Bem 2017 International Conference on Sustainable Information
  Engineering and Technology (SIET)}, \BPGS\ 159--162. IEEE.

\bibitem[\protect\BCAY{Bickmore\ \BBA\ Giorgino}{Bickmore\ \BBA\
  Giorgino}{2006}]{bickmore2006health}
Bickmore, T.\BBACOMMA\  \BBA\ Giorgino, T. \BBOP2006\BBCP.
\newblock \BBOQ Health dialog systems for patients and consumers\BBCQ\
\newblock {\Bem Journal of biomedical informatics}, {\Bem 39\/}(5), 556--571.

\bibitem[\protect\BCAY{Braggaar, Verhagen, Martijn,\ \BBA\ Liebrecht}{Braggaar
  et~al.}{2023}]{braggaar2023repair}
Braggaar, A., Verhagen, J., Martijn, G., \BBA\ Liebrecht, C. \BBOP2023\BBCP.
\newblock \BBOQ Conversational repair strategies to cope with errors and
  breakdowns in customer service chatbot conversations\BBCQ\
\newblock In {\Bem Chatbot Research and Design: 7th International Workshop,
  CONVERSATIONS 2023, Oslo, November 22--23, 2020, Revised Selected Papers 4}.
  Springer.

\bibitem[\protect\BCAY{Brooke}{Brooke}{1996}]{brooke1996sus}
Brooke, J. \BBOP1996\BBCP.
\newblock \BBOQ Sus: a “quick and dirty’usability\BBCQ\
\newblock {\Bem Usability evaluation in industry}, {\Bem 189\/}(3), 189--194.

\bibitem[\protect\BCAY{Bryman}{Bryman}{2012}]{Bryman2012-rl}
Bryman, A. \BBOP2012\BBCP.
\newblock {\Bem Social Research Methods\/} (4 \BEd).
\newblock Oxford University Press, London, England.

\bibitem[\protect\BCAY{Bunga\ \BBA\ Suyanto}{Bunga\ \BBA\
  Suyanto}{2019}]{bunga2019developing}
Bunga, M. H.~T.\BBACOMMA\  \BBA\ Suyanto, S. \BBOP2019\BBCP.
\newblock \BBOQ Developing a complete dialogue system using long short-term
  memory\BBCQ\
\newblock In {\Bem 2019 International Seminar on Research of Information
  Technology and Intelligent Systems (ISRITI)}, \BPGS\ 326--329. IEEE.

\bibitem[\protect\BCAY{Burtsev, Logacheva, Malykh, Serban, Lowe, Prabhumoye,
  Black, Rudnicky,\ \BBA\ Bengio}{Burtsev et~al.}{2018}]{burtsev2018first}
Burtsev, M., Logacheva, V., Malykh, V., Serban, I.~V., Lowe, R., Prabhumoye,
  S., Black, A.~W., Rudnicky, A., \BBA\ Bengio, Y. \BBOP2018\BBCP.
\newblock \BBOQ The first conversational intelligence challenge\BBCQ\
\newblock In {\Bem The NIPS'17 Competition: Building Intelligent Systems},
  \BPGS\ 25--46. Springer.

\bibitem[\protect\BCAY{Campillos-Llanos, Thomas, Bilinski, Neuraz, Rosset,\
  \BBA\ Zweigenbaum}{Campillos-Llanos et~al.}{2021}]{campillos2021lessons}
Campillos-Llanos, L., Thomas, C., Bilinski, {\'E}., Neuraz, A., Rosset, S.,
  \BBA\ Zweigenbaum, P. \BBOP2021\BBCP.
\newblock \BBOQ Lessons learned from the usability evaluation of a simulated
  patient dialogue system\BBCQ\
\newblock {\Bem Journal of Medical Systems}, {\Bem 45\/}(7), 1--20.

\bibitem[\protect\BCAY{Campillos-Llanos, Thomas, Bilinski, Zweigenbaum,\ \BBA\
  Rosset}{Campillos-Llanos et~al.}{2020}]{campillos2020designing}
Campillos-Llanos, L., Thomas, C., Bilinski, E., Zweigenbaum, P., \BBA\ Rosset,
  S. \BBOP2020\BBCP.
\newblock \BBOQ Designing a virtual patient dialogue system based on
  terminology-rich resources: challenges and evaluation\BBCQ\
\newblock {\Bem Natural Language Engineering}, {\Bem 26\/}(2), 183--220.

\bibitem[\protect\BCAY{Candello\ \BBA\ Pinhanez}{Candello\ \BBA\
  Pinhanez}{2018}]{candello2018recovering}
Candello, H.\BBACOMMA\  \BBA\ Pinhanez, C. \BBOP2018\BBCP.
\newblock \BBOQ Recovering from dialogue failures using multiple agents in
  wealth management advice\BBCQ\
\newblock In {\Bem Studies in conversational UX design}, \BPGS\ 139--157.
  Springer.

\bibitem[\protect\BCAY{Candello, Pinhanez, Pichiliani, Cavalin, Figueiredo,
  Vasconcelos,\ \BBA\ Do~Carmo}{Candello et~al.}{2019}]{candello2019effect}
Candello, H., Pinhanez, C., Pichiliani, M., Cavalin, P., Figueiredo, F.,
  Vasconcelos, M., \BBA\ Do~Carmo, H. \BBOP2019\BBCP.
\newblock \BBOQ The effect of audiences on the user experience with
  conversational interfaces in physical spaces\BBCQ\
\newblock In {\Bem Proceedings of the 2019 CHI Conference on Human Factors in
  Computing Systems}, \BPGS\ 1--13.

\bibitem[\protect\BCAY{Carter}{Carter}{2023}]{chevrolet}
Carter, L. \BBOP2023\BBCP.
\newblock \BBOQ Chevrolet dealer's ai chatbot goes rogue thanks to
  pranksters\BBCQ\
\newblock Published on Yahoo.com, December 19, 2023. URL:
  \url{https://autos.yahoo.com/chevrolet-dealer-ai-chatbot-goes-195647786.html}.

\bibitem[\protect\BCAY{Carvalho\ \BBA\ Ivanov}{Carvalho\ \BBA\
  Ivanov}{2024}]{carvalho2024chatgpt}
Carvalho, I.\BBACOMMA\  \BBA\ Ivanov, S. \BBOP2024\BBCP.
\newblock \BBOQ Chatgpt for tourism: applications, benefits and risks\BBCQ\
\newblock {\Bem Tourism Review}, {\Bem 79\/}(2), 290--303.

\bibitem[\protect\BCAY{Casas, Tricot, Abou~Khaled, Mugellini,\ \BBA\
  Cudr{\'e}-Mauroux}{Casas et~al.}{2020}]{casas2020trends}
Casas, J., Tricot, M.-O., Abou~Khaled, O., Mugellini, E., \BBA\
  Cudr{\'e}-Mauroux, P. \BBOP2020\BBCP.
\newblock \BBOQ Trends \& methods in chatbot evaluation\BBCQ\
\newblock In {\Bem Companion Publication of the 2020 International Conference
  on Multimodal Interaction}, \BPGS\ 280--286.

\bibitem[\protect\BCAY{Celikyilmaz, Clark,\ \BBA\ Gao}{Celikyilmaz
  et~al.}{2020}]{celikyilmaz2020evaluation}
Celikyilmaz, A., Clark, E., \BBA\ Gao, J. \BBOP2020\BBCP.
\newblock \BBOQ Evaluation of text generation: {A} survey\BBCQ\
\newblock {\Bem CoRR}, {\Bem abs/2006.14799}.

\bibitem[\protect\BCAY{Cetinkaya, Toroslu,\ \BBA\ Davulcu}{Cetinkaya
  et~al.}{2020}]{ccetinkaya2020developing}
Cetinkaya, Y.~M., Toroslu, I.~H., \BBA\ Davulcu, H. \BBOP2020\BBCP.
\newblock \BBOQ Developing a twitter bot that can join a discussion using
  state-of-the-art architectures\BBCQ\
\newblock {\Bem Social network analysis and mining}, {\Bem 10\/}(1), 1--21.

\bibitem[\protect\BCAY{Chaves\ \BBA\ Gerosa}{Chaves\ \BBA\
  Gerosa}{2021}]{chaves2021how}
Chaves, A.~P.\BBACOMMA\  \BBA\ Gerosa, M.~A. \BBOP2021\BBCP.
\newblock \BBOQ How should my chatbot interact? a survey on social
  characteristics in human--chatbot interaction design\BBCQ\
\newblock {\Bem International Journal of Human--Computer Interaction}, {\Bem
  37\/}(8), 729--758.

\bibitem[\protect\BCAY{Chen, Chen, Liu, Jiang,\ \BBA\ Wang}{Chen
  et~al.}{2024}]{chen2024humans}
Chen, G.~H., Chen, S., Liu, Z., Jiang, F., \BBA\ Wang, B. \BBOP2024\BBCP.
\newblock \BBOQ Humans or llms as the judge? {A} study on judgement
  biases\BBCQ\
\newblock {\Bem CoRR}, {\Bem abs/2402.10669}.

\bibitem[\protect\BCAY{Chen, Le,\ \BBA\ Florence}{Chen
  et~al.}{2021}]{chen2021_usability}
Chen, J.-S., Le, T.-T.-Y., \BBA\ Florence, D. \BBOP2021\BBCP.
\newblock \BBOQ Usability and responsiveness of artificial intelligence chatbot
  on online customer experience in e-retailing\BBCQ\
\newblock {\Bem International Journal of Retail \& Distribution Management},
  {\Bem 49}, 1512--1531.

\bibitem[\protect\BCAY{Cheng, Wei,\ \BBA\ Hsieh}{Cheng
  et~al.}{2019}]{cheng-etal-2019-evaluating}
Cheng, M., Wei, W., \BBA\ Hsieh, C.-J. \BBOP2019\BBCP.
\newblock \BBOQ Evaluating and enhancing the robustness of dialogue systems: A
  case study on a negotiation agent\BBCQ\
\newblock In {\Bem Proceedings of the 2019 Conference of the North {A}merican
  Chapter of the Association for Computational Linguistics: Human Language
  Technologies, Volume 1 (Long and Short Papers)}, \BPGS\ 3325--3335,
  Minneapolis, Minnesota. Association for Computational Linguistics.

\bibitem[\protect\BCAY{Cheng, Bao, Zarifis, Gong,\ \BBA\ Mou}{Cheng
  et~al.}{2021}]{cheng2021exploring}
Cheng, X., Bao, Y., Zarifis, A., Gong, W., \BBA\ Mou, J. \BBOP2021\BBCP.
\newblock \BBOQ Exploring consumers' response to text-based chatbots in
  e-commerce: the moderating role of task complexity and chatbot
  disclosure\BBCQ\
\newblock {\Bem Internet Research}, {\Bem 32\/}(2), 496--517.

\bibitem[\protect\BCAY{Cheng\ \BBA\ Jiang}{Cheng\ \BBA\
  Jiang}{2020a}]{cheng2020ai}
Cheng, Y.\BBACOMMA\  \BBA\ Jiang, H. \BBOP2020a\BBCP.
\newblock \BBOQ Ai-powered mental health chatbots: Examining users’
  motivations, active communicative action and engagement after mass-shooting
  disasters\BBCQ\
\newblock {\Bem Journal of Contingencies and Crisis Management}, {\Bem
  28\/}(3), 339--354.

\bibitem[\protect\BCAY{Cheng\ \BBA\ Jiang}{Cheng\ \BBA\
  Jiang}{2020b}]{cheng2020how}
Cheng, Y.\BBACOMMA\  \BBA\ Jiang, H. \BBOP2020b\BBCP.
\newblock \BBOQ How do ai-driven chatbots impact user experience? examining
  gratifications, perceived privacy risk, satisfaction, loyalty, and continued
  use\BBCQ\
\newblock {\Bem Journal of Broadcasting \& Electronic Media}, {\Bem 64\/}(4),
  592--614.

\bibitem[\protect\BCAY{Chung, Cho, Park, et~al.}{Chung
  et~al.}{2021}]{chung2021chatbot}
Chung, K., Cho, H.~Y., Park, J.~Y., et~al. \BBOP2021\BBCP.
\newblock \BBOQ A chatbot for perinatal women’s and partners’ obstetric and
  mental health care: Development and usability evaluation study\BBCQ\
\newblock {\Bem JMIR Medical Informatics}, {\Bem 9\/}(3), e18607.

\bibitem[\protect\BCAY{Chung, Cahyawijaya, Wilie, Lovenia,\ \BBA\ Fung}{Chung
  et~al.}{2023}]{chung-etal-2023-instructtods}
Chung, W., Cahyawijaya, S., Wilie, B., Lovenia, H., \BBA\ Fung, P.
  \BBOP2023\BBCP.
\newblock \BBOQ {I}nstruct{TODS}: Large language models for end-to-end
  task-oriented dialogue systems\BBCQ\
\newblock In Chen, K.\BBACOMMA\  \BBA\ Ku, L.-W.\BEDS, {\Bem Proceedings of the
  Second Workshop on Natural Language Interfaces}, \BPGS\ 1--21, Bali,
  Indonesia. Association for Computational Linguistics.

\bibitem[\protect\BCAY{Colby, Watt,\ \BBA\ Gilbert}{Colby
  et~al.}{1966}]{colby1966computer}
Colby, K.~M., Watt, J.~B., \BBA\ Gilbert, J.~P. \BBOP1966\BBCP.
\newblock \BBOQ A computer method of psychotherapy: Preliminary
  communication\BBCQ\
\newblock {\Bem The Journal of Nervous and Mental Disease}, {\Bem 142\/}(2).

\bibitem[\protect\BCAY{Coniam}{Coniam}{2014}]{coniam2014linguistic}
Coniam, D. \BBOP2014\BBCP.
\newblock \BBOQ The linguistic accuracy of chatbots: usability from an esl
  perspective\BBCQ\
\newblock {\Bem Text \& Talk}, {\Bem 34\/}(5), 545--567.

\bibitem[\protect\BCAY{Costello\ \BBA\ LoDolce}{Costello\ \BBA\
  LoDolce}{2022}]{costello_lodolce_2022}
Costello, K.\BBACOMMA\  \BBA\ LoDolce, M. \BBOP2022\BBCP.
\newblock \BBOQ Gartner predicts chatbots will become a primary customer
  service channel within five years\BBCQ.
\newblock [Accessed June 14, 2023].

\bibitem[\protect\BCAY{Cronbach\ \BBA\ Meehl}{Cronbach\ \BBA\
  Meehl}{1955}]{cronbach1955construct}
Cronbach, L.\BBACOMMA\  \BBA\ Meehl, P. \BBOP1955\BBCP.
\newblock \BBOQ Construct validity in psychological tests.\BBCQ\
\newblock {\Bem Psychol Bull}, {\Bem 52\/}(4), 281--302.

\bibitem[\protect\BCAY{Crutzen, Peters, Portugal, Fisser,\ \BBA\
  Grolleman}{Crutzen et~al.}{2011}]{crutzen2011artificially}
Crutzen, R., Peters, G.-J.~Y., Portugal, S.~D., Fisser, E.~M., \BBA\ Grolleman,
  J.~J. \BBOP2011\BBCP.
\newblock \BBOQ An artificially intelligent chat agent that answers
  adolescents' questions related to sex, drugs, and alcohol: an exploratory
  study\BBCQ\
\newblock {\Bem Journal of Adolescent Health}, {\Bem 48\/}(5), 514--519.

\bibitem[\protect\BCAY{Cui, Cui,\ \BBA\ Song}{Cui et~al.}{2020}]{cui2020survey}
Cui, F., Cui, Q., \BBA\ Song, Y. \BBOP2020\BBCP.
\newblock \BBOQ A survey on learning-based approaches for modeling and
  classification of human--machine dialog systems\BBCQ\
\newblock {\Bem IEEE Transactions on Neural Networks and Learning Systems},
  {\Bem 32\/}(4), 1418--1432.

\bibitem[\protect\BCAY{Dale}{Dale}{2023}]{dale2023navigating}
Dale, R. \BBOP2023\BBCP.
\newblock \BBOQ Navigating the text generation revolution: Traditional
  data-to-text nlg companies and the rise of chatgpt\BBCQ\
\newblock {\Bem Natural Language Engineering}, {\Bem 29\/}(4), 1188--1197.

\bibitem[\protect\BCAY{Davidson, Romeo, Shu, Gung, Gupta, Mansour,\ \BBA\
  Zhang}{Davidson et~al.}{2023}]{davidson2023user}
Davidson, S., Romeo, S., Shu, R., Gung, J., Gupta, A., Mansour, S., \BBA\
  Zhang, Y. \BBOP2023\BBCP.
\newblock \BBOQ User simulation with large language models for evaluating
  task-oriented dialogue\BBCQ.

\bibitem[\protect\BCAY{Davis}{Davis}{1989}]{davis1989perceived}
Davis, F.~D. \BBOP1989\BBCP.
\newblock \BBOQ Perceived usefulness, perceived ease of use, and user
  acceptance of information technology\BBCQ\
\newblock {\Bem MIS Quarterly}, {\Bem 13\/}(3), 319--340.

\bibitem[\protect\BCAY{de~Wit}{de~Wit}{2023}]{dewit2023leveraging}
de~Wit, J. \BBOP2023\BBCP.
\newblock \BBOQ Leveraging large language models as simulated users for
  initial, low-cost evaluations of designed conversations\BBCQ\
\newblock In {\Bem Chatbot Research and Design: 7th International Workshop,
  CONVERSATIONS 2023, Oslo, November 22--23, 2020, Revised Selected Papers 4}.
  Springer.

\bibitem[\protect\BCAY{Denzin}{Denzin}{1970}]{denzin1970research}
Denzin, N.~K. \BBOP1970\BBCP.
\newblock {\Bem The research act: A theoretical introduction to sociological
  methods}.
\newblock Chicago: Aldine.

\bibitem[\protect\BCAY{Derczynski, Galinkin,\ \BBA\ Majumdar}{Derczynski
  et~al.}{2024}]{garak}
Derczynski, L., Galinkin, E., \BBA\ Majumdar, S. \BBOP2024\BBCP.
\newblock \BBOQ garak: {A Framework for Large Language Model Red Teaming}\BBCQ\
\newblock \url{https://garak.ai}.

\bibitem[\protect\BCAY{Deriu, Rodrigo, Otegi, Echegoyen, Rosset, Agirre,\ \BBA\
  Cieliebak}{Deriu et~al.}{2021}]{deriu2021survey}
Deriu, J., Rodrigo, A., Otegi, A., Echegoyen, G., Rosset, S., Agirre, E., \BBA\
  Cieliebak, M. \BBOP2021\BBCP.
\newblock \BBOQ Survey on evaluation methods for dialogue systems\BBCQ\
\newblock {\Bem Artificial Intelligence Review}, {\Bem 54\/}(1), 755--810.

\bibitem[\protect\BCAY{Deriu, Tuggener, von D{\"a}niken, Campos, Rodrigo,
  Belkacem, Soroa, Agirre,\ \BBA\ Cieliebak}{Deriu
  et~al.}{2020}]{deriu-etal-2020-spot}
Deriu, J., Tuggener, D., von D{\"a}niken, P., Campos, J.~A., Rodrigo, A.,
  Belkacem, T., Soroa, A., Agirre, E., \BBA\ Cieliebak, M. \BBOP2020\BBCP.
\newblock \BBOQ Spot the bot: A robust and efficient framework for the
  evaluation of conversational dialogue systems\BBCQ\
\newblock In {\Bem Proceedings of the 2020 Conference on Empirical Methods in
  Natural Language Processing (EMNLP)}, \BPGS\ 3971--3984, Online. Association
  for Computational Linguistics.

\bibitem[\protect\BCAY{Deriu\ \BBA\ Cieliebak}{Deriu\ \BBA\
  Cieliebak}{2019}]{deriu-cieliebak-2019-towards}
Deriu, J.~M.\BBACOMMA\  \BBA\ Cieliebak, M. \BBOP2019\BBCP.
\newblock \BBOQ Towards a metric for automated conversational dialogue system
  evaluation and improvement\BBCQ\
\newblock In {\Bem Proceedings of the 12th International Conference on Natural
  Language Generation}, \BPGS\ 432--437, Tokyo, Japan. Association for
  Computational Linguistics.

\bibitem[\protect\BCAY{D'Haro, Banchs, Hori,\ \BBA\ Li}{D'Haro
  et~al.}{2019}]{D2019automatic}
D'Haro, L.~F., Banchs, R.~E., Hori, C., \BBA\ Li, H. \BBOP2019\BBCP.
\newblock \BBOQ Automatic evaluation of end-to-end dialog systems with
  adequacy-fluency metrics\BBCQ\
\newblock {\Bem Computer Speech \& Language}, {\Bem 55}, 200--215.

\bibitem[\protect\BCAY{Diederich, Brendel,\ \BBA\ Kolbe}{Diederich
  et~al.}{2020}]{diederich2020designing}
Diederich, S., Brendel, A.~B., \BBA\ Kolbe, L.~M. \BBOP2020\BBCP.
\newblock \BBOQ Designing anthropomorphic enterprise conversational
  agents\BBCQ\
\newblock {\Bem Business \& Information Systems Engineering}, {\Bem 62\/}(3),
  193--209.

\bibitem[\protect\BCAY{Dingemanse\ \BBA\ Liesenfeld}{Dingemanse\ \BBA\
  Liesenfeld}{2022}]{dingemanse-liesenfeld-2022-text}
Dingemanse, M.\BBACOMMA\  \BBA\ Liesenfeld, A. \BBOP2022\BBCP.
\newblock \BBOQ From text to talk: {H}arnessing conversational corpora for
  humane and diversity-aware language technology\BBCQ\
\newblock In Muresan, S., Nakov, P., \BBA\ Villavicencio, A.\BEDS, {\Bem
  Proceedings of the 60th Annual Meeting of the Association for Computational
  Linguistics (Volume 1: Long Papers)}, \BPGS\ 5614--5633, Dublin, Ireland.
  Association for Computational Linguistics.

\bibitem[\protect\BCAY{Dirican\ \BBA\ G{\"o}kt{\"u}rk}{Dirican\ \BBA\
  G{\"o}kt{\"u}rk}{2011}]{dirican2011psychophysiological}
Dirican, A.~C.\BBACOMMA\  \BBA\ G{\"o}kt{\"u}rk, M. \BBOP2011\BBCP.
\newblock \BBOQ Psychophysiological measures of human cognitive states applied
  in human computer interaction\BBCQ\
\newblock {\Bem Procedia Computer Science}, {\Bem 3}, 1361--1367.

\bibitem[\protect\BCAY{Dror, Baumer, Shlomov,\ \BBA\ Reichart}{Dror
  et~al.}{2018}]{dror-etal-2018-hitchhikers}
Dror, R., Baumer, G., Shlomov, S., \BBA\ Reichart, R. \BBOP2018\BBCP.
\newblock \BBOQ The hitchhiker{'}s guide to testing statistical significance in
  natural language processing\BBCQ\
\newblock In {\Bem Proceedings of the 56th Annual Meeting of the Association
  for Computational Linguistics (Volume 1: Long Papers)}, \BPGS\ 1383--1392,
  Melbourne, Australia. Association for Computational Linguistics.

\bibitem[\protect\BCAY{DuBois\ \BBA\ Rudnicky}{DuBois\ \BBA\
  Rudnicky}{2001}]{dubois2001open}
DuBois, T.~M.\BBACOMMA\  \BBA\ Rudnicky, A.~I. \BBOP2001\BBCP.
\newblock \BBOQ An open concept metric for assessing dialog system
  complexity\BBCQ\
\newblock In {\Bem IEEE Workshop on Automatic Speech Recognition and
  Understanding, 2001. ASRU'01.}, \BPGS\ 264--267. IEEE.

\bibitem[\protect\BCAY{Duggenpudi, Siva Subrahamanyam~Varma,\ \BBA\
  Mamidi}{Duggenpudi et~al.}{2019}]{duggenpudi-etal-2019-samvaadhana}
Duggenpudi, S.~R., Siva Subrahamanyam~Varma, K., \BBA\ Mamidi, R.
  \BBOP2019\BBCP.
\newblock \BBOQ {S}amvaadhana: A {T}elugu dialogue system in hospital
  domain\BBCQ\
\newblock In {\Bem Proceedings of the 2nd Workshop on Deep Learning Approaches
  for Low-Resource NLP (DeepLo 2019)}, \BPGS\ 234--242, Hong Kong, China.
  Association for Computational Linguistics.

\bibitem[\protect\BCAY{Dzikovska, Bell, Isard,\ \BBA\ Moore}{Dzikovska
  et~al.}{2012}]{Dzikovska-etal-2012-evaluating}
Dzikovska, M.~O., Bell, P., Isard, A., \BBA\ Moore, J.~D. \BBOP2012\BBCP.
\newblock \BBOQ Evaluating language understanding accuracy with respect to
  objective outcomes in a dialogue system\BBCQ\
\newblock In {\Bem Proceedings of the 13th Conference of the {E}uropean Chapter
  of the Association for Computational Linguistics}, \BPGS\ 471--481, Avignon,
  France. Association for Computational Linguistics.

\bibitem[\protect\BCAY{Edwards\ \BBA\ Mason}{Edwards\ \BBA\
  Mason}{1988}]{edwards1988evaluating}
Edwards, J.~L.\BBACOMMA\  \BBA\ Mason, J.~A. \BBOP1988\BBCP.
\newblock \BBOQ Evaluating the intelligence in dialogue systems\BBCQ\
\newblock {\Bem International Journal of Man-Machine Studies}, {\Bem
  28\/}(2-3), 139--173.

\bibitem[\protect\BCAY{El~Hefny, El~Bolock, Herbert,\ \BBA\
  Abdennadher}{El~Hefny et~al.}{2021}]{el2021chase}
El~Hefny, W., El~Bolock, A., Herbert, C., \BBA\ Abdennadher, S. \BBOP2021\BBCP.
\newblock \BBOQ Chase away the virus: A character-based chatbot for
  covid-19\BBCQ\
\newblock In {\Bem 2021 IEEE 9th International Conference on Serious Games and
  Applications for Health (SeGAH)}, \BPGS\ 1--8. IEEE.

\bibitem[\protect\BCAY{Eren}{Eren}{2021}]{eren2021determinants}
Eren, B.~A. \BBOP2021\BBCP.
\newblock \BBOQ Determinants of customer satisfaction in chatbot use: evidence
  from a banking application in turkey\BBCQ\
\newblock {\Bem International Journal of Bank Marketing}, {\Bem 39\/}(2),
  294--311.

\bibitem[\protect\BCAY{Eric, Krishnan, Charette,\ \BBA\ Manning}{Eric
  et~al.}{2017}]{Eric-etal-2017-key}
Eric, M., Krishnan, L., Charette, F., \BBA\ Manning, C.~D. \BBOP2017\BBCP.
\newblock \BBOQ Key-value retrieval networks for task-oriented dialogue\BBCQ\
\newblock In {\Bem Proceedings of the 18th Annual {SIG}dial Meeting on
  Discourse and Dialogue}, \BPGS\ 37--49, Saarbr{\"u}cken, Germany. Association
  for Computational Linguistics.

\bibitem[\protect\BCAY{Fan, Shi,\ \BBA\ Truong}{Fan
  et~al.}{2020}]{fan2020practices}
Fan, M., Shi, S., \BBA\ Truong, K.~N. \BBOP2020\BBCP.
\newblock \BBOQ Practices and challenges of using think-aloud protocols in
  industry: An international survey.\BBCQ\
\newblock {\Bem Journal of Usability Studies}, {\Bem 15\/}(2).

\bibitem[\protect\BCAY{Fan\ \BBA\ Luo}{Fan\ \BBA\ Luo}{2020}]{fan2020survey}
Fan, Y.\BBACOMMA\  \BBA\ Luo, X. \BBOP2020\BBCP.
\newblock \BBOQ A survey of dialogue system evaluation\BBCQ\
\newblock In {\Bem 2020 IEEE 32nd International Conference on Tools with
  Artificial Intelligence (ICTAI)}, \BPGS\ 1202--1209. IEEE.

\bibitem[\protect\BCAY{Federici, de~Filippis, Mele, Borsci, Bracalenti,
  Gaudino, Cocco, Amendola,\ \BBA\ Simonetti}{Federici
  et~al.}{2020}]{federici2020inside}
Federici, S., de~Filippis, M.~L., Mele, M.~L., Borsci, S., Bracalenti, M.,
  Gaudino, G., Cocco, A., Amendola, M., \BBA\ Simonetti, E. \BBOP2020\BBCP.
\newblock \BBOQ Inside pandora’s box: a systematic review of the assessment
  of the perceived quality of chatbots for people with disabilities or special
  needs\BBCQ\
\newblock {\Bem Disability and rehabilitation: assistive technology}, {\Bem
  15\/}(7), 832--837.

\bibitem[\protect\BCAY{Feng, Feng,\ \BBA\ Qin}{Feng
  et~al.}{2022}]{ijcai2022p764}
Feng, X., Feng, X., \BBA\ Qin, B. \BBOP2022\BBCP.
\newblock \BBOQ A survey on dialogue summarization: Recent advances and new
  frontiers\BBCQ\
\newblock In Raedt, L.~D.\BED, {\Bem Proceedings of the Thirty-First
  International Joint Conference on Artificial Intelligence, {IJCAI-22}},
  \BPGS\ 5453--5460. International Joint Conferences on Artificial Intelligence
  Organization.
\newblock Survey Track.

\bibitem[\protect\BCAY{Finch, Finch,\ \BBA\ Choi}{Finch
  et~al.}{2021}]{Finch-etal-2021-went}
Finch, J.~D., Finch, S.~E., \BBA\ Choi, J.~D. \BBOP2021\BBCP.
\newblock \BBOQ What went wrong? explaining overall dialogue quality through
  utterance-level impacts\BBCQ\
\newblock In {\Bem Proceedings of the 3rd Workshop on Natural Language
  Processing for Conversational AI}, \BPGS\ 93--101, Online. Association for
  Computational Linguistics.

\bibitem[\protect\BCAY{Finch\ \BBA\ Choi}{Finch\ \BBA\
  Choi}{2020}]{finch-choi-2020-towards}
Finch, S.~E.\BBACOMMA\  \BBA\ Choi, J.~D. \BBOP2020\BBCP.
\newblock \BBOQ Towards unified dialogue system evaluation: A comprehensive
  analysis of current evaluation protocols\BBCQ\
\newblock In {\Bem Proceedings of the 21th Annual Meeting of the Special
  Interest Group on Discourse and Dialogue}, \BPGS\ 236--245, 1st virtual
  meeting. Association for Computational Linguistics.

\bibitem[\protect\BCAY{Fiore, Baldauf,\ \BBA\ Thiel}{Fiore
  et~al.}{2019}]{Fiore2019forgot}
Fiore, D., Baldauf, M., \BBA\ Thiel, C. \BBOP2019\BBCP.
\newblock \BBOQ " forgot your password again?" acceptance and user experience
  of a chatbot for in-company it support\BBCQ\
\newblock In {\Bem Proceedings of the 18th International Conference on Mobile
  and Ubiquitous Multimedia}, \BPGS\ 1--11.

\bibitem[\protect\BCAY{Firdaus, Pratap~Shandeelya,\ \BBA\ Ekbal}{Firdaus
  et~al.}{2020}]{Firdaus2020more}
Firdaus, M., Pratap~Shandeelya, A., \BBA\ Ekbal, A. \BBOP2020\BBCP.
\newblock \BBOQ More to diverse: Generating diversified responses in a task
  oriented multimodal dialog system\BBCQ\
\newblock {\Bem PloS one}, {\Bem 15\/}(11), e0241271.

\bibitem[\protect\BCAY{Fitrianie, Bruijnes, Richards, B\"{o}nsch,\ \BBA\
  Brinkman}{Fitrianie et~al.}{2020}]{10.1145/3383652.3423873}
Fitrianie, S., Bruijnes, M., Richards, D., B\"{o}nsch, A., \BBA\ Brinkman,
  W.-P. \BBOP2020\BBCP.
\newblock \BBOQ The 19 unifying questionnaire constructs of artificial social
  agents: An iva community analysis\BBCQ\
\newblock In {\Bem Proceedings of the 20th ACM International Conference on
  Intelligent Virtual Agents}, IVA '20, New York, NY, USA. Association for
  Computing Machinery.

\bibitem[\protect\BCAY{Flake\ \BBA\ Fried}{Flake\ \BBA\
  Fried}{2020}]{flake2020measurement}
Flake, J.~K.\BBACOMMA\  \BBA\ Fried, E.~I. \BBOP2020\BBCP.
\newblock \BBOQ Measurement schmeasurement: Questionable measurement practices
  and how to avoid them\BBCQ\
\newblock {\Bem Advances in Methods and Practices in Psychological Science},
  {\Bem 3\/}(4), 456--465.

\bibitem[\protect\BCAY{F{\o}lstad, Nordheim,\ \BBA\ Bj{\o}rkli}{F{\o}lstad
  et~al.}{2018}]{folstad2018makes}
F{\o}lstad, A., Nordheim, C.~B., \BBA\ Bj{\o}rkli, C.~A. \BBOP2018\BBCP.
\newblock \BBOQ What makes users trust a chatbot for customer service? an
  exploratory interview study\BBCQ\
\newblock In {\Bem Internet Science: 5th International Conference, INSCI 2018,
  St. Petersburg, Russia, October 24--26, 2018, Proceedings 5}, \BPGS\
  194--208. Springer.

\bibitem[\protect\BCAY{Fonseca, Yankovskaya, Martins, Fishel,\ \BBA\
  Federmann}{Fonseca et~al.}{2019}]{fonseca-etal-2019-findings}
Fonseca, E., Yankovskaya, L., Martins, A. F.~T., Fishel, M., \BBA\ Federmann,
  C. \BBOP2019\BBCP.
\newblock \BBOQ Findings of the {WMT} 2019 shared tasks on quality
  estimation\BBCQ\
\newblock In Bojar, O., Chatterjee, R., Federmann, C., Fishel, M., Graham, Y.,
  Haddow, B., Huck, M., Yepes, A.~J., Koehn, P., Martins, A., Monz, C., Negri,
  M., N{\'e}v{\'e}ol, A., Neves, M., Post, M., Turchi, M., \BBA\ Verspoor,
  K.\BEDS, {\Bem Proceedings of the Fourth Conference on Machine Translation
  (Volume 3: Shared Task Papers, Day 2)}, \BPGS\ 1--10, Florence, Italy.
  Association for Computational Linguistics.

\bibitem[\protect\BCAY{Forkan, Jayaraman, Kang,\ \BBA\ Morshed}{Forkan
  et~al.}{2020}]{forkan2020echo}
Forkan, A. R.~M., Jayaraman, P.~P., Kang, Y.-B., \BBA\ Morshed, A.
  \BBOP2020\BBCP.
\newblock \BBOQ Echo: A tool for empirical evaluation cloud chatbots\BBCQ\
\newblock In {\Bem 2020 20th IEEE/ACM International Symposium on Cluster, Cloud
  and Internet Computing (CCGRID)}, \BPGS\ 669--672. IEEE.

\bibitem[\protect\BCAY{Foster, Giuliani,\ \BBA\ Knoll}{Foster
  et~al.}{2009}]{Foster-etal-2009-comparing}
Foster, M.~E., Giuliani, M., \BBA\ Knoll, A. \BBOP2009\BBCP.
\newblock \BBOQ Comparing objective and subjective measures of usability in a
  human-robot dialogue system\BBCQ\
\newblock In {\Bem Proceedings of the Joint Conference of the 47th Annual
  Meeting of the {ACL} and the 4th International Joint Conference on Natural
  Language Processing of the {AFNLP}}, \BPGS\ 879--887, Suntec, Singapore.
  Association for Computational Linguistics.

\bibitem[\protect\BCAY{Fried\ \BBA\ Flake}{Fried\ \BBA\
  Flake}{2018}]{fried2018measurement}
Fried, E.~I.\BBACOMMA\  \BBA\ Flake, J.~K. \BBOP2018\BBCP.
\newblock \BBOQ Measurement matters\BBCQ\
\newblock {\Bem APS Observer}, {\Bem 31\/}(3), 29--31.

\bibitem[\protect\BCAY{Gandhe\ \BBA\ Traum}{Gandhe\ \BBA\
  Traum}{2008}]{gandhe-traum-2008-evaluation}
Gandhe, S.\BBACOMMA\  \BBA\ Traum, D. \BBOP2008\BBCP.
\newblock \BBOQ Evaluation understudy for dialogue coherence models\BBCQ\
\newblock In {\Bem Proceedings of the 9th {SIG}dial Workshop on Discourse and
  Dialogue}, \BPGS\ 172--181, Columbus, Ohio. Association for Computational
  Linguistics.

\bibitem[\protect\BCAY{Gehrmann, Adewumi, Aggarwal, Ammanamanchi, Aremu,
  Bosselut, Chandu, Clinciu, Das, Dhole, Du, Durmus, Du{\v{s}}ek, Emezue,
  Gangal, Garbacea, Hashimoto, Hou, Jernite, \ldots,\ \BBA\ Zhou}{Gehrmann
  et~al.}{2021}]{gehrmann-etal-2021-gem}
Gehrmann, S., Adewumi, T., Aggarwal, K., Ammanamanchi, P., Aremu, A., Bosselut,
  A., Chandu, K., Clinciu, M., Das, D., Dhole, K., Du, W., Durmus, E.,
  Du{\v{s}}ek, O., Emezue, C., Gangal, V., Garbacea, C., Hashimoto, T., Hou,
  Y., Jernite, Y., \ldots, \BBA\ Zhou, J. \BBOP2021\BBCP.
\newblock \BBOQ The {GEM} benchmark: Natural language generation, its
  evaluation and metrics\BBCQ\
\newblock In {\Bem Proceedings of the 1st Workshop on Natural Language
  Generation, Evaluation, and Metrics (GEM 2021)}, \BPGS\ 96--120, Online.
  Association for Computational Linguistics.

\bibitem[\protect\BCAY{Gehrmann, Bhattacharjee, Mahendiran, Wang, Papangelis,
  Madaan, Mcmillan-major, Shvets, Upadhyay, Bohnet, Yao, Wilie, Bhagavatula,
  You, Thomson, Garbacea, Wang, \ldots,\ \BBA\ Hou}{Gehrmann
  et~al.}{2022}]{gehrmann-etal-2022-gemv2}
Gehrmann, S., Bhattacharjee, A., Mahendiran, A., Wang, A., Papangelis, A.,
  Madaan, A., Mcmillan-major, A., Shvets, A., Upadhyay, A., Bohnet, B., Yao,
  B., Wilie, B., Bhagavatula, C., You, C., Thomson, C., Garbacea, C., Wang, D.,
  \ldots, \BBA\ Hou, Y. \BBOP2022\BBCP.
\newblock \BBOQ {GEM}v2: Multilingual {NLG} benchmarking in a single line of
  code\BBCQ\
\newblock In Che, W.\BBACOMMA\  \BBA\ Shutova, E.\BEDS, {\Bem Proceedings of
  the 2022 Conference on Empirical Methods in Natural Language Processing:
  System Demonstrations}, \BPGS\ 266--281, Abu Dhabi, UAE. Association for
  Computational Linguistics.

\bibitem[\protect\BCAY{Gerken}{Gerken}{2024}]{dpdbbc}
Gerken, T. \BBOP2024\BBCP.
\newblock \BBOQ Dpd error caused chatbot to swear at customer\BBCQ\
\newblock Published on bbc.com, January 19, 2024. URL:
  \url{https://www.bbc.com/news/technology-68025677}.
\newblock Accessed March 5, 2024.

\bibitem[\protect\BCAY{Ghosh\ \BBA\ Mandal}{Ghosh\ \BBA\
  Mandal}{2020}]{ghosh2020webcare}
Ghosh, T.\BBACOMMA\  \BBA\ Mandal, S. \BBOP2020\BBCP.
\newblock \BBOQ Webcare quality: conceptualisation, scale development and
  validation\BBCQ\
\newblock {\Bem Journal of Marketing Management}, {\Bem 36\/}(15-16),
  1556--1590.

\bibitem[\protect\BCAY{Gonzales\ \BBA\ Gonz{\'a}lez}{Gonzales\ \BBA\
  Gonz{\'a}lez}{2017}]{Gonzales2017bots}
Gonzales, H. M.~S.\BBACOMMA\  \BBA\ Gonz{\'a}lez, M.~S. \BBOP2017\BBCP.
\newblock \BBOQ Bots as a news service and its emotional connection with
  audiences. the case of politibot\BBCQ\
\newblock {\Bem Doxa Comunicaci{\'o}n: Revista interdisciplinar de estudios de
  Comunicaci{\'o}n y Ciencias Sociales}, {\Bem 25}, 51--68.

\bibitem[\protect\BCAY{Gururangan, Card, Dreier, Gade, Wang, Wang,
  Zettlemoyer,\ \BBA\ Smith}{Gururangan
  et~al.}{2022}]{gururangan-etal-2022-whose}
Gururangan, S., Card, D., Dreier, S., Gade, E., Wang, L., Wang, Z.,
  Zettlemoyer, L., \BBA\ Smith, N.~A. \BBOP2022\BBCP.
\newblock \BBOQ Whose language counts as high quality? measuring language
  ideologies in text data selection\BBCQ\
\newblock In Goldberg, Y., Kozareva, Z., \BBA\ Zhang, Y.\BEDS, {\Bem
  Proceedings of the 2022 Conference on Empirical Methods in Natural Language
  Processing}, \BPGS\ 2562--2580, Abu Dhabi, United Arab Emirates. Association
  for Computational Linguistics.

\bibitem[\protect\BCAY{Ham, Lee, Jang,\ \BBA\ Kim}{Ham
  et~al.}{2020}]{ham-etal-2020-end}
Ham, D., Lee, J.-G., Jang, Y., \BBA\ Kim, K.-E. \BBOP2020\BBCP.
\newblock \BBOQ End-to-end neural pipeline for goal-oriented dialogue systems
  using {GPT}-2\BBCQ\
\newblock In {\Bem Proceedings of the 58th Annual Meeting of the Association
  for Computational Linguistics}, \BPGS\ 583--592, Online. Association for
  Computational Linguistics.

\bibitem[\protect\BCAY{Han, Smeaton,\ \BBA\ Jones}{Han
  et~al.}{2021}]{han-etal-2021-translation}
Han, L., Smeaton, A., \BBA\ Jones, G. \BBOP2021\BBCP.
\newblock \BBOQ Translation quality assessment: A brief survey on manual and
  automatic methods\BBCQ\
\newblock In Bizzoni, Y., Teich, E., Espa{\~n}a-Bonet, C., \BBA\ van Genabith,
  J.\BEDS, {\Bem Proceedings for the First Workshop on Modelling Translation:
  Translatology in the Digital Age}, \BPGS\ 15--33, online. Association for
  Computational Linguistics.

\bibitem[\protect\BCAY{Han}{Han}{2021}]{han2021impact}
Han, M.~C. \BBOP2021\BBCP.
\newblock \BBOQ The impact of anthropomorphism on consumers' purchase decision
  in chatbot commerce\BBCQ\
\newblock {\Bem Journal of Internet Commerce}, {\Bem 20\/}(1), 46--65.

\bibitem[\protect\BCAY{Han, Zhou, Turner,\ \BBA\ Yeh}{Han
  et~al.}{2021}]{han2021designing}
Han, X., Zhou, M., Turner, M.~J., \BBA\ Yeh, T. \BBOP2021\BBCP.
\newblock \BBOQ Designing effective interview chatbots: Automatic chatbot
  profiling and design suggestion generation for chatbot debugging\BBCQ\
\newblock In {\Bem Proceedings of the 2021 CHI Conference on Human Factors in
  Computing Systems}, \BPGS\ 1--15.

\bibitem[\protect\BCAY{Harms, Kucherbaev, Bozzon,\ \BBA\ Houben}{Harms
  et~al.}{2018}]{harms2018approaches}
Harms, J.-G., Kucherbaev, P., Bozzon, A., \BBA\ Houben, G.-J. \BBOP2018\BBCP.
\newblock \BBOQ Approaches for dialog management in conversational agents\BBCQ\
\newblock {\Bem IEEE Internet Computing}, {\Bem 23\/}(2), 13--22.

\bibitem[\protect\BCAY{Harris\ \BBA\ de~Chernatony}{Harris\ \BBA\
  de~Chernatony}{2001}]{harris2001corporate}
Harris, F.\BBACOMMA\  \BBA\ de~Chernatony, L. \BBOP2001\BBCP.
\newblock \BBOQ Corporate branding and corporate brand performance\BBCQ\
\newblock {\Bem European Journal of Marketing}, {\Bem 35\/}(3/4), 441--456.

\bibitem[\protect\BCAY{Hart\ \BBA\ Staveland}{Hart\ \BBA\
  Staveland}{1988}]{hart1988development}
Hart, S.~G.\BBACOMMA\  \BBA\ Staveland, L.~E. \BBOP1988\BBCP.
\newblock \BBOQ Development of nasa-tlx (task load index): Results of empirical
  and theoretical research\BBCQ\
\newblock In {\Bem Advances in psychology}, \lowercase{\BVOL}~52, \BPGS\
  139--183. Elsevier.

\bibitem[\protect\BCAY{Hassenzahl\ \BBA\ Tractinsky}{Hassenzahl\ \BBA\
  Tractinsky}{2006}]{hassenzahl2006user}
Hassenzahl, M.\BBACOMMA\  \BBA\ Tractinsky, N. \BBOP2006\BBCP.
\newblock \BBOQ User experience-a research agenda\BBCQ\
\newblock {\Bem Behaviour \& information technology}, {\Bem 25\/}(2), 91--97.

\bibitem[\protect\BCAY{Higashinaka, Funakoshi, Araki, Tsukahara, Kobayashi,\
  \BBA\ Mizukami}{Higashinaka et~al.}{2015}]{higashinaka-etal-2015-towards}
Higashinaka, R., Funakoshi, K., Araki, M., Tsukahara, H., Kobayashi, Y., \BBA\
  Mizukami, M. \BBOP2015\BBCP.
\newblock \BBOQ Towards taxonomy of errors in chat-oriented dialogue
  systems\BBCQ\
\newblock In Koller, A., Skantze, G., Jurcicek, F., Araki, M., \BBA\ Rose,
  C.~P.\BEDS, {\Bem Proceedings of the 16th Annual Meeting of the Special
  Interest Group on Discourse and Dialogue}, \BPGS\ 87--95, Prague, Czech
  Republic. Association for Computational Linguistics.

\bibitem[\protect\BCAY{Holmes, Moorhead, Bond, Zheng, Coates,\ \BBA\
  McTear}{Holmes et~al.}{2019}]{holmes2019usability}
Holmes, S., Moorhead, A., Bond, R., Zheng, H., Coates, V., \BBA\ McTear, M.
  \BBOP2019\BBCP.
\newblock \BBOQ Usability testing of a healthcare chatbot: Can we use
  conventional methods to assess conversational user interfaces?\BBCQ\
\newblock In {\Bem Proceedings of the 31st European Conference on Cognitive
  Ergonomics}, \BPGS\ 207--214.

\bibitem[\protect\BCAY{Horton}{Horton}{2023}]{horton2023large}
Horton, J.~J. \BBOP2023\BBCP.
\newblock \BBOQ Large language models as simulated economic agents: What can we
  learn from homo silicus?\BBCQ\
\newblock \BTR, National Bureau of Economic Research.

\bibitem[\protect\BCAY{Howcroft, Belz, Clinciu, Gkatzia, Hasan, Mahamood,
  Mille, van Miltenburg, Santhanam,\ \BBA\ Rieser}{Howcroft
  et~al.}{2020}]{howcroft-etal-2020-twenty}
Howcroft, D.~M., Belz, A., Clinciu, M.-A., Gkatzia, D., Hasan, S.~A., Mahamood,
  S., Mille, S., van Miltenburg, E., Santhanam, S., \BBA\ Rieser, V.
  \BBOP2020\BBCP.
\newblock \BBOQ Twenty years of confusion in human evaluation: {NLG} needs
  evaluation sheets and standardised definitions\BBCQ\
\newblock In {\Bem Proceedings of the 13th International Conference on Natural
  Language Generation}, \BPGS\ 169--182, Dublin, Ireland. Association for
  Computational Linguistics.

\bibitem[\protect\BCAY{Hu, Gao, Hu, Zhang, Chen, Xu,\ \BBA\ Wan}{Hu
  et~al.}{2024}]{hu2024llmbased}
Hu, X., Gao, M., Hu, S., Zhang, Y., Chen, Y., Xu, T., \BBA\ Wan, X.
  \BBOP2024\BBCP.
\newblock \BBOQ Are llm-based evaluators confusing nlg quality criteria?\BBCQ\
\newblock {\Bem CoRR}, {\Bem abs/2402.12055}.

\bibitem[\protect\BCAY{Huang, Wu, Liang, Wang, Shi, Wu, Yang,\ \BBA\
  Zhao}{Huang et~al.}{2023}]{huang2023towards}
Huang, H., Wu, S., Liang, X., Wang, B., Shi, Y., Wu, P., Yang, M., \BBA\ Zhao,
  T. \BBOP2023\BBCP.
\newblock \BBOQ Towards making the most of llm for translation quality
  estimation\BBCQ\
\newblock In {\Bem CCF International Conference on Natural Language Processing
  and Chinese Computing}, \BPGS\ 375--386. Springer.

\bibitem[\protect\BCAY{Hwang, Macdonald,\ \BBA\ Ahn}{Hwang
  et~al.}{2019}]{hwang2019end}
Hwang, E.~J., Macdonald, B.~A., \BBA\ Ahn, H.~S. \BBOP2019\BBCP.
\newblock \BBOQ End-to-end dialogue system with multi languages for hospital
  receptionist robot\BBCQ\
\newblock In {\Bem 2019 16th International Conference on Ubiquitous Robots
  (UR)}, \BPGS\ 278--283. IEEE.

\bibitem[\protect\BCAY{Ihsani, Baizal,\ \BBA\ Ikhsan}{Ihsani
  et~al.}{2021}]{ihsani2021conversational}
Ihsani, N.~S., Baizal, Z., \BBA\ Ikhsan, N. \BBOP2021\BBCP.
\newblock \BBOQ Conversational recommender system based on functional
  requirements and technical specifications\BBCQ\
\newblock In {\Bem 2021 International Conference on Data Science and Its
  Applications (ICoDSA)}, \BPGS\ 203--208. IEEE.

\bibitem[\protect\BCAY{Inie, Stray,\ \BBA\ Derczynski}{Inie
  et~al.}{2023}]{inie2023summon}
Inie, N., Stray, J., \BBA\ Derczynski, L. \BBOP2023\BBCP.
\newblock \BBOQ Summon a demon and bind it: {A} grounded theory of {LLM} red
  teaming in the wild\BBCQ\
\newblock {\Bem CoRR}, {\Bem abs/2311.06237}.

\bibitem[\protect\BCAY{Jang, Kim, Kim, Hong, Kim,\ \BBA\ Kim}{Jang
  et~al.}{2021}]{Jang2021mobile}
Jang, S., Kim, J.-J., Kim, S.-J., Hong, J., Kim, S., \BBA\ Kim, E.
  \BBOP2021\BBCP.
\newblock \BBOQ Mobile app-based chatbot to deliver cognitive behavioral
  therapy and psychoeducation for adults with attention deficit: A development
  and feasibility/usability study\BBCQ\
\newblock {\Bem International journal of medical informatics}, {\Bem 150},
  104440.

\bibitem[\protect\BCAY{Jannach, Manzoor, Cai,\ \BBA\ Chen}{Jannach
  et~al.}{2021}]{jannach2021survey}
Jannach, D., Manzoor, A., Cai, W., \BBA\ Chen, L. \BBOP2021\BBCP.
\newblock \BBOQ A survey on conversational recommender systems\BBCQ\
\newblock {\Bem ACM Computing Surveys (CSUR)}, {\Bem 54\/}(5), 1--36.

\bibitem[\protect\BCAY{Ji, Lee, Frieske, Yu, Su, Xu, Ishii, Bang, Madotto,\
  \BBA\ Fung}{Ji et~al.}{2023}]{ji2023survey}
Ji, Z., Lee, N., Frieske, R., Yu, T., Su, D., Xu, Y., Ishii, E., Bang, Y.~J.,
  Madotto, A., \BBA\ Fung, P. \BBOP2023\BBCP.
\newblock \BBOQ Survey of hallucination in natural language generation\BBCQ\
\newblock {\Bem ACM Comput. Surv.}, {\Bem 55\/}(12).

\bibitem[\protect\BCAY{Jiang, Dai, Yang, Zhao,\ \BBA\ Wei}{Jiang
  et~al.}{2021}]{jiang-etal-2021-towards}
Jiang, H., Dai, B., Yang, M., Zhao, T., \BBA\ Wei, W. \BBOP2021\BBCP.
\newblock \BBOQ Towards automatic evaluation of dialog systems: A model-free
  off-policy evaluation approach\BBCQ\
\newblock In {\Bem Proceedings of the 2021 Conference on Empirical Methods in
  Natural Language Processing}, \BPGS\ 7419--7451, Online and Punta Cana,
  Dominican Republic. Association for Computational Linguistics.

\bibitem[\protect\BCAY{Jiang, Cheng, Yang,\ \BBA\ Gao}{Jiang
  et~al.}{2022}]{JIANG2022107329}
Jiang, H., Cheng, Y., Yang, J., \BBA\ Gao, S. \BBOP2022\BBCP.
\newblock \BBOQ Ai-powered chatbot communication with customers: Dialogic
  interactions, satisfaction, engagement, and customer behavior\BBCQ\
\newblock {\Bem Computers in Human Behavior}, {\Bem 134}, 107329.

\bibitem[\protect\BCAY{Jim{\'e}nez-Barreto, Rubio,\ \BBA\
  Molinillo}{Jim{\'e}nez-Barreto et~al.}{2021}]{jimenez2021find}
Jim{\'e}nez-Barreto, J., Rubio, N., \BBA\ Molinillo, S. \BBOP2021\BBCP.
\newblock \BBOQ “find a flight for me, oscar!” motivational customer
  experiences with chatbots\BBCQ\
\newblock {\Bem International Journal of Contemporary Hospitality Management},
  {\Bem 33\/}(11), 3860--3882.

\bibitem[\protect\BCAY{Jurafsky\ \BBA\ Martin}{Jurafsky\ \BBA\
  Martin}{2021}]{jurafsky2021}
Jurafsky, D.\BBACOMMA\  \BBA\ Martin, J.~H. \BBOP2021\BBCP.
\newblock {\Bem Speech and Language Processing}.
\newblock Pre-publication draft.

\bibitem[\protect\BCAY{Jwalapuram}{Jwalapuram}{2017}]{jwalapuram2017evaluating}
Jwalapuram, P. \BBOP2017\BBCP.
\newblock \BBOQ Evaluating dialogs based on grice’s maxims\BBCQ\
\newblock In {\Bem Proceedings of the Student Research Workshop associated with
  RANLP}, \BPGS\ 17--24.

\bibitem[\protect\BCAY{Kadariya, Venkataramanan, Yip, Kalra, Thirunarayanan,\
  \BBA\ Sheth}{Kadariya et~al.}{2019}]{kadariya2019kbot}
Kadariya, D., Venkataramanan, R., Yip, H.~Y., Kalra, M., Thirunarayanan, K.,
  \BBA\ Sheth, A. \BBOP2019\BBCP.
\newblock \BBOQ kbot: knowledge-enabled personalized chatbot for asthma
  self-management\BBCQ\
\newblock In {\Bem 2019 IEEE International Conference on Smart Computing
  (SMARTCOMP)}, \BPGS\ 138--143. IEEE.

\bibitem[\protect\BCAY{Karpinska, Akoury,\ \BBA\ Iyyer}{Karpinska
  et~al.}{2021}]{karpinska-etal-2021-perils}
Karpinska, M., Akoury, N., \BBA\ Iyyer, M. \BBOP2021\BBCP.
\newblock \BBOQ The perils of using {M}echanical {T}urk to evaluate open-ended
  text generation\BBCQ\
\newblock In Moens, M.-F., Huang, X., Specia, L., \BBA\ Yih, S. W.-t.\BEDS,
  {\Bem Proceedings of the 2021 Conference on Empirical Methods in Natural
  Language Processing}, \BPGS\ 1265--1285, Online and Punta Cana, Dominican
  Republic. Association for Computational Linguistics.

\bibitem[\protect\BCAY{Kataoka, Takemura, Sasajima, Katoh, et~al.}{Kataoka
  et~al.}{2021}]{kataoka2021development}
Kataoka, Y., Takemura, T., Sasajima, M., Katoh, N., et~al. \BBOP2021\BBCP.
\newblock \BBOQ Development and early feasibility of chatbots for educating
  patients with lung cancer and their caregivers in japan: Mixed methods
  study\BBCQ\
\newblock {\Bem JMIR cancer}, {\Bem 7\/}(1), e26911.

\bibitem[\protect\BCAY{Kattenbeck, Kilian, Ferstl, Alt,\ \BBA\
  Ludwig}{Kattenbeck et~al.}{2018}]{kattenbeck2018airbot}
Kattenbeck, M., Kilian, M.~A., Ferstl, M., Alt, F., \BBA\ Ludwig, B.
  \BBOP2018\BBCP.
\newblock \BBOQ Airbot: using a work flow model for proactive assistance in
  public spaces\BBCQ\
\newblock In {\Bem Proceedings of the 20th International Conference on
  Human-Computer Interaction with Mobile Devices and Services Adjunct}, \BPGS\
  213--220.

\bibitem[\protect\BCAY{Khadpe, Krishna, Fei-Fei, Hancock,\ \BBA\
  Bernstein}{Khadpe et~al.}{2020}]{khadpe2020conceptual}
Khadpe, P., Krishna, R., Fei-Fei, L., Hancock, J.~T., \BBA\ Bernstein, M.~S.
  \BBOP2020\BBCP.
\newblock \BBOQ Conceptual metaphors impact perceptions of human-ai
  collaboration\BBCQ\
\newblock {\Bem Proc. ACM Hum.-Comput. Interact.}, {\Bem 4\/}(CSCW2).

\bibitem[\protect\BCAY{Khayrallah\ \BBA\ Sedoc}{Khayrallah\ \BBA\
  Sedoc}{2021}]{Khayrallah-sedoc-2021-measuring}
Khayrallah, H.\BBACOMMA\  \BBA\ Sedoc, J. \BBOP2021\BBCP.
\newblock \BBOQ Measuring the {`}{I} don{'}t know{'} problem through the lens
  of {G}ricean quantity\BBCQ\
\newblock In {\Bem Proceedings of the 2021 Conference of the North American
  Chapter of the Association for Computational Linguistics: Human Language
  Technologies}, \BPGS\ 5659--5670, Online. Association for Computational
  Linguistics.

\bibitem[\protect\BCAY{Kimchi, Polivka,\ \BBA\ Stevenson}{Kimchi
  et~al.}{1991}]{kimchi1991triangulation}
Kimchi, J., Polivka, B., \BBA\ Stevenson, J.~S. \BBOP1991\BBCP.
\newblock \BBOQ Triangulation: Operational definitions\BBCQ\
\newblock {\Bem Nursing Research}, {\Bem 40\/}(6).

\bibitem[\protect\BCAY{Klarna}{Klarna}{2024}]{klarna}
Klarna \BBOP2024\BBCP.
\newblock \BBOQ Klarna ai assistant handles two-thirds of customer service
  chats in its first month\BBCQ\
\newblock Published on klarna.com, February 27, 2024. URL:
  \url{https://www.klarna.com/international/press/klarna-ai-assistant-handles-two-thirds-of-customer-service-chats-in-its-first-month/}.
\newblock Accessed March 5, 2024.

\bibitem[\protect\BCAY{Kocaballi}{Kocaballi}{2023}]{kocaballi2023conversational}
Kocaballi, A.~B. \BBOP2023\BBCP.
\newblock \BBOQ Conversational ai-powered design: Chatgpt as designer, user,
  and product\BBCQ\
\newblock {\Bem CoRR}, {\Bem abs/2302.07406}.

\bibitem[\protect\BCAY{Kocmi\ \BBA\ Federmann}{Kocmi\ \BBA\
  Federmann}{2023}]{kocmi-federmann-2023-large}
Kocmi, T.\BBACOMMA\  \BBA\ Federmann, C. \BBOP2023\BBCP.
\newblock \BBOQ Large language models are state-of-the-art evaluators of
  translation quality\BBCQ\
\newblock In Nurminen, M., Brenner, J., Koponen, M., Latomaa, S., Mikhailov,
  M., Schierl, F., Ranasinghe, T., Vanmassenhove, E., Vidal, S.~A., Aranberri,
  N., Nunziatini, M., Escart{\'\i}n, C.~P., Forcada, M., Popovic, M., Scarton,
  C., \BBA\ Moniz, H.\BEDS, {\Bem Proceedings of the 24th Annual Conference of
  the European Association for Machine Translation}, \BPGS\ 193--203, Tampere,
  Finland. European Association for Machine Translation.

\bibitem[\protect\BCAY{Kocmi, Federmann, Grundkiewicz, Junczys-Dowmunt,
  Matsushita,\ \BBA\ Menezes}{Kocmi et~al.}{2021}]{kocmi-etal-2021-ship}
Kocmi, T., Federmann, C., Grundkiewicz, R., Junczys-Dowmunt, M., Matsushita,
  H., \BBA\ Menezes, A. \BBOP2021\BBCP.
\newblock \BBOQ To ship or not to ship: An extensive evaluation of automatic
  metrics for machine translation\BBCQ\
\newblock In Barrault, L., Bojar, O., Bougares, F., Chatterjee, R.,
  Costa-jussa, M.~R., Federmann, C., Fishel, M., Fraser, A., Freitag, M.,
  Graham, Y., Grundkiewicz, R., Guzman, P., Haddow, B., Huck, M., Yepes, A.~J.,
  Koehn, P., Kocmi, T., Martins, A., Morishita, M., \BBA\ Monz, C.\BEDS, {\Bem
  Proceedings of the Sixth Conference on Machine Translation}, \BPGS\ 478--494,
  Online. Association for Computational Linguistics.

\bibitem[\protect\BCAY{Lancaster}{Lancaster}{2023}]{forbes}
Lancaster, A. \BBOP2023\BBCP.
\newblock \BBOQ Beyond chatbots: the rise of large language models\BBCQ\
\newblock Published on forbes.com, March 20, 2023. URL:
  \url{https://www.forbes.com/sites/forbestechcouncil/2023/03/20/beyond-chatbots-the-rise-of-large-language-models/?sh=4b7438182319}.
\newblock Accessed March 5, 2024.

\bibitem[\protect\BCAY{Langevin, Lordon, Avrahami, Cowan, Hirsch,\ \BBA\
  Hsieh}{Langevin et~al.}{2021}]{langevin2021heuristic}
Langevin, R., Lordon, R.~J., Avrahami, T., Cowan, B.~R., Hirsch, T., \BBA\
  Hsieh, G. \BBOP2021\BBCP.
\newblock \BBOQ Heuristic evaluation of conversational agents\BBCQ\
\newblock In {\Bem Proceedings of the 2021 CHI Conference on Human Factors in
  Computing Systems}, \BPGS\ 1--15.

\bibitem[\protect\BCAY{Lazarsfeld}{Lazarsfeld}{1958}]{lazarsfeld1958evidence}
Lazarsfeld, P.~F. \BBOP1958\BBCP.
\newblock \BBOQ Evidence and inference in social research\BBCQ\
\newblock {\Bem Daedalus}, {\Bem 87\/}(4), 99--130.

\bibitem[\protect\BCAY{Lee, Pan,\ \BBA\ Hsieh}{Lee
  et~al.}{2022}]{lee2022artificial}
Lee, C.~T., Pan, L.-Y., \BBA\ Hsieh, S.~H. \BBOP2022\BBCP.
\newblock \BBOQ Artificial intelligent chatbots as brand promoters: a two-stage
  structural equation modeling-artificial neural network approach\BBCQ\
\newblock {\Bem Internet Research}, {\Bem 32\/}(4), 1329--1356.

\bibitem[\protect\BCAY{Lee, Jo, Kim, Jung,\ \BBA\ Kim}{Lee
  et~al.}{2021}]{Lee2021sumbt+}
Lee, H., Jo, S., Kim, H., Jung, S., \BBA\ Kim, T.-Y. \BBOP2021\BBCP.
\newblock \BBOQ Sumbt+ larl: Effective multi-domain end-to-end neural
  task-oriented dialog system\BBCQ\
\newblock {\Bem IEEE Access}, {\Bem 9}, 116133--116146.

\bibitem[\protect\BCAY{Leiter, Opitz, Deutsch, Gao, Dror,\ \BBA\ Eger}{Leiter
  et~al.}{2023}]{leiter-etal-2023-eval4nlp}
Leiter, C., Opitz, J., Deutsch, D., Gao, Y., Dror, R., \BBA\ Eger, S.
  \BBOP2023\BBCP.
\newblock \BBOQ The {E}val4{NLP} 2023 shared task on prompting large language
  models as explainable metrics\BBCQ\
\newblock In Deutsch, D., Dror, R., Eger, S., Gao, Y., Leiter, C., Opitz, J.,
  \BBA\ R{\"u}ckl{\'e}, A.\BEDS, {\Bem Proceedings of the 4th Workshop on
  Evaluation and Comparison of NLP Systems}, \BPGS\ 117--138, Bali, Indonesia.
  Association for Computational Linguistics.

\bibitem[\protect\BCAY{Lewis\ \BBA\ Mitchell}{Lewis\ \BBA\
  Mitchell}{1990}]{lewis1990defining}
Lewis, B.~R.\BBACOMMA\  \BBA\ Mitchell, V.~W. \BBOP1990\BBCP.
\newblock \BBOQ Defining and measuring the quality of customer service\BBCQ\
\newblock {\Bem Marketing intelligence \& planning}, {\Bem 8\/}(6), 11--17.

\bibitem[\protect\BCAY{Li, Lee, Emokpae,\ \BBA\ Yang}{Li
  et~al.}{2021a}]{li2021makes}
Li, L., Lee, K.~Y., Emokpae, E., \BBA\ Yang, S.-B. \BBOP2021a\BBCP.
\newblock \BBOQ What makes you continuously use chatbot services? evidence from
  chinese online travel agencies\BBCQ\
\newblock {\Bem Electronic Markets}, {\Bem 31\/}(3), 575--599.

\bibitem[\protect\BCAY{Li, Arnold, Yan, Shi,\ \BBA\ Yu}{Li
  et~al.}{2021b}]{li-etal-2021-legoeval}
Li, Y., Arnold, J., Yan, F., Shi, W., \BBA\ Yu, Z. \BBOP2021b\BBCP.
\newblock \BBOQ {LEGOE}val: An open-source toolkit for dialogue system
  evaluation via crowdsourcing\BBCQ\
\newblock In {\Bem Proceedings of the 59th Annual Meeting of the Association
  for Computational Linguistics and the 11th International Joint Conference on
  Natural Language Processing: System Demonstrations}, \BPGS\ 317--324, Online.
  Association for Computational Linguistics.

\bibitem[\protect\BCAY{Li, Xu, Shen, Xu, Gu,\ \BBA\ Tao}{Li
  et~al.}{2024}]{li2024leveraging}
Li, Z., Xu, X., Shen, T., Xu, C., Gu, J., \BBA\ Tao, C. \BBOP2024\BBCP.
\newblock \BBOQ Leveraging large language models for {NLG} evaluation: {A}
  survey\BBCQ\
\newblock {\Bem CoRR}, {\Bem abs/2401.07103}.

\bibitem[\protect\BCAY{Liebrecht\ \BBA\ van~der Weegen}{Liebrecht\ \BBA\
  van~der Weegen}{2019}]{liebrecht2019mensachtige}
Liebrecht, C.\BBACOMMA\  \BBA\ van~der Weegen, E. \BBOP2019\BBCP.
\newblock \BBOQ Menselijke chatbots: een zegen voor online klantcontact?\BBCQ\
\newblock {\Bem Tijdschrift voor Communicatiewetenschap}, {\Bem 47\/}(3).

\bibitem[\protect\BCAY{Liesenfeld\ \BBA\ Dingemanse}{Liesenfeld\ \BBA\
  Dingemanse}{2024}]{liesenfeld2024interactive}
Liesenfeld, A.\BBACOMMA\  \BBA\ Dingemanse, M. \BBOP2024\BBCP.
\newblock \BBOQ Interactive probes: action-level evaluation for dialogue
  systems\BBCQ\
\newblock \emph{Discourse and Communication.} In press.

\bibitem[\protect\BCAY{Limna\ \BBA\ Kraiwanit}{Limna\ \BBA\
  Kraiwanit}{2023}]{limna2023role}
Limna, P.\BBACOMMA\  \BBA\ Kraiwanit, T. \BBOP2023\BBCP.
\newblock \BBOQ The role of chatgpt on customer service in the hospitality
  industry: An exploratory study of hospitality workers' experiences and
  perceptions\BBCQ\
\newblock {\Bem Tourism and hospitality management}, {\Bem 29\/}(4), 583--592.

\bibitem[\protect\BCAY{Lin\ \BBA\ Och}{Lin\ \BBA\ Och}{2004}]{lin2004looking}
Lin, C.-Y.\BBACOMMA\  \BBA\ Och, F. \BBOP2004\BBCP.
\newblock \BBOQ Looking for a few good metrics: Rouge and its evaluation\BBCQ\
\newblock In {\Bem Ntcir workshop}.

\bibitem[\protect\BCAY{Liu, Lowe, Serban, Noseworthy, Charlin,\ \BBA\
  Pineau}{Liu et~al.}{2016}]{liu-etal-2016-evaluate}
Liu, C.-W., Lowe, R., Serban, I., Noseworthy, M., Charlin, L., \BBA\ Pineau, J.
  \BBOP2016\BBCP.
\newblock \BBOQ How {NOT} to evaluate your dialogue system: An empirical study
  of unsupervised evaluation metrics for dialogue response generation\BBCQ\
\newblock In {\Bem Proceedings of the 2016 Conference on Empirical Methods in
  Natural Language Processing}, \BPGS\ 2122--2132, Austin, Texas. Association
  for Computational Linguistics.

\bibitem[\protect\BCAY{Liu, Cai, Ou, Huang,\ \BBA\ Feng}{Liu
  et~al.}{2022}]{liu-etal-2022-generative}
Liu, H., Cai, Y., Ou, Z., Huang, Y., \BBA\ Feng, J. \BBOP2022\BBCP.
\newblock \BBOQ A generative user simulator with {GPT}-based architecture and
  goal state tracking for reinforced multi-domain dialog systems\BBCQ\
\newblock In Ou, Z., Feng, J., \BBA\ Li, J.\BEDS, {\Bem Proceedings of the
  Towards Semi-Supervised and Reinforced Task-Oriented Dialog Systems
  (SereTOD)}, \BPGS\ 85--97, Abu Dhabi, Beijing (Hybrid). Association for
  Computational Linguistics.

\bibitem[\protect\BCAY{Liu, Lu,\ \BBA\ Yang}{Liu
  et~al.}{2005}]{liu2005research}
Liu, H., Lu, R., \BBA\ Yang, F. \BBOP2005\BBCP.
\newblock \BBOQ Research on the multi-talk dialog system\BBCQ\
\newblock In {\Bem IRI-2005 IEEE International Conference on Information Reuse
  and Integration, Conf, 2005.}, \BPGS\ 344--349. IEEE.

\bibitem[\protect\BCAY{Liu, Takanobu, Wen, Wan, Li, Nie, Li, Peng,\ \BBA\
  Huang}{Liu et~al.}{2021}]{liu-etal-2021-robustness}
Liu, J., Takanobu, R., Wen, J., Wan, D., Li, H., Nie, W., Li, C., Peng, W.,
  \BBA\ Huang, M. \BBOP2021\BBCP.
\newblock \BBOQ Robustness testing of language understanding in task-oriented
  dialog\BBCQ\
\newblock In {\Bem Proceedings of the 59th Annual Meeting of the Association
  for Computational Linguistics and the 11th International Joint Conference on
  Natural Language Processing (Volume 1: Long Papers)}, \BPGS\ 2467--2480,
  Online. Association for Computational Linguistics.

\bibitem[\protect\BCAY{Liu, Huang, Wu, Zhu,\ \BBA\ Ba}{Liu
  et~al.}{2020}]{liu2020cbet}
Liu, Q., Huang, J., Wu, L., Zhu, K., \BBA\ Ba, S. \BBOP2020\BBCP.
\newblock \BBOQ Cbet: design and evaluation of a domain-specific chatbot for
  mobile learning\BBCQ\
\newblock {\Bem Universal Access in the Information Society}, {\Bem 19\/}(3),
  655--673.

\bibitem[\protect\BCAY{Liu, Zhang,\ \BBA\ Feng}{Liu
  et~al.}{2015}]{liu2015ergonomics}
Liu, W., Zhang, J., \BBA\ Feng, S. \BBOP2015\BBCP.
\newblock \BBOQ An ergonomics evaluation to chatbot equipped with
  knowledge-rich mind\BBCQ\
\newblock In {\Bem 2015 3rd International Symposium on Computational and
  Business Intelligence (ISCBI)}, \BPGS\ 95--99. IEEE.

\bibitem[\protect\BCAY{Liu, Han, Ma, Zhang, Yang, Tian, He, Li, He, Liu, Wu,
  Zhao, Zhu, Li, Qiang, Shen, Liu,\ \BBA\ Ge}{Liu
  et~al.}{2023}]{liu2023summary}
Liu, Y., Han, T., Ma, S., Zhang, J., Yang, Y., Tian, J., He, H., Li, A., He,
  M., Liu, Z., Wu, Z., Zhao, L., Zhu, D., Li, X., Qiang, N., Shen, D., Liu, T.,
  \BBA\ Ge, B. \BBOP2023\BBCP.
\newblock \BBOQ Summary of chatgpt-related research and perspective towards the
  future of large language models\BBCQ\
\newblock {\Bem Meta-Radiology}, {\Bem 1\/}(2), 100017.

\bibitem[\protect\BCAY{Liu, Feng,\ \BBA\ Chen}{Liu
  et~al.}{2021}]{liu2021dialtest}
Liu, Z., Feng, Y., \BBA\ Chen, Z. \BBOP2021\BBCP.
\newblock \BBOQ Dialtest: automated testing for recurrent-neural-network-driven
  dialogue systems\BBCQ\
\newblock In {\Bem Proceedings of the 30th ACM SIGSOFT International Symposium
  on Software Testing and Analysis}, \BPGS\ 115--126.

\bibitem[\protect\BCAY{L{\'o}pez-C{\'o}zar, De~la Torre, Segura,\ \BBA\
  Rubio}{L{\'o}pez-C{\'o}zar et~al.}{2003}]{lopez2003assessment}
L{\'o}pez-C{\'o}zar, R., De~la Torre, A., Segura, J.~C., \BBA\ Rubio, A.~J.
  \BBOP2003\BBCP.
\newblock \BBOQ Assessment of dialogue systems by means of a new simulation
  technique\BBCQ\
\newblock {\Bem Speech Communication}, {\Bem 40\/}(3), 387--407.

\bibitem[\protect\BCAY{Lowe, Serban, Noseworthy, Charlin,\ \BBA\ Pineau}{Lowe
  et~al.}{2016}]{lowe-etal-2016-evaluation}
Lowe, R., Serban, I.~V., Noseworthy, M., Charlin, L., \BBA\ Pineau, J.
  \BBOP2016\BBCP.
\newblock \BBOQ On the evaluation of dialogue systems with next utterance
  classification\BBCQ\
\newblock In {\Bem Proceedings of the 17th Annual Meeting of the Special
  Interest Group on Discourse and Dialogue}, \BPGS\ 264--269, Los Angeles.
  Association for Computational Linguistics.

\bibitem[\protect\BCAY{Maniou\ \BBA\ Veglis}{Maniou\ \BBA\
  Veglis}{2020}]{maniou2020employing}
Maniou, T.~A.\BBACOMMA\  \BBA\ Veglis, A. \BBOP2020\BBCP.
\newblock \BBOQ Employing a chatbot for news dissemination during crisis:
  Design, implementation and evaluation\BBCQ\
\newblock {\Bem Future Internet}, {\Bem 12\/}(7), 109.

\bibitem[\protect\BCAY{Margaretha\ \BBA\ DeVault}{Margaretha\ \BBA\
  DeVault}{2011}]{margaretha-devault-2011-approach}
Margaretha, E.\BBACOMMA\  \BBA\ DeVault, D. \BBOP2011\BBCP.
\newblock \BBOQ An approach to the automated evaluation of pipeline
  architectures in natural language dialogue systems\BBCQ\
\newblock In {\Bem Proceedings of the {SIGDIAL} 2011 Conference}, \BPGS\
  279--285, Portland, Oregon. Association for Computational Linguistics.

\bibitem[\protect\BCAY{Maroengsit, Piyakulpinyo, Phonyiam, Pongnumkul,
  Chaovalit,\ \BBA\ Theeramunkong}{Maroengsit
  et~al.}{2019}]{maroengsit2019survey}
Maroengsit, W., Piyakulpinyo, T., Phonyiam, K., Pongnumkul, S., Chaovalit, P.,
  \BBA\ Theeramunkong, T. \BBOP2019\BBCP.
\newblock \BBOQ A survey on evaluation methods for chatbots\BBCQ\
\newblock In {\Bem Proceedings of the 2019 7th International conference on
  information and education technology}, \BPGS\ 111--119.

\bibitem[\protect\BCAY{Martijn, van Hooijdonk, Kunneman,\ \BBA\ Hoeken}{Martijn
  et~al.}{ND}]{martijnreconfiguring}
Martijn, G., van Hooijdonk, C., Kunneman, F., \BBA\ Hoeken, H.
  \BBOP{N.D.}\BBCP.
\newblock \BBOQ Reconfiguring the customer service domain: perspectives of
  managers, conversational designers, and human agents on human-chatbot
  collaboration\BBCQ\
\newblock Accepted.

\bibitem[\protect\BCAY{Messick}{Messick}{1995}]{messick_validity_1995}
Messick, S. \BBOP1995\BBCP.
\newblock \BBOQ Validity of psychological assessment: {Validation} of
  inferences from persons' responses and performances as scientific inquiry
  into score meaning.\BBCQ\
\newblock {\Bem American Psychologist}, {\Bem 50\/}(9), 741--749.
\newblock Place: US Publisher: American Psychological Association.

\bibitem[\protect\BCAY{Meyer, Elsweiler, Ludwig, Fernandez-Pichel,\ \BBA\
  Losada}{Meyer et~al.}{2022}]{10.1145/3543829.3544529}
Meyer, S., Elsweiler, D., Ludwig, B., Fernandez-Pichel, M., \BBA\ Losada, D.~E.
  \BBOP2022\BBCP.
\newblock \BBOQ Do we still need human assessors? {P}rompt-based gpt-3 user
  simulation in conversational ai\BBCQ\
\newblock In {\Bem Proceedings of the 4th Conference on Conversational User
  Interfaces}, CUI '22, New York, NY, USA. Association for Computing Machinery.

\bibitem[\protect\BCAY{Mi, Zhou, Kong, Cai, Huang,\ \BBA\ Faltings}{Mi
  et~al.}{2021}]{mi-etal-2021-self}
Mi, F., Zhou, W., Kong, L., Cai, F., Huang, M., \BBA\ Faltings, B.
  \BBOP2021\BBCP.
\newblock \BBOQ Self-training improves pre-training for few-shot learning in
  task-oriented dialog systems\BBCQ\
\newblock In {\Bem Proceedings of the 2021 Conference on Empirical Methods in
  Natural Language Processing}, \BPGS\ 1887--1898, Online and Punta Cana,
  Dominican Republic. Association for Computational Linguistics.

\bibitem[\protect\BCAY{Miraj, Raza, Hussain, Siddiqi, Habib, Khan,\ \BBA\
  Chandir}{Miraj et~al.}{2021}]{miraj2021development}
Miraj, F., Raza, H., Hussain, O., Siddiqi, D., Habib, A., Khan, A., \BBA\
  Chandir, S. \BBOP2021\BBCP.
\newblock \BBOQ Development and feasibility-testing of an artificially
  intelligent chatbot to answer immunization-related queries of caregivers in
  pakistan: A mixed-methods evaluation\BBCQ\
\newblock In {\Bem TROPICAL MEDICINE \& INTERNATIONAL HEALTH},
  \lowercase{\BVOL}~26, \BPGS\ 140--141. WILEY 111 RIVER ST, HOBOKEN
  07030-5774, NJ USA.

\bibitem[\protect\BCAY{Mitchell}{Mitchell}{2021}]{mitchell2021why}
Mitchell, M. \BBOP2021\BBCP.
\newblock \BBOQ Why {AI} is harder than we think\BBCQ\
\newblock {\Bem CoRR}, {\Bem abs/2104.12871}.

\bibitem[\protect\BCAY{Mokmin\ \BBA\ Ibrahim}{Mokmin\ \BBA\
  Ibrahim}{2021}]{mokmin2021evaluation}
Mokmin, N. A.~M.\BBACOMMA\  \BBA\ Ibrahim, N.~A. \BBOP2021\BBCP.
\newblock \BBOQ The evaluation of chatbot as a tool for health literacy
  education among undergraduate students\BBCQ\
\newblock {\Bem Education and Information Technologies}, {\Bem 26\/}(5),
  6033--6049.

\bibitem[\protect\BCAY{Moramarco, Papadopoulos~Korfiatis, Perera, Juric, Flann,
  Reiter, Belz,\ \BBA\ Savkov}{Moramarco
  et~al.}{2022}]{moramarco-etal-2022-human}
Moramarco, F., Papadopoulos~Korfiatis, A., Perera, M., Juric, D., Flann, J.,
  Reiter, E., Belz, A., \BBA\ Savkov, A. \BBOP2022\BBCP.
\newblock \BBOQ Human evaluation and correlation with automatic metrics in
  consultation note generation\BBCQ\
\newblock In Muresan, S., Nakov, P., \BBA\ Villavicencio, A.\BEDS, {\Bem
  Proceedings of the 60th Annual Meeting of the Association for Computational
  Linguistics (Volume 1: Long Papers)}, \BPGS\ 5739--5754, Dublin, Ireland.
  Association for Computational Linguistics.

\bibitem[\protect\BCAY{Morgan\ \BBA\ Hunt}{Morgan\ \BBA\
  Hunt}{1994}]{morgan1994commitment}
Morgan, R.~M.\BBACOMMA\  \BBA\ Hunt, S.~D. \BBOP1994\BBCP.
\newblock \BBOQ The commitment-trust theory of relationship marketing\BBCQ\
\newblock {\Bem Journal of marketing}, {\Bem 58\/}(3), 20--38.

\bibitem[\protect\BCAY{Mori, MacDorman,\ \BBA\ Kageki}{Mori
  et~al.}{2012}]{mori2012uncanny}
Mori, M., MacDorman, K.~F., \BBA\ Kageki, N. \BBOP2012\BBCP.
\newblock \BBOQ The uncanny valley [from the field]\BBCQ\
\newblock {\Bem IEEE Robotics \& automation magazine}, {\Bem 19\/}(2), 98--100.

\bibitem[\protect\BCAY{Mostafa\ \BBA\ Kasamani}{Mostafa\ \BBA\
  Kasamani}{2022}]{mostafa2022antecedents}
Mostafa, R.~B.\BBACOMMA\  \BBA\ Kasamani, T. \BBOP2022\BBCP.
\newblock \BBOQ Antecedents and consequences of chatbot initial trust\BBCQ\
\newblock {\Bem European journal of marketing}, {\Bem 56\/}(6), 1748--1771.

\bibitem[\protect\BCAY{Nazir, Khan, Ahmed, Jami,\ \BBA\ Wasi}{Nazir
  et~al.}{2019}]{nazir2019novel}
Nazir, A., Khan, M.~Y., Ahmed, T., Jami, S.~I., \BBA\ Wasi, S. \BBOP2019\BBCP.
\newblock \BBOQ A novel approach for ontology-driven information retrieving
  chatbot for fashion brands\BBCQ\
\newblock {\Bem Int. J. Adv. Comput. Sci. Appl. IJACSA}, {\Bem 10\/}(9).

\bibitem[\protect\BCAY{Nedelchev, Usbeck,\ \BBA\ Lehmann}{Nedelchev
  et~al.}{2020}]{nedelchev2020treating}
Nedelchev, R., Usbeck, R., \BBA\ Lehmann, J. \BBOP2020\BBCP.
\newblock \BBOQ Treating dialogue quality evaluation as an anomaly detection
  problem\BBCQ\
\newblock In {\Bem Proceedings of The 12th Language Resources and Evaluation
  Conference}, \BPGS\ 508--512.

\bibitem[\protect\BCAY{Niculescu, Kukanov,\ \BBA\ Wadhwa}{Niculescu
  et~al.}{2020}]{niculescu2020digimo}
Niculescu, A.~I., Kukanov, I., \BBA\ Wadhwa, B. \BBOP2020\BBCP.
\newblock \BBOQ Digimo-towards developing an emotional intelligent chatbot in
  singapore\BBCQ\
\newblock In {\Bem Proceedings of the 2020 Symposium on Emerging Research from
  Asia and on Asian Contexts and Cultures}, \BPGS\ 29--32.

\bibitem[\protect\BCAY{Nielsen, Clemmensen,\ \BBA\ Yssing}{Nielsen
  et~al.}{2002}]{10.1145/572020.572033}
Nielsen, J., Clemmensen, T., \BBA\ Yssing, C. \BBOP2002\BBCP.
\newblock \BBOQ Getting access to what goes on in people's heads? reflections
  on the think-aloud technique\BBCQ\
\newblock In {\Bem Proceedings of the Second Nordic Conference on
  Human-Computer Interaction}, NordiCHI '02, \BPG\ 101–110, New York, NY,
  USA. Association for Computing Machinery.

\bibitem[\protect\BCAY{Noble\ \BBA\ Heale}{Noble\ \BBA\ Heale}{2019}]{Noble67}
Noble, H.\BBACOMMA\  \BBA\ Heale, R. \BBOP2019\BBCP.
\newblock \BBOQ Triangulation in research, with examples\BBCQ\
\newblock {\Bem Evidence-Based Nursing}, {\Bem 22\/}(3), 67--68.

\bibitem[\protect\BCAY{Novikova, Du{\v{s}}ek, Curry,\ \BBA\ Rieser}{Novikova
  et~al.}{2017}]{novikova_etal_2017}
Novikova, J., Du{\v{s}}ek, O., Curry, A.~C., \BBA\ Rieser, V. \BBOP2017\BBCP.
\newblock \BBOQ Why we need new evaluation metrics for {NLG}\BBCQ\
\newblock In {\Bem Proceedings of the 2017 Conference on Empirical Methods in
  Natural Language Processing}, \BPGS\ 2241--2252.

\bibitem[\protect\BCAY{Nuruzzaman\ \BBA\ Hussain}{Nuruzzaman\ \BBA\
  Hussain}{2020}]{nuruzzaman2020intellibot}
Nuruzzaman, M.\BBACOMMA\  \BBA\ Hussain, O.~K. \BBOP2020\BBCP.
\newblock \BBOQ Intellibot: a dialogue-based chatbot for the insurance
  industry\BBCQ\
\newblock {\Bem Knowledge-Based Systems}, {\Bem 196}, 105810.

\bibitem[\protect\BCAY{Okanovi{\'c}, Beck, Merz, Zorn, Merino, van Hoorn,\
  \BBA\ Beck}{Okanovi{\'c} et~al.}{2020}]{okanovic2020can}
Okanovi{\'c}, D., Beck, S., Merz, L., Zorn, C., Merino, L., van Hoorn, A.,
  \BBA\ Beck, F. \BBOP2020\BBCP.
\newblock \BBOQ Can a chatbot support software engineers with load testing?
  approach and experiences\BBCQ\
\newblock In {\Bem Proceedings of the ACM/SPEC international conference on
  performance engineering}, \BPGS\ 120--129.

\bibitem[\protect\BCAY{Okonkwo\ \BBA\ Ade-Ibijola}{Okonkwo\ \BBA\
  Ade-Ibijola}{2020}]{Okonkwo2020python}
Okonkwo, C.~W.\BBACOMMA\  \BBA\ Ade-Ibijola, A. \BBOP2020\BBCP.
\newblock \BBOQ Python-bot: A chatbot for teaching python programming.\BBCQ\
\newblock {\Bem Engineering Letters}, {\Bem 29\/}(1).

\bibitem[\protect\BCAY{Oliver}{Oliver}{1980}]{oliver1980cognitive}
Oliver, R.~L. \BBOP1980\BBCP.
\newblock \BBOQ A cognitive model of the antecedents and consequences of
  satisfaction decisions\BBCQ\
\newblock {\Bem Journal of Marketing Research}, {\Bem 17\/}(4), 460--469.

\bibitem[\protect\BCAY{Oliver\ \BBA\ DeSarbo}{Oliver\ \BBA\
  DeSarbo}{1988}]{oliver1988response}
Oliver, R.~L.\BBACOMMA\  \BBA\ DeSarbo, W.~S. \BBOP1988\BBCP.
\newblock \BBOQ Response determinants in satisfaction judgments\BBCQ\
\newblock {\Bem Journal of Consumer Research}, {\Bem 14\/}(4), 495--507.

\bibitem[\protect\BCAY{Oniani\ \BBA\ Wang}{Oniani\ \BBA\
  Wang}{2020}]{oniani2020qualitative}
Oniani, D.\BBACOMMA\  \BBA\ Wang, Y. \BBOP2020\BBCP.
\newblock \BBOQ A qualitative evaluation of language models on automatic
  question-answering for covid-19\BBCQ\
\newblock In {\Bem Proceedings of the 11th ACM International Conference on
  Bioinformatics, Computational Biology and Health Informatics}, \BPGS\ 1--9.

\bibitem[\protect\BCAY{OpenAI}{OpenAI}{2022}]{chatgpt}
OpenAI \BBOP2022\BBCP.
\newblock \BBOQ Introducing chatgpt\BBCQ\
\newblock First published on November 30 2022. Last checked on 26 February
  2024.

\bibitem[\protect\BCAY{Orden-Mej{\'\i}a\ \BBA\ Huertas}{Orden-Mej{\'\i}a\ \BBA\
  Huertas}{2021}]{orden2021analysis}
Orden-Mej{\'\i}a, M.\BBACOMMA\  \BBA\ Huertas, A. \BBOP2021\BBCP.
\newblock \BBOQ Analysis of the attributes of smart tourism technologies in
  destination chatbots that influence tourist satisfaction\BBCQ\
\newblock {\Bem Current Issues in Tourism}, {\Bem 25\/}(17), 1--16.

\bibitem[\protect\BCAY{Ouzzani, Hammady, Fedorowicz,\ \BBA\ Elmagarmid}{Ouzzani
  et~al.}{2016}]{ouzzani2016rayyan}
Ouzzani, M., Hammady, H., Fedorowicz, Z., \BBA\ Elmagarmid, A. \BBOP2016\BBCP.
\newblock \BBOQ Rayyan—a web and mobile app for systematic reviews\BBCQ\
\newblock {\Bem Systematic reviews}, {\Bem 5\/}(1), 1--10.

\bibitem[\protect\BCAY{Paek}{Paek}{2001}]{paek-2001-empirical}
Paek, T. \BBOP2001\BBCP.
\newblock \BBOQ Empirical methods for evaluating dialog systems\BBCQ\
\newblock In {\Bem Proceedings of the {ACL} 2001 Workshop on Evaluation
  Methodologies for Language and Dialogue Systems}.

\bibitem[\protect\BCAY{Papineni, Roukos, Ward,\ \BBA\ Zhu}{Papineni
  et~al.}{2002}]{papineni2002bleu}
Papineni, K., Roukos, S., Ward, T., \BBA\ Zhu, W.-J. \BBOP2002\BBCP.
\newblock \BBOQ Bleu: a method for automatic evaluation of machine
  translation\BBCQ\
\newblock In {\Bem Proceedings of the 40th annual meeting of the Association
  for Computational Linguistics}, \BPGS\ 311--318.

\bibitem[\protect\BCAY{Parasuraman, Zeithaml,\ \BBA\ Berry}{Parasuraman
  et~al.}{1988}]{parasuraman1988servqual}
Parasuraman, A., Zeithaml, V.~A., \BBA\ Berry, L. \BBOP1988\BBCP.
\newblock \BBOQ Servqual: A multiple-item scale for measuring consumer
  perceptions of service quality\BBCQ\
\newblock {\Bem 1988}, {\Bem 64\/}(1), 12--40.

\bibitem[\protect\BCAY{Pavone, Meyer-Waarden,\ \BBA\ Munzel}{Pavone
  et~al.}{2023}]{pavone2023rage}
Pavone, G., Meyer-Waarden, L., \BBA\ Munzel, A. \BBOP2023\BBCP.
\newblock \BBOQ Rage against the machine: experimental insights into
  customers’ negative emotional responses, attributions of responsibility,
  and coping strategies in artificial intelligence--based service
  failures\BBCQ\
\newblock {\Bem Journal of Interactive Marketing}, {\Bem 58\/}(1), 52--71.

\bibitem[\protect\BCAY{Peng, Li, Zhang, Zhu, Li,\ \BBA\ Gao}{Peng
  et~al.}{2021}]{peng-etal-2021-raddle}
Peng, B., Li, C., Zhang, Z., Zhu, C., Li, J., \BBA\ Gao, J. \BBOP2021\BBCP.
\newblock \BBOQ {RADDLE}: An evaluation benchmark and analysis platform for
  robust task-oriented dialog systems\BBCQ\
\newblock In {\Bem Proceedings of the 59th Annual Meeting of the Association
  for Computational Linguistics and the 11th International Joint Conference on
  Natural Language Processing (Volume 1: Long Papers)}, \BPGS\ 4418--4429,
  Online. Association for Computational Linguistics.

\bibitem[\protect\BCAY{Peng\ \BBA\ Ma}{Peng\ \BBA\ Ma}{2019}]{peng2019survey}
Peng, Z.\BBACOMMA\  \BBA\ Ma, X. \BBOP2019\BBCP.
\newblock \BBOQ A survey on construction and enhancement methods in service
  chatbots design\BBCQ\
\newblock {\Bem CCF Transactions on Pervasive Computing and Interaction}, {\Bem
  1\/}(3), 204--223.

\bibitem[\protect\BCAY{Pereira\ \BBA\ D{\'\i}az}{Pereira\ \BBA\
  D{\'\i}az}{2018}]{pereira2018quality}
Pereira, J.\BBACOMMA\  \BBA\ D{\'\i}az, O. \BBOP2018\BBCP.
\newblock \BBOQ A quality analysis of facebook messenger's most popular
  chatbots\BBCQ\
\newblock In {\Bem Proceedings of the 33rd annual ACM symposium on applied
  computing}, \BPGS\ 2144--2150.

\bibitem[\protect\BCAY{Piao, Kim, Ryu,\ \BBA\ Lee}{Piao
  et~al.}{2020}]{piao2020development}
Piao, M., Kim, J., Ryu, H., \BBA\ Lee, H. \BBOP2020\BBCP.
\newblock \BBOQ Development and usability evaluation of a healthy lifestyle
  coaching chatbot using a habit formation model\BBCQ\
\newblock {\Bem Healthcare Informatics Research}, {\Bem 26\/}(4), 255--264.

\bibitem[\protect\BCAY{Piau, Crissey, Brechemier, Balardy,\ \BBA\
  Nourhashemi}{Piau et~al.}{2019}]{piau2019smartphone}
Piau, A., Crissey, R., Brechemier, D., Balardy, L., \BBA\ Nourhashemi, F.
  \BBOP2019\BBCP.
\newblock \BBOQ A smartphone chatbot application to optimize monitoring of
  older patients with cancer\BBCQ\
\newblock {\Bem International journal of medical informatics}, {\Bem 128},
  18--23.

\bibitem[\protect\BCAY{Pradhan\ \BBA\ Todi}{Pradhan\ \BBA\
  Todi}{2023}]{pradhan-todi-2023-understanding}
Pradhan, A.\BBACOMMA\  \BBA\ Todi, K. \BBOP2023\BBCP.
\newblock \BBOQ Understanding large language model based metrics for text
  summarization\BBCQ\
\newblock In Deutsch, D., Dror, R., Eger, S., Gao, Y., Leiter, C., Opitz, J.,
  \BBA\ R{\"u}ckl{\'e}, A.\BEDS, {\Bem Proceedings of the 4th Workshop on
  Evaluation and Comparison of NLP Systems}, \BPGS\ 149--155, Bali, Indonesia.
  Association for Computational Linguistics.

\bibitem[\protect\BCAY{Prasad, Blagsvedt, Pochiraju,\ \BBA\ Medhi~Thies}{Prasad
  et~al.}{2019}]{prasad2019dara}
Prasad, A., Blagsvedt, S., Pochiraju, T., \BBA\ Medhi~Thies, I. \BBOP2019\BBCP.
\newblock \BBOQ Dara: A chatbot to help indian artists and designers discover
  international opportunities\BBCQ\
\newblock In {\Bem Proceedings of the 2019 Conference on Creativity and
  Cognition}, C\&C '19, \BPG\ 626–632, New York, NY, USA. Association for
  Computing Machinery.

\bibitem[\protect\BCAY{Pricilla, Lestari,\ \BBA\ Dharma}{Pricilla
  et~al.}{2018}]{pricilla2018designing}
Pricilla, C., Lestari, D.~P., \BBA\ Dharma, D. \BBOP2018\BBCP.
\newblock \BBOQ Designing interaction for chatbot-based conversational commerce
  with user-centered design\BBCQ\
\newblock In {\Bem 2018 5th International Conference on Advanced Informatics:
  Concept Theory and Applications (ICAICTA)}, \BPGS\ 244--249. IEEE.

\bibitem[\protect\BCAY{Przegalinska, Ciechanowski, Stroz, Gloor,\ \BBA\
  Mazurek}{Przegalinska et~al.}{2019}]{przegalinska2019bot}
Przegalinska, A., Ciechanowski, L., Stroz, A., Gloor, P., \BBA\ Mazurek, G.
  \BBOP2019\BBCP.
\newblock \BBOQ In bot we trust: A new methodology of chatbot performance
  measures\BBCQ\
\newblock {\Bem Business Horizons}, {\Bem 62\/}(6), 785--797.

\bibitem[\protect\BCAY{Puron, Pierres, Javelot, Prosa, Gillot, Fernandes,
  Morisseau,\ \BBA\ Frenzel}{Puron et~al.}{2021}]{puron2021p}
Puron, L., Pierres, M., Javelot, M., Prosa, N., Gillot, L., Fernandes, P.,
  Morisseau, V., \BBA\ Frenzel, L. \BBOP2021\BBCP.
\newblock \BBOQ P-135: Evaluation of the support and knowledge provided by the
  chatbot (virtual conversation agent) vik multiple myeloma in the management
  of the disease\BBCQ\
\newblock {\Bem Clinical Lymphoma, Myeloma and Leukemia}, {\Bem 21},
  S108--S109.

\bibitem[\protect\BCAY{Qiu\ \BBA\ Benbasat}{Qiu\ \BBA\
  Benbasat}{2009}]{qiu2009evaluating}
Qiu, L.\BBACOMMA\  \BBA\ Benbasat, I. \BBOP2009\BBCP.
\newblock \BBOQ Evaluating anthropomorphic product recommendation agents: A
  social relationship perspective to designing information systems\BBCQ\
\newblock {\Bem Journal of management information systems}, {\Bem 25\/}(4),
  145--182.

\bibitem[\protect\BCAY{Rapp, Curti,\ \BBA\ Boldi}{Rapp
  et~al.}{2021}]{rapp2021human}
Rapp, A., Curti, L., \BBA\ Boldi, A. \BBOP2021\BBCP.
\newblock \BBOQ The human side of human-chatbot interaction: A systematic
  literature review of ten years of research on text-based chatbots\BBCQ\
\newblock {\Bem International Journal of Human-Computer Studies}, {\Bem 151},
  102630.

\bibitem[\protect\BCAY{Rei, Guerreiro, Pombal, van Stigt, Treviso, Coheur,
  de~Souza,\ \BBA\ Martins}{Rei et~al.}{2023}]{rei2023scaling}
Rei, R., Guerreiro, N.~M., Pombal, J., van Stigt, D., Treviso, M., Coheur, L.,
  de~Souza, J. G.~C., \BBA\ Martins, A. F.~T. \BBOP2023\BBCP.
\newblock \BBOQ Scaling up cometkiwi: Unbabel-ist 2023 submission for the
  quality estimation shared task\BBCQ.

\bibitem[\protect\BCAY{Rei, Stewart, Farinha,\ \BBA\ Lavie}{Rei
  et~al.}{2020}]{rei-etal-2020-comet}
Rei, R., Stewart, C., Farinha, A.~C., \BBA\ Lavie, A. \BBOP2020\BBCP.
\newblock \BBOQ {COMET}: A neural framework for {MT} evaluation\BBCQ\
\newblock In Webber, B., Cohn, T., He, Y., \BBA\ Liu, Y.\BEDS, {\Bem
  Proceedings of the 2020 Conference on Empirical Methods in Natural Language
  Processing (EMNLP)}, \BPGS\ 2685--2702, Online. Association for Computational
  Linguistics.

\bibitem[\protect\BCAY{Rei, Treviso, Guerreiro, Zerva, Farinha, Maroti, C.~de
  Souza, Glushkova, Alves, Coheur, Lavie,\ \BBA\ Martins}{Rei
  et~al.}{2022}]{rei-etal-2022-cometkiwi}
Rei, R., Treviso, M., Guerreiro, N.~M., Zerva, C., Farinha, A.~C., Maroti, C.,
  C.~de Souza, J.~G., Glushkova, T., Alves, D., Coheur, L., Lavie, A., \BBA\
  Martins, A. F.~T. \BBOP2022\BBCP.
\newblock \BBOQ {C}omet{K}iwi: {IST}-unbabel 2022 submission for the quality
  estimation shared task\BBCQ\
\newblock In Koehn, P., Barrault, L., Bojar, O., Bougares, F., Chatterjee, R.,
  Costa-juss{\`a}, M.~R., Federmann, C., Fishel, M., Fraser, A., Freitag, M.,
  Graham, Y., Grundkiewicz, R., Guzman, P., Haddow, B., Huck, M., Jimeno~Yepes,
  A., Kocmi, T., Martins, A., Morishita, M., Monz, C., Nagata, M., Nakazawa,
  T., Negri, M., N{\'e}v{\'e}ol, A., Neves, M., Popel, M., Turchi, M., \BBA\
  Zampieri, M.\BEDS, {\Bem Proceedings of the Seventh Conference on Machine
  Translation (WMT)}, \BPGS\ 634--645, Abu Dhabi, United Arab Emirates
  (Hybrid). Association for Computational Linguistics.

\bibitem[\protect\BCAY{Reiter}{Reiter}{2017}]{reiter2017howto}
Reiter, E. \BBOP2017\BBCP.
\newblock \BBOQ How to do an nlg evaluation: Metrics\BBCQ\
\newblock Published on Ehud Reiter's personal blog. Last accessed 24 October
  2023.

\bibitem[\protect\BCAY{Reiter}{Reiter}{2018}]{reiter-2018-structured}
Reiter, E. \BBOP2018\BBCP.
\newblock \BBOQ A structured review of the validity of {BLEU}\BBCQ\
\newblock {\Bem Computational Linguistics}, {\Bem 44\/}(3), 393--401.

\bibitem[\protect\BCAY{Reiter}{Reiter}{2022}]{reiter22understand}
Reiter, E. \BBOP2022\BBCP.
\newblock \BBOQ We need to understand what users want\BBCQ\
\newblock Published on Ehud Reiter's personal blog. Last accessed 24 October
  2023.

\bibitem[\protect\BCAY{Ren, Castro, Acu{\~n}a,\ \BBA\ de~Lara}{Ren
  et~al.}{2019}]{ren2019evaluation}
Ren, R., Castro, J.~W., Acu{\~n}a, S.~T., \BBA\ de~Lara, J. \BBOP2019\BBCP.
\newblock \BBOQ Evaluation techniques for chatbot usability: A systematic
  mapping study\BBCQ\
\newblock {\Bem International Journal of Software Engineering and Knowledge
  Engineering}, {\Bem 29\/}(11n12), 1673--1702.

\bibitem[\protect\BCAY{Ren, Spina, De~Vries, Bijkerk, Faber,\ \BBA\
  Geraedts}{Ren et~al.}{2020}]{ren2020understanding}
Ren, X., Spina, G., De~Vries, S., Bijkerk, A., Faber, B., \BBA\ Geraedts, A.
  \BBOP2020\BBCP.
\newblock \BBOQ Understanding physician’s experience with conversational
  interfaces during occupational health consultation\BBCQ\
\newblock {\Bem IEEE Access}, {\Bem 8}, 119158--119169.

\bibitem[\protect\BCAY{Rese, Ganster,\ \BBA\ Baier}{Rese
  et~al.}{2020}]{rese2020chatbots}
Rese, A., Ganster, L., \BBA\ Baier, D. \BBOP2020\BBCP.
\newblock \BBOQ Chatbots in retailers’ customer communication: How to measure
  their acceptance?\BBCQ\
\newblock {\Bem Journal of Retailing and Consumer Services}, {\Bem 56}, 102176.

\bibitem[\protect\BCAY{Resnik\ \BBA\ Lin}{Resnik\ \BBA\
  Lin}{2010}]{resnik2010evaluation}
Resnik, P.\BBACOMMA\  \BBA\ Lin, J. \BBOP2010\BBCP.
\newblock {\Bem Evaluation of NLP Systems}, \BCH~11, \BPGS\ 271--295.
\newblock John Wiley \& Sons, Ltd.

\bibitem[\protect\BCAY{Rietz\ \BBA\ Maedche}{Rietz\ \BBA\
  Maedche}{2019}]{Rietz2019ladderbot}
Rietz, T.\BBACOMMA\  \BBA\ Maedche, A. \BBOP2019\BBCP.
\newblock \BBOQ Ladderbot: A requirements self-elicitation system\BBCQ\
\newblock In {\Bem 2019 IEEE 27th International Requirements Engineering
  Conference (RE)}, \BPGS\ 357--362. IEEE.

\bibitem[\protect\BCAY{Riyadh\ \BBA\ Shafiq}{Riyadh\ \BBA\
  Shafiq}{2023}]{riyadh2023towards}
Riyadh, M.\BBACOMMA\  \BBA\ Shafiq, M.~O. \BBOP2023\BBCP.
\newblock \BBOQ Towards automatic evaluation of nlg tasks using conversational
  large language models\BBCQ\
\newblock In {\Bem IFIP International Conference on Artificial Intelligence
  Applications and Innovations}, \BPGS\ 425--437. Springer.

\bibitem[\protect\BCAY{Rodr{\'\i}guez-Cantelar, Zhang, Tang, Shi, Ghazarian,
  Sedoc, Fernando~D{'}Haro,\ \BBA\ Rudnicky}{Rodr{\'\i}guez-Cantelar
  et~al.}{2023}]{rodriguez-cantelar-etal-2023-overview}
Rodr{\'\i}guez-Cantelar, M., Zhang, C., Tang, C., Shi, K., Ghazarian, S.,
  Sedoc, J., Fernando~D{'}Haro, L., \BBA\ Rudnicky, A.~I. \BBOP2023\BBCP.
\newblock \BBOQ Overview of robust and multilingual automatic evaluation
  metricsfor open-domain dialogue systems at {DSTC} 11 track 4\BBCQ\
\newblock In Chen, Y.-N., Crook, P., Galley, M., Ghazarian, S., Gunasekara, C.,
  Gupta, R., Hedayatnia, B., Kottur, S., Moon, S., \BBA\ Zhang, C.\BEDS, {\Bem
  Proceedings of The Eleventh Dialog System Technology Challenge}, \BPGS\
  260--273, Prague, Czech Republic. Association for Computational Linguistics.

\bibitem[\protect\BCAY{Roque, de~Souza, do~Nascimento, de~Campos~Filho,
  de~Melo~Queiroz,\ \BBA\ Santos}{Roque et~al.}{2021}]{roque2021content}
Roque, G. d. S.~L., de~Souza, R.~R., do~Nascimento, J. W.~A., de~Campos~Filho,
  A.~S., de~Melo~Queiroz, S.~R., \BBA\ Santos, I. C. R.~V. \BBOP2021\BBCP.
\newblock \BBOQ Content validation and usability of a chatbot of guidelines for
  wound dressing\BBCQ\
\newblock {\Bem International Journal of Medical Informatics}, {\Bem 151},
  104473.

\bibitem[\protect\BCAY{Santhanam\ \BBA\ Shaikh}{Santhanam\ \BBA\
  Shaikh}{2019}]{santhanam-shaikh-2019-towards}
Santhanam, S.\BBACOMMA\  \BBA\ Shaikh, S. \BBOP2019\BBCP.
\newblock \BBOQ Towards best experiment design for evaluating dialogue system
  output\BBCQ\
\newblock In {\Bem Proceedings of the 12th International Conference on Natural
  Language Generation}, \BPGS\ 88--94, Tokyo, Japan. Association for
  Computational Linguistics.

\bibitem[\protect\BCAY{Schiavone, Quinn,\ \BBA\ Vazire}{Schiavone
  et~al.}{2023}]{schiavone_quinn_vazire_2023}
Schiavone, S.~R., Quinn, K.~A., \BBA\ Vazire, S. \BBOP2023\BBCP.
\newblock \BBOQ A consensus-based tool for evaluating threats to the validity
  of empirical research\BBCQ.

\bibitem[\protect\BCAY{Schlangen}{Schlangen}{2021}]{schlangen-2021-targeting}
Schlangen, D. \BBOP2021\BBCP.
\newblock \BBOQ Targeting the benchmark: On methodology in current natural
  language processing research\BBCQ\
\newblock In Zong, C., Xia, F., Li, W., \BBA\ Navigli, R.\BEDS, {\Bem
  Proceedings of the 59th Annual Meeting of the Association for Computational
  Linguistics and the 11th International Joint Conference on Natural Language
  Processing (Volume 2: Short Papers)}, \BPGS\ 670--674, Online. Association
  for Computational Linguistics.

\bibitem[\protect\BCAY{Schmidlen, Schwartz, DiLoreto, Kirchner,\ \BBA\
  Sturm}{Schmidlen et~al.}{2019}]{schmidlen2019patient}
Schmidlen, T., Schwartz, M., DiLoreto, K., Kirchner, H.~L., \BBA\ Sturm, A.~C.
  \BBOP2019\BBCP.
\newblock \BBOQ Patient assessment of chatbots for the scalable delivery of
  genetic counseling\BBCQ\
\newblock {\Bem Journal of genetic counseling}, {\Bem 28\/}(6), 1166--1177.

\bibitem[\protect\BCAY{Schumaker, Ginsburg, Chen,\ \BBA\ Liu}{Schumaker
  et~al.}{2007}]{schumaker2007evaluation}
Schumaker, R.~P., Ginsburg, M., Chen, H., \BBA\ Liu, Y. \BBOP2007\BBCP.
\newblock \BBOQ An evaluation of the chat and knowledge delivery components of
  a low-level dialog system: The az-alice experiment\BBCQ\
\newblock {\Bem Decision Support Systems}, {\Bem 42\/}(4), 2236--2246.

\bibitem[\protect\BCAY{Sedoc, Ippolito, Kirubarajan, Thirani, Ungar,\ \BBA\
  Callison-Burch}{Sedoc et~al.}{2018}]{sedoc-etal-2018-chateval}
Sedoc, J., Ippolito, D., Kirubarajan, A., Thirani, J., Ungar, L., \BBA\
  Callison-Burch, C. \BBOP2018\BBCP.
\newblock \BBOQ {C}hat{E}val: A tool for the systematic evaluation of
  chatbots\BBCQ\
\newblock In Alonso, J.~M., Catala, A., \BBA\ Theune, M.\BEDS, {\Bem
  Proceedings of the Workshop on Intelligent Interactive Systems and Language
  Generation (2{IS}{\&}{NLG})}, \BPGS\ 42--44, Tilburg, the Netherlands.
  Association for Computational Linguistics.

\bibitem[\protect\BCAY{Sedoc, Ippolito, Kirubarajan, Thirani, Ungar,\ \BBA\
  Callison-Burch}{Sedoc et~al.}{2019}]{sedoc-etal-2019-chateval}
Sedoc, J., Ippolito, D., Kirubarajan, A., Thirani, J., Ungar, L., \BBA\
  Callison-Burch, C. \BBOP2019\BBCP.
\newblock \BBOQ {C}hat{E}val: A tool for chatbot evaluation\BBCQ\
\newblock In {\Bem Proceedings of the 2019 Conference of the North {A}merican
  Chapter of the Association for Computational Linguistics (Demonstrations)},
  \BPGS\ 60--65, Minneapolis, Minnesota. Association for Computational
  Linguistics.

\bibitem[\protect\BCAY{Sedoc\ \BBA\ Ungar}{Sedoc\ \BBA\
  Ungar}{2020}]{sedoc-ungar-2020-item}
Sedoc, J.\BBACOMMA\  \BBA\ Ungar, L. \BBOP2020\BBCP.
\newblock \BBOQ Item response theory for efficient human evaluation of
  chatbots\BBCQ\
\newblock In {\Bem Proceedings of the First Workshop on Evaluation and
  Comparison of NLP Systems}, \BPGS\ 21--33, Online. Association for
  Computational Linguistics.

\bibitem[\protect\BCAY{Sekulić, Terragni, Guimarães, Khau, Guedes,
  Filipavicius, Manso,\ \BBA\ Mathis}{Sekulić
  et~al.}{2024}]{sekulic2024reliable}
Sekulić, I., Terragni, S., Guimarães, V., Khau, N., Guedes, B., Filipavicius,
  M., Manso, A.~F., \BBA\ Mathis, R. \BBOP2024\BBCP.
\newblock \BBOQ Reliable llm-based user simulator for task-oriented dialogue
  systems\BBCQ.

\bibitem[\protect\BCAY{Sellam, Das,\ \BBA\ Parikh}{Sellam
  et~al.}{2020}]{sellam-etal-2020-bleurt}
Sellam, T., Das, D., \BBA\ Parikh, A. \BBOP2020\BBCP.
\newblock \BBOQ {BLEURT}: Learning robust metrics for text generation\BBCQ\
\newblock In Jurafsky, D., Chai, J., Schluter, N., \BBA\ Tetreault, J.\BEDS,
  {\Bem Proceedings of the 58th Annual Meeting of the Association for
  Computational Linguistics}, \BPGS\ 7881--7892, Online. Association for
  Computational Linguistics.

\bibitem[\protect\BCAY{Sensuse, Dhevanty, Rahmanasari, Permatasari, Putra,
  Lusa, Misbah,\ \BBA\ Prima}{Sensuse et~al.}{2019}]{sensuse2019chatbot}
Sensuse, D.~I., Dhevanty, V., Rahmanasari, E., Permatasari, D., Putra, B.~E.,
  Lusa, J.~S., Misbah, M., \BBA\ Prima, P. \BBOP2019\BBCP.
\newblock \BBOQ Chatbot evaluation as knowledge application: a case study of pt
  abc\BBCQ\
\newblock In {\Bem 2019 11th International Conference on Information Technology
  and Electrical Engineering (ICITEE)}, \BPGS\ 1--6. IEEE.

\bibitem[\protect\BCAY{Shackel}{Shackel}{2009}]{shackel2009usability}
Shackel, B. \BBOP2009\BBCP.
\newblock \BBOQ Usability--context, framework, definition, design and
  evaluation\BBCQ\
\newblock {\Bem Interacting with computers}, {\Bem 21\/}(5-6), 339--346.

\bibitem[\protect\BCAY{Shadish, Cook,\ \BBA\ Campbell}{Shadish
  et~al.}{2002}]{shadish2002experimental}
Shadish, W.~R., Cook, T.~D., \BBA\ Campbell, D.~T. \BBOP2002\BBCP.
\newblock {\Bem Experimental and Quasi-experimental Designs for Generalized
  Causal Inference}.
\newblock Experimental and Quasi-experimental Designs for Generalized Causal
  Inference. Houghton Mifflin.

\bibitem[\protect\BCAY{Shi, Li, Sahay,\ \BBA\ Yu}{Shi
  et~al.}{2021}]{Shi-etal-2021-refine-imitate}
Shi, W., Li, Y., Sahay, S., \BBA\ Yu, Z. \BBOP2021\BBCP.
\newblock \BBOQ Refine and imitate: Reducing repetition and inconsistency in
  persuasion dialogues via reinforcement learning and human demonstration\BBCQ\
\newblock In {\Bem Findings of the Association for Computational Linguistics:
  EMNLP 2021}, \BPGS\ 3478--3492, Punta Cana, Dominican Republic. Association
  for Computational Linguistics.

\bibitem[\protect\BCAY{Shimorina\ \BBA\ Belz}{Shimorina\ \BBA\
  Belz}{2022}]{shimorina-belz-2022-human}
Shimorina, A.\BBACOMMA\  \BBA\ Belz, A. \BBOP2022\BBCP.
\newblock \BBOQ The human evaluation datasheet: A template for recording
  details of human evaluation experiments in {NLP}\BBCQ\
\newblock In {\Bem Proceedings of the 2nd Workshop on Human Evaluation of NLP
  Systems (HumEval)}, \BPGS\ 54--75, Dublin, Ireland. Association for
  Computational Linguistics.

\bibitem[\protect\BCAY{Song, Chen, Niu,\ \BBA\ Haihong}{Song
  et~al.}{2019}]{song2019task}
Song, M., Chen, Z., Niu, P., \BBA\ Haihong, E. \BBOP2019\BBCP.
\newblock \BBOQ Task-oriented dialogue system based on reinforcement
  learning\BBCQ\
\newblock In {\Bem 2019 IEEE 10th International Conference on Software
  Engineering and Service Science (ICSESS)}, \BPGS\ 1--6. IEEE.

\bibitem[\protect\BCAY{Specia, Scarton,\ \BBA\ Paetzold}{Specia
  et~al.}{2018}]{qualityestimation}
Specia, L., Scarton, C., \BBA\ Paetzold, G.~H. \BBOP2018\BBCP.
\newblock {\Bem Quality Estimation for Machine Translation}.
\newblock Synthesis Lectures on Human Language Technologies. Springer Cham.

\bibitem[\protect\BCAY{Spellman, Gilbert,\ \BBA\ Corker}{Spellman
  et~al.}{2018}]{spellman2018open}
Spellman, B.~A., Gilbert, E.~A., \BBA\ Corker, K.~S. \BBOP2018\BBCP.
\newblock \BBOQ Open science\BBCQ\
\newblock {\Bem Stevens' Handbook of Experimental Psychology and Cognitive
  Neuroscience}, {\Bem 5}, 1--47.

\bibitem[\protect\BCAY{Sperl{\'\i}}{Sperl{\'\i}}{2020}]{sperli2020deep}
Sperl{\'\i}, G. \BBOP2020\BBCP.
\newblock \BBOQ A deep learning based chatbot for cultural heritage\BBCQ\
\newblock In {\Bem Proceedings of the 35th Annual ACM Symposium on Applied
  Computing}, \BPGS\ 935--937.

\bibitem[\protect\BCAY{Strauss\ \BBA\ Smith}{Strauss\ \BBA\
  Smith}{2009}]{strauss2009construct}
Strauss, M.~E.\BBACOMMA\  \BBA\ Smith, G.~T. \BBOP2009\BBCP.
\newblock \BBOQ Construct validity: advances in theory and methodology.\BBCQ\
\newblock {\Bem Annu Rev Clin Psychol}, {\Bem 5}, 1--25.

\bibitem[\protect\BCAY{Su, Shen, Xiao, Zhang, Chang, Zhang, Niu,\ \BBA\
  Zhou}{Su et~al.}{2020}]{su-etal-2020-moviechats}
Su, H., Shen, X., Xiao, Z., Zhang, Z., Chang, E., Zhang, C., Niu, C., \BBA\
  Zhou, J. \BBOP2020\BBCP.
\newblock \BBOQ {M}ovie{C}hats: Chat like humans in a closed domain\BBCQ\
\newblock In {\Bem Proceedings of the 2020 Conference on Empirical Methods in
  Natural Language Processing (EMNLP)}, \BPGS\ 6605--6619, Online. Association
  for Computational Linguistics.

\bibitem[\protect\BCAY{Subramonian, Yuan, Daum{\'e}~III,\ \BBA\
  Blodgett}{Subramonian et~al.}{2023}]{subramonian-etal-2023-takes}
Subramonian, A., Yuan, X., Daum{\'e}~III, H., \BBA\ Blodgett, S.~L.
  \BBOP2023\BBCP.
\newblock \BBOQ It takes two to tango: Navigating conceptualizations of {NLP}
  tasks and measurements of performance\BBCQ\
\newblock In Rogers, A., Boyd-Graber, J., \BBA\ Okazaki, N.\BEDS, {\Bem
  Findings of the Association for Computational Linguistics: ACL 2023}, \BPGS\
  3234--3279, Toronto, Canada. Association for Computational Linguistics.

\bibitem[\protect\BCAY{Sugawara, Stenetorp,\ \BBA\ Aizawa}{Sugawara
  et~al.}{2021}]{sugawara-etal-2021-benchmarking}
Sugawara, S., Stenetorp, P., \BBA\ Aizawa, A. \BBOP2021\BBCP.
\newblock \BBOQ Benchmarking machine reading comprehension: A psychological
  perspective\BBCQ\
\newblock In Merlo, P., Tiedemann, J., \BBA\ Tsarfaty, R.\BEDS, {\Bem
  Proceedings of the 16th Conference of the European Chapter of the Association
  for Computational Linguistics: Main Volume}, \BPGS\ 1592--1612, Online.
  Association for Computational Linguistics.

\bibitem[\protect\BCAY{Sugawara\ \BBA\ Tsugita}{Sugawara\ \BBA\
  Tsugita}{2023}]{sugawara-tsugita-2023-degrees}
Sugawara, S.\BBACOMMA\  \BBA\ Tsugita, S. \BBOP2023\BBCP.
\newblock \BBOQ On degrees of freedom in defining and testing natural language
  understanding\BBCQ\
\newblock In Rogers, A., Boyd-Graber, J., \BBA\ Okazaki, N.\BEDS, {\Bem
  Findings of the Association for Computational Linguistics: ACL 2023}, \BPGS\
  13625--13649, Toronto, Canada. Association for Computational Linguistics.

\bibitem[\protect\BCAY{Sugisaki\ \BBA\ Bleiker}{Sugisaki\ \BBA\
  Bleiker}{2020}]{sugisaki2020usability}
Sugisaki, K.\BBACOMMA\  \BBA\ Bleiker, A. \BBOP2020\BBCP.
\newblock \BBOQ Usability guidelines and evaluation criteria for conversational
  user interfaces: a heuristic and linguistic approach\BBCQ\
\newblock In {\Bem Proceedings of the Conference on Mensch und Computer},
  \BPGS\ 309--319.

\bibitem[\protect\BCAY{Sulem, Abend,\ \BBA\ Rappoport}{Sulem
  et~al.}{2018}]{sulem2018bleu}
Sulem, E., Abend, O., \BBA\ Rappoport, A. \BBOP2018\BBCP.
\newblock \BBOQ {BLEU} is not suitable for the evaluation of text
  simplification\BBCQ\
\newblock In {\Bem Proceedings of the 2018 Conference on Empirical Methods in
  Natural Language Processing}, \BPGS\ 738--744, Brussels, Belgium. Association
  for Computational Linguistics.

\bibitem[\protect\BCAY{Sun, Williams,\ \BBA\ Hupkes}{Sun
  et~al.}{2023}]{sun2023validity}
Sun, K., Williams, A., \BBA\ Hupkes, D. \BBOP2023\BBCP.
\newblock \BBOQ The validity of evaluation results: Assessing concurrence
  across compositionality benchmarks\BBCQ.

\bibitem[\protect\BCAY{Syv{\"a}nen\ \BBA\ Valentini}{Syv{\"a}nen\ \BBA\
  Valentini}{2020}]{syvanen2020conversational}
Syv{\"a}nen, S.\BBACOMMA\  \BBA\ Valentini, C. \BBOP2020\BBCP.
\newblock \BBOQ Conversational agents in online organization--stakeholder
  interactions: a state-of-the-art analysis and implications for further
  research\BBCQ\
\newblock {\Bem Journal of Communication Management}, {\Bem 24\/}(4), 339--362.

\bibitem[\protect\BCAY{Takanobu, Zhu, Li, Peng, Gao,\ \BBA\ Huang}{Takanobu
  et~al.}{2020}]{takanobu-etal-2020-goal}
Takanobu, R., Zhu, Q., Li, J., Peng, B., Gao, J., \BBA\ Huang, M.
  \BBOP2020\BBCP.
\newblock \BBOQ Is your goal-oriented dialog model performing really well?
  empirical analysis of system-wise evaluation\BBCQ\
\newblock In {\Bem Proceedings of the 21th Annual Meeting of the Special
  Interest Group on Discourse and Dialogue}, \BPGS\ 297--310, 1st virtual
  meeting. Association for Computational Linguistics.

\bibitem[\protect\BCAY{Theosaksomo\ \BBA\ Widyantoro}{Theosaksomo\ \BBA\
  Widyantoro}{2019}]{theosaksomo2019conversational}
Theosaksomo, D.\BBACOMMA\  \BBA\ Widyantoro, D.~H. \BBOP2019\BBCP.
\newblock \BBOQ Conversational recommender system chatbot based on functional
  requirement\BBCQ\
\newblock In {\Bem 2019 IEEE 13th International Conference on Telecommunication
  Systems, Services, and Applications (TSSA)}, \BPGS\ 154--159. IEEE.

\bibitem[\protect\BCAY{Thompson, Yarnold, Williams,\ \BBA\ Adams}{Thompson
  et~al.}{1996}]{thompson1996effects}
Thompson, D.~A., Yarnold, P.~R., Williams, D.~R., \BBA\ Adams, S.~L.
  \BBOP1996\BBCP.
\newblock \BBOQ Effects of actual waiting time, perceived waiting time,
  information delivery, and expressive quality on patient satisfaction in the
  emergency department\BBCQ\
\newblock {\Bem Annals of emergency medicine}, {\Bem 28\/}(6), 657--665.

\bibitem[\protect\BCAY{Thomson\ \BBA\ Reiter}{Thomson\ \BBA\
  Reiter}{2021}]{thomson-reiter-2021-generation}
Thomson, C.\BBACOMMA\  \BBA\ Reiter, E. \BBOP2021\BBCP.
\newblock \BBOQ Generation challenges: Results of the accuracy evaluation
  shared task\BBCQ\
\newblock In Belz, A., Fan, A., Reiter, E., \BBA\ Sripada, Y.\BEDS, {\Bem
  Proceedings of the 14th International Conference on Natural Language
  Generation}, \BPGS\ 240--248, Aberdeen, Scotland, UK. Association for
  Computational Linguistics.

\bibitem[\protect\BCAY{Thomson, Reiter,\ \BBA\ Sundararajan}{Thomson
  et~al.}{2023}]{thomson2023evaluating}
Thomson, C., Reiter, E., \BBA\ Sundararajan, B. \BBOP2023\BBCP.
\newblock \BBOQ Evaluating factual accuracy in complex data-to-text\BBCQ\
\newblock {\Bem Computer Speech \& Language}, {\Bem 80}, 101482.

\bibitem[\protect\BCAY{Thurmond}{Thurmond}{2001}]{thurmond2001point}
Thurmond, V.~A. \BBOP2001\BBCP.
\newblock \BBOQ The point of triangulation\BBCQ\
\newblock {\Bem Journal of Nursing Scholarship}, {\Bem 33\/}(3), 253--258.

\bibitem[\protect\BCAY{Trapero, Ilao,\ \BBA\ Lacaza}{Trapero
  et~al.}{2020}]{trapero2020integrated}
Trapero, H., Ilao, J., \BBA\ Lacaza, R. \BBOP2020\BBCP.
\newblock \BBOQ An integrated theory for chatbot use in air travel:
  Questionnaire development and validation\BBCQ\
\newblock In {\Bem 2020 IEEE REGION 10 CONFERENCE (TENCON)}, \BPGS\ 652--657.
  IEEE.

\bibitem[\protect\BCAY{Treadwell\ \BBA\ Davis}{Treadwell\ \BBA\
  Davis}{2020}]{Treadwell_Davis_2020}
Treadwell, D.\BBACOMMA\  \BBA\ Davis, A. \BBOP2020\BBCP.
\newblock {\Bem Introducing communication research: Paths of inquiry}.
\newblock SAGE.

\bibitem[\protect\BCAY{Tsai, Yang, Wang, Yang, Kang, et~al.}{Tsai
  et~al.}{2022}]{tsai2022sema}
Tsai, M.-H., Yang, C.-H., Wang, C.-H., Yang, I., Kang, S.-C., et~al.
  \BBOP2022\BBCP.
\newblock \BBOQ Sema: A site equipment management assistant for construction
  management\BBCQ\
\newblock {\Bem KSCE Journal of Civil Engineering}, {\Bem 26\/}(3), 1144--1162.

\bibitem[\protect\BCAY{Urban, Sultan,\ \BBA\ Qualls}{Urban
  et~al.}{1998}]{urban1998trust}
Urban, G.~L., Sultan, F., \BBA\ Qualls, W.~J. \BBOP1998\BBCP.
\newblock \BBOQ Trust based marketing on the internet\BBCQ\
\newblock Working Paper (WP 4035-98). Sloan School of Management, Massachusetts
  Institute of Technology.

\bibitem[\protect\BCAY{van Deemter\ \BBA\ Reiter}{van Deemter\ \BBA\
  Reiter}{2018}]{deemter2018lying}
van Deemter, K.\BBACOMMA\  \BBA\ Reiter, E. \BBOP2018\BBCP.
\newblock \BBOQ Lying and computational linguistics\BBCQ\
\newblock In Meibauer, J.\BED, {\Bem The Oxford Handbook of Lying}. Oxford
  University Press.

\bibitem[\protect\BCAY{van~den Broeck, Zarouali,\ \BBA\ Poels}{van~den Broeck
  et~al.}{2019}]{van2019chatbot}
van~den Broeck, E., Zarouali, B., \BBA\ Poels, K. \BBOP2019\BBCP.
\newblock \BBOQ Chatbot advertising effectiveness: When does the message get
  through?\BBCQ\
\newblock {\Bem Computers in Human Behavior}, {\Bem 98}, 150--157.

\bibitem[\protect\BCAY{van~der Goot, Hafkamp,\ \BBA\ Dankfort}{van~der Goot
  et~al.}{2021}]{goot2021customer}
van~der Goot, M.~J., Hafkamp, L., \BBA\ Dankfort, Z. \BBOP2021\BBCP.
\newblock \BBOQ Customer service chatbots: A qualitative interview study into
  the communication journey of customers\BBCQ\
\newblock In F{\o}lstad, A., Araujo, T., Papadopoulos, S., Law, E. L.-C.,
  Luger, E., Goodwin, M., \BBA\ Brandtzaeg, P.~B.\BEDS, {\Bem Chatbot Research
  and Design}, \BPGS\ 190--204, Cham. Springer International Publishing.

\bibitem[\protect\BCAY{{van der Lee}, Gatt, {van Miltenburg},\ \BBA\
  Krahmer}{{van der Lee} et~al.}{2021}]{VANDERLEE2021101151}
{van der Lee}, C., Gatt, A., {van Miltenburg}, E., \BBA\ Krahmer, E.
  \BBOP2021\BBCP.
\newblock \BBOQ Human evaluation of automatically generated text: Current
  trends and best practice guidelines\BBCQ\
\newblock {\Bem Computer Speech \& Language}, {\Bem 67}, 101151.

\bibitem[\protect\BCAY{van~der Meulen}{van~der Meulen}{2023}]{gartnerrisks}
van~der Meulen, R. \BBOP2023\BBCP.
\newblock \BBOQ Gartner identifies six chatgpt risks legal and compliance
  leaders must evaluate\BBCQ\
\newblock Published on gartner.com, May 18, 2023. URL:
  \url{https://www.gartner.com/en/newsroom/press-releases/2023-05-18-gartner-identifies-six-chatgpt-risks-legal-and-compliance-must-evaluate}.
\newblock Accessed March 5, 2024.

\bibitem[\protect\BCAY{van~der Wal, Bachmann, Leidinger, van Maanen, Zuidema,\
  \BBA\ Schulz}{van~der Wal et~al.}{2024}]{van2024undesirable}
van~der Wal, O., Bachmann, D., Leidinger, A., van Maanen, L., Zuidema, W.,
  \BBA\ Schulz, K. \BBOP2024\BBCP.
\newblock \BBOQ Undesirable biases in nlp: Addressing challenges of
  measurement\BBCQ\
\newblock {\Bem Journal of Artificial Intelligence Research}, {\Bem 79}, 1--40.

\bibitem[\protect\BCAY{van Hooijdonk}{van
  Hooijdonk}{2021}]{hooijdonk2021chatbots}
van Hooijdonk, C. \BBOP2021\BBCP.
\newblock \BBOQ Chatbots in the tourism industry: The effects of communication
  style and brand familiarity on social presence and brand attitude\BBCQ\
\newblock In {\Bem Adjunct Proceedings of the 29th ACM Conference on User
  Modeling, Adaptation and Personalization}, UMAP '21, \BPG\ 375–381, New
  York, NY, USA. Association for Computing Machinery.

\bibitem[\protect\BCAY{van Miltenburg, Clinciu, Du{\v{s}}ek, Gkatzia, Inglis,
  Lepp{\"a}nen, Mahamood, Manning, Schoch, Thomson,\ \BBA\ Wen}{van Miltenburg
  et~al.}{2021a}]{van-miltenburg-etal-2021-underreporting}
van Miltenburg, E., Clinciu, M., Du{\v{s}}ek, O., Gkatzia, D., Inglis, S.,
  Lepp{\"a}nen, L., Mahamood, S., Manning, E., Schoch, S., Thomson, C., \BBA\
  Wen, L. \BBOP2021a\BBCP.
\newblock \BBOQ Underreporting of errors in {NLG} output, and what to do about
  it\BBCQ\
\newblock In {\Bem Proceedings of the 14th International Conference on Natural
  Language Generation}, \BPGS\ 140--153, Aberdeen, Scotland, UK. Association
  for Computational Linguistics.

\bibitem[\protect\BCAY{van Miltenburg, van~der Lee,\ \BBA\ Krahmer}{van
  Miltenburg et~al.}{2021b}]{van-miltenburg-etal-2021-preregistering}
van Miltenburg, E., van~der Lee, C., \BBA\ Krahmer, E. \BBOP2021b\BBCP.
\newblock \BBOQ Preregistering {NLP} research\BBCQ\
\newblock In {\Bem Proceedings of the 2021 Conference of the North American
  Chapter of the Association for Computational Linguistics: Human Language
  Technologies}, \BPGS\ 613--623, Online. Association for Computational
  Linguistics.

\bibitem[\protect\BCAY{Vanderlyn, Weber, Neumann, V{\"a}th, Meyer,\ \BBA\
  Vu}{Vanderlyn et~al.}{2021}]{vanderlyn2021seemed}
Vanderlyn, L., Weber, G., Neumann, M., V{\"a}th, D., Meyer, S., \BBA\ Vu, N.~T.
  \BBOP2021\BBCP.
\newblock \BBOQ “it seemed like an annoying woman”: On the perception and
  ethical considerations of affective language in text-based conversational
  agents\BBCQ\
\newblock In {\Bem Proceedings of the 25th Conference on Computational Natural
  Language Learning}, \BPGS\ 44--57.

\bibitem[\protect\BCAY{Vasconcelos, Candello, Pinhanez,\ \BBA\ dos
  Santos}{Vasconcelos et~al.}{2017}]{vasconcelos2017bottester}
Vasconcelos, M., Candello, H., Pinhanez, C., \BBA\ dos Santos, T.
  \BBOP2017\BBCP.
\newblock \BBOQ Bottester: testing conversational systems with simulated
  users\BBCQ\
\newblock In {\Bem Proceedings of the XVI Brazilian Symposium on Human Factors
  in Computing Systems}, \BPGS\ 1--4.

\bibitem[\protect\BCAY{Vazire, Schiavone,\ \BBA\ Bottesini}{Vazire
  et~al.}{2022}]{vazire2022credibility}
Vazire, S., Schiavone, S.~R., \BBA\ Bottesini, J.~G. \BBOP2022\BBCP.
\newblock \BBOQ Credibility beyond replicability: Improving the four validities
  in psychological science\BBCQ\
\newblock {\Bem Current Directions in Psychological Science}, {\Bem 31\/}(2),
  162--168.

\bibitem[\protect\BCAY{Venkatesh, Khatri, Ram, Guo, Gabriel, Nagar, Prasad,
  Cheng, Hedayatnia, Metallinou, Goel, Yang,\ \BBA\ Raju}{Venkatesh
  et~al.}{2017}]{venkatesh2017evaluating}
Venkatesh, A., Khatri, C., Ram, A., Guo, F., Gabriel, R., Nagar, A., Prasad,
  R., Cheng, M., Hedayatnia, B., Metallinou, A., Goel, R., Yang, S., \BBA\
  Raju, A. \BBOP2017\BBCP.
\newblock \BBOQ On evaluating and comparing conversational agents\BBCQ\
\newblock {\Bem CoRR}, {\Bem abs/1801.03625}.

\bibitem[\protect\BCAY{Venkatesh\ \BBA\ Davis}{Venkatesh\ \BBA\
  Davis}{2000}]{venkatesh2000theoretical}
Venkatesh, V.\BBACOMMA\  \BBA\ Davis, F.~D. \BBOP2000\BBCP.
\newblock \BBOQ A theoretical extension of the technology acceptance model:
  Four longitudinal field studies\BBCQ\
\newblock {\Bem Management Science}, {\Bem 46\/}(2), 186--204.

\bibitem[\protect\BCAY{Venkatesh, Morris, Davis,\ \BBA\ Davis}{Venkatesh
  et~al.}{2003}]{venkatesh2003user}
Venkatesh, V., Morris, M.~G., Davis, G.~B., \BBA\ Davis, F.~D. \BBOP2003\BBCP.
\newblock \BBOQ User acceptance of information technology: Toward a unified
  view\BBCQ\
\newblock {\Bem MIS Quarterly}, {\Bem 27\/}(3), 425--478.

\bibitem[\protect\BCAY{Venkatesh, Thong,\ \BBA\ Xu}{Venkatesh
  et~al.}{2012}]{venkatesh2012consumer}
Venkatesh, V., Thong, J. Y.~L., \BBA\ Xu, X. \BBOP2012\BBCP.
\newblock \BBOQ Consumer acceptance and use of information technology:
  Extending the unified theory of acceptance and use of technology\BBCQ\
\newblock {\Bem MIS Quarterly}, {\Bem 36\/}(1), 157--178.

\bibitem[\protect\BCAY{Wang, Liang, Meng, Sun, Shi, Li, Xu, Qu,\ \BBA\
  Zhou}{Wang et~al.}{2023}]{wang-etal-2023-chatgpt}
Wang, J., Liang, Y., Meng, F., Sun, Z., Shi, H., Li, Z., Xu, J., Qu, J., \BBA\
  Zhou, J. \BBOP2023\BBCP.
\newblock \BBOQ Is {C}hat{GPT} a good {NLG} evaluator? a preliminary
  study\BBCQ\
\newblock In Dong, Y., Xiao, W., Wang, L., Liu, F., \BBA\ Carenini, G.\BEDS,
  {\Bem Proceedings of the 4th New Frontiers in Summarization Workshop}, \BPGS\
  1--11, Singapore. Association for Computational Linguistics.

\bibitem[\protect\BCAY{Wang, Chen, Deng, Wen, You, Liu, Li,\ \BBA\ Li}{Wang
  et~al.}{2024}]{wang2024prompt}
Wang, L., Chen, X., Deng, X., Wen, H., You, M., Liu, W., Li, Q., \BBA\ Li, J.
  \BBOP2024\BBCP.
\newblock \BBOQ Prompt engineering in consistency and reliability with the
  evidence-based guideline for llms\BBCQ\
\newblock {\Bem npj Digital Medicine}, {\Bem 7\/}(1), 41.

\bibitem[\protect\BCAY{Wang, Li, Chen, Zhu, Lin, Cao, Liu, Liu,\ \BBA\
  Sui}{Wang et~al.}{2023}]{wang2023large}
Wang, P., Li, L., Chen, L., Zhu, D., Lin, B., Cao, Y., Liu, Q., Liu, T., \BBA\
  Sui, Z. \BBOP2023\BBCP.
\newblock \BBOQ Large language models are not fair evaluators\BBCQ\
\newblock {\Bem CoRR}, {\Bem abs/2305.17926}.

\bibitem[\protect\BCAY{Weisz, Jain, Joshi, Johnson,\ \BBA\ Lange}{Weisz
  et~al.}{2019}]{weisz2019bigbluebot}
Weisz, J.~D., Jain, M., Joshi, N.~N., Johnson, J., \BBA\ Lange, I.
  \BBOP2019\BBCP.
\newblock \BBOQ Bigbluebot: teaching strategies for successful human-agent
  interactions\BBCQ\
\newblock In {\Bem Proceedings of the 24th International Conference on
  Intelligent User Interfaces}, \BPGS\ 448--459.

\bibitem[\protect\BCAY{Weizenbaum}{Weizenbaum}{1966}]{weizenbaum_eliza}
Weizenbaum, J. \BBOP1966\BBCP.
\newblock \BBOQ Eliza—a computer program for the study of natural language
  communication between man and machine\BBCQ\
\newblock {\Bem Commun. ACM}, {\Bem 9\/}(1), 36–45.

\bibitem[\protect\BCAY{Weizenbaum}{Weizenbaum}{1976}]{Weizenbaum1976computer}
Weizenbaum, J. \BBOP1976\BBCP.
\newblock {\Bem Computer power and human reason}.
\newblock W.H. Freeman and Company, New York, NY.

\bibitem[\protect\BCAY{Wu\ \BBA\ Chien}{Wu\ \BBA\
  Chien}{2020}]{wu-chien-2020-learning}
Wu, S.-H.\BBACOMMA\  \BBA\ Chien, S.-L. \BBOP2020\BBCP.
\newblock \BBOQ Learning the human judgment for the automatic evaluation of
  chatbot\BBCQ\
\newblock In {\Bem Proceedings of the 12th Language Resources and Evaluation
  Conference}, \BPGS\ 1598--1602, Marseille, France. European Language
  Resources Association.

\bibitem[\protect\BCAY{Xiao, Zhang, Lai,\ \BBA\ Liao}{Xiao
  et~al.}{2023}]{xiao2023evaluating}
Xiao, Z., Zhang, S., Lai, V., \BBA\ Liao, Q.~V. \BBOP2023\BBCP.
\newblock \BBOQ Evaluating evaluation metrics: A framework for analyzing nlg
  evaluation metrics using measurement theory\BBCQ.

\bibitem[\protect\BCAY{Xiao, Zhou, Chen, Yang,\ \BBA\ Chi}{Xiao
  et~al.}{2020}]{xiao2020if}
Xiao, Z., Zhou, M.~X., Chen, W., Yang, H., \BBA\ Chi, C. \BBOP2020\BBCP.
\newblock \BBOQ If i hear you correctly: Building and evaluating interview
  chatbots with active listening skills\BBCQ\
\newblock In {\Bem Proceedings of the 2020 CHI Conference on Human Factors in
  Computing Systems}, \BPGS\ 1--14.

\bibitem[\protect\BCAY{Xu, Jiang, Li, Gao, Guo, Liu, Hei,\ \BBA\ Wang}{Xu
  et~al.}{2022}]{xu2020healthcare}
Xu, Y., Jiang, Y., Li, R., Gao, H., Guo, J., Liu, Y., Hei, L., \BBA\ Wang, Y.
  \BBOP2022\BBCP.
\newblock \BBOQ A healthcare-oriented mobile question-and-answering system for
  smart cities\BBCQ\
\newblock {\Bem Transactions on Emerging Telecommunications Technologies},
  {\Bem 33\/}(10), e4012.

\bibitem[\protect\BCAY{Yao\ \BBA\ Kabir}{Yao\ \BBA\
  Kabir}{2023}]{yao2023person}
Yao, L.\BBACOMMA\  \BBA\ Kabir, R. \BBOP2023\BBCP.
\newblock \BBOQ Person-centered therapy (rogerian therapy) [updated 2023 feb
  9]\BBCQ\
\newblock StatPearls [Internet]. Treasure Island (FL): StatPearls Publishing;
  2023 Jan-.

\bibitem[\protect\BCAY{Ye, Lu, Huang, Lin,\ \BBA\ Liang}{Ye
  et~al.}{2021}]{ye-etal-2021-towards-quantifiable}
Ye, Z., Lu, L., Huang, L., Lin, L., \BBA\ Liang, X. \BBOP2021\BBCP.
\newblock \BBOQ Towards quantifiable dialogue coherence evaluation\BBCQ\
\newblock In {\Bem Proceedings of the 59th Annual Meeting of the Association
  for Computational Linguistics and the 11th International Joint Conference on
  Natural Language Processing (Volume 1: Long Papers)}, \BPGS\ 2718--2729,
  Online. Association for Computational Linguistics.

\bibitem[\protect\BCAY{Yeh, Eskenazi,\ \BBA\ Mehri}{Yeh
  et~al.}{2021}]{yeh-etal-2021-comprehensive}
Yeh, Y.-T., Eskenazi, M., \BBA\ Mehri, S. \BBOP2021\BBCP.
\newblock \BBOQ A comprehensive assessment of dialog evaluation metrics\BBCQ\
\newblock In {\Bem The First Workshop on Evaluations and Assessments of Neural
  Conversation Systems}, \BPGS\ 15--33, Online. Association for Computational
  Linguistics.

\bibitem[\protect\BCAY{Yen\ \BBA\ Chiang}{Yen\ \BBA\
  Chiang}{2021}]{yen2021trust}
Yen, C.\BBACOMMA\  \BBA\ Chiang, M.-C. \BBOP2021\BBCP.
\newblock \BBOQ Trust me, if you can: a study on the factors that influence
  consumers' purchase intention triggered by chatbots based on brain image
  evidence and self-reported assessments\BBCQ\
\newblock {\Bem Behaviour \& Information Technology}, {\Bem 40\/}(11),
  1177--1194.

\bibitem[\protect\BCAY{Yin, Chang,\ \BBA\ Zhang}{Yin
  et~al.}{2017}]{yin2017deepprobe}
Yin, Z., Chang, K.-h., \BBA\ Zhang, R. \BBOP2017\BBCP.
\newblock \BBOQ Deepprobe: Information directed sequence understanding and
  chatbot design via recurrent neural networks\BBCQ\
\newblock In {\Bem Proceedings of the 23rd ACM SIGKDD International Conference
  on Knowledge Discovery and Data Mining}, \BPGS\ 2131--2139.

\bibitem[\protect\BCAY{Yuan, Moore,\ \BBA\ Grierson}{Yuan
  et~al.}{2008}]{yuan2008human}
Yuan, T., Moore, D., \BBA\ Grierson, A. \BBOP2008\BBCP.
\newblock \BBOQ A human-computer dialogue system for educational debate: A
  computational dialectics approach\BBCQ\
\newblock {\Bem International Journal of Artificial Intelligence in Education},
  {\Bem 18\/}(1), 3--26.

\bibitem[\protect\BCAY{Yuan, Moore, Ravenscroft,\ \BBA\ Zhong}{Yuan
  et~al.}{2010}]{yuan2010evaluation}
Yuan, T., Moore, D., Ravenscroft, A., \BBA\ Zhong, G. \BBOP2010\BBCP.
\newblock \BBOQ Evaluation of a human-computer dialogue system for educational
  debate\BBCQ\
\newblock In {\Bem 2010 Second WRI Global Congress on Intelligent Systems},
  \lowercase{\BVOL}~2, \BPGS\ 359--362. IEEE.

\bibitem[\protect\BCAY{Zhang, F{\o}lstad,\ \BBA\ Bj{\o}rkli}{Zhang
  et~al.}{2023}]{zhang2023organizational}
Zhang, J.~J., F{\o}lstad, A., \BBA\ Bj{\o}rkli, C.~A. \BBOP2023\BBCP.
\newblock \BBOQ Organizational factors affecting successful implementation of
  chatbots for customer service\BBCQ\
\newblock {\Bem Journal of internet commerce}, {\Bem 22\/}(1), 122--156.

\bibitem[\protect\BCAY{Zhang, Dinan, Urbanek, Szlam, Kiela,\ \BBA\
  Weston}{Zhang et~al.}{2018}]{zhang2018personalizing}
Zhang, S., Dinan, E., Urbanek, J., Szlam, A., Kiela, D., \BBA\ Weston, J.
  \BBOP2018\BBCP.
\newblock \BBOQ Personalizing dialogue agents: {I} have a dog, do you have pets
  too?\BBCQ\
\newblock In {\Bem Proceedings of the 56th Annual Meeting of the Association
  for Computational Linguistics (Volume 1: Long Papers)}, \BPGS\ 2204--2213,
  Melbourne, Australia. Association for Computational Linguistics.

\bibitem[\protect\BCAY{Zhang, Kishore, Wu, Weinberger,\ \BBA\ Artzi}{Zhang
  et~al.}{2020a}]{bert-score}
Zhang, T., Kishore, V., Wu, F., Weinberger, K.~Q., \BBA\ Artzi, Y.
  \BBOP2020a\BBCP.
\newblock \BBOQ Bertscore: Evaluating text generation with bert\BBCQ\
\newblock In {\Bem International Conference on Learning Representations}.

\bibitem[\protect\BCAY{Zhang, Takanobu, Zhu, Huang,\ \BBA\ Zhu}{Zhang
  et~al.}{2020b}]{zhang2020recent}
Zhang, Z., Takanobu, R., Zhu, Q., Huang, M., \BBA\ Zhu, X. \BBOP2020b\BBCP.
\newblock \BBOQ Recent advances and challenges in task-oriented dialog
  systems\BBCQ\
\newblock {\Bem Science China Technological Sciences}, {\Bem 63\/}(10),
  2011--2027.

\bibitem[\protect\BCAY{Zhang, Zhang,\ \BBA\ Chen}{Zhang
  et~al.}{2021}]{zhang2021informing}
Zhang, Z., Zhang, X., \BBA\ Chen, L. \BBOP2021\BBCP.
\newblock \BBOQ Informing the design of a news chatbot\BBCQ\
\newblock In {\Bem Proceedings of the 21st ACM International Conference on
  Intelligent Virtual Agents}, \BPGS\ 224--231.

\bibitem[\protect\BCAY{Zhao, Zhang, Che, Chen,\ \BBA\ Zhang}{Zhao
  et~al.}{2019}]{zhao2019evaluation}
Zhao, Z., Zhang, W., Che, W., Chen, Z., \BBA\ Zhang, Y. \BBOP2019\BBCP.
\newblock \BBOQ An evaluation of chinese human-computer dialogue
  technology\BBCQ\
\newblock {\Bem Data Intelligence}, {\Bem 1\/}(2), 187--200.

\bibitem[\protect\BCAY{Zheng, Chiang, Sheng, Zhuang, Wu, Zhuang, Lin, Li, Li,
  Xing, Zhang, Gonzalez,\ \BBA\ Stoica}{Zheng et~al.}{2023}]{zheng2023judging}
Zheng, L., Chiang, W.-L., Sheng, Y., Zhuang, S., Wu, Z., Zhuang, Y., Lin, Z.,
  Li, Z., Li, D., Xing, E.~P., Zhang, H., Gonzalez, J.~E., \BBA\ Stoica, I.
  \BBOP2023\BBCP.
\newblock \BBOQ Judging llm-as-a-judge with mt-bench and chatbot arena\BBCQ.

\bibitem[\protect\BCAY{Zhou, Blodgett, Trischler, Daum{\'e}~III, Suleman,\
  \BBA\ Olteanu}{Zhou et~al.}{2022}]{zhou-etal-2022-deconstructing}
Zhou, K., Blodgett, S.~L., Trischler, A., Daum{\'e}~III, H., Suleman, K., \BBA\
  Olteanu, A. \BBOP2022\BBCP.
\newblock \BBOQ Deconstructing {NLG} evaluation: Evaluation practices,
  assumptions, and their implications\BBCQ\
\newblock In Carpuat, M., de~Marneffe, M.-C., \BBA\ Meza~Ruiz, I.~V.\BEDS,
  {\Bem Proceedings of the 2022 Conference of the North American Chapter of the
  Association for Computational Linguistics: Human Language Technologies},
  \BPGS\ 314--324, Seattle, United States. Association for Computational
  Linguistics.

\bibitem[\protect\BCAY{Zhou, Gao, Li,\ \BBA\ Shum}{Zhou
  et~al.}{2020}]{zhou2020design}
Zhou, L., Gao, J., Li, D., \BBA\ Shum, H.-Y. \BBOP2020\BBCP.
\newblock \BBOQ The design and implementation of xiaoice, an empathetic social
  chatbot\BBCQ\
\newblock {\Bem Computational Linguistics}, {\Bem 46\/}(1), 53--93.

\bibitem[\protect\BCAY{Zhu, Zhang, Fang, Li, Takanobu, Li, Peng, Gao, Zhu,\
  \BBA\ Huang}{Zhu et~al.}{2020}]{zhu-etal-2020-convlab}
Zhu, Q., Zhang, Z., Fang, Y., Li, X., Takanobu, R., Li, J., Peng, B., Gao, J.,
  Zhu, X., \BBA\ Huang, M. \BBOP2020\BBCP.
\newblock \BBOQ {C}onv{L}ab-2: An open-source toolkit for building, evaluating,
  and diagnosing dialogue systems\BBCQ\
\newblock In {\Bem Proceedings of the 58th Annual Meeting of the Association
  for Computational Linguistics: System Demonstrations}, \BPGS\ 142--149,
  Online. Association for Computational Linguistics.

\bibitem[\protect\BCAY{Zhu, Janssen, Wang,\ \BBA\ Liu}{Zhu
  et~al.}{2022}]{zhu2022me}
Zhu, Y., Janssen, M., Wang, R., \BBA\ Liu, Y. \BBOP2022\BBCP.
\newblock \BBOQ It is me, chatbot: working to address the covid-19
  outbreak-related mental health issues in china. user experience,
  satisfaction, and influencing factors\BBCQ\
\newblock {\Bem International Journal of Human--Computer Interaction}, {\Bem
  38\/}(12), 1182--1194.

\end{thebibliography}
\bibliographystyle{theapa}

\end{document}